%% file: main.tex
\documentclass[runningheads]{llncs}

\usepackage{eccv}

\usepackage{eccvabbrv}

\usepackage{graphicx}
\usepackage{booktabs}

\usepackage{algorithm}
\usepackage{algorithmic}

\usepackage{etoc}
\usepackage{makecell}

\usepackage[accsupp]{axessibility}  
\PassOptionsToPackage{breaklinks,colorlinks,citecolor=eccvblue}{hyperref}

\usepackage{orcidlink}
\usepackage{longtable}
\usepackage{placeins}
\begin{document}

\title{Event-Driven Video Generation} 

\title{Event-Driven Video Generation} 

\author{Chika Maduabuchi\orcidlink{0000-0001-9947-5855} \and
Jindong Wang\orcidlink{0000-0002-4833-0880}\thanks{Corresponding author: \texttt{jdw@wm.edu}.}}

\authorrunning{C. Maduabuchi et al.}

\institute{William \& Mary, Williamsburg, VA, USA}

\maketitle

\begin{center}
    \centering
    \captionsetup{type=figure}
    \includegraphics[width=\textwidth]
    {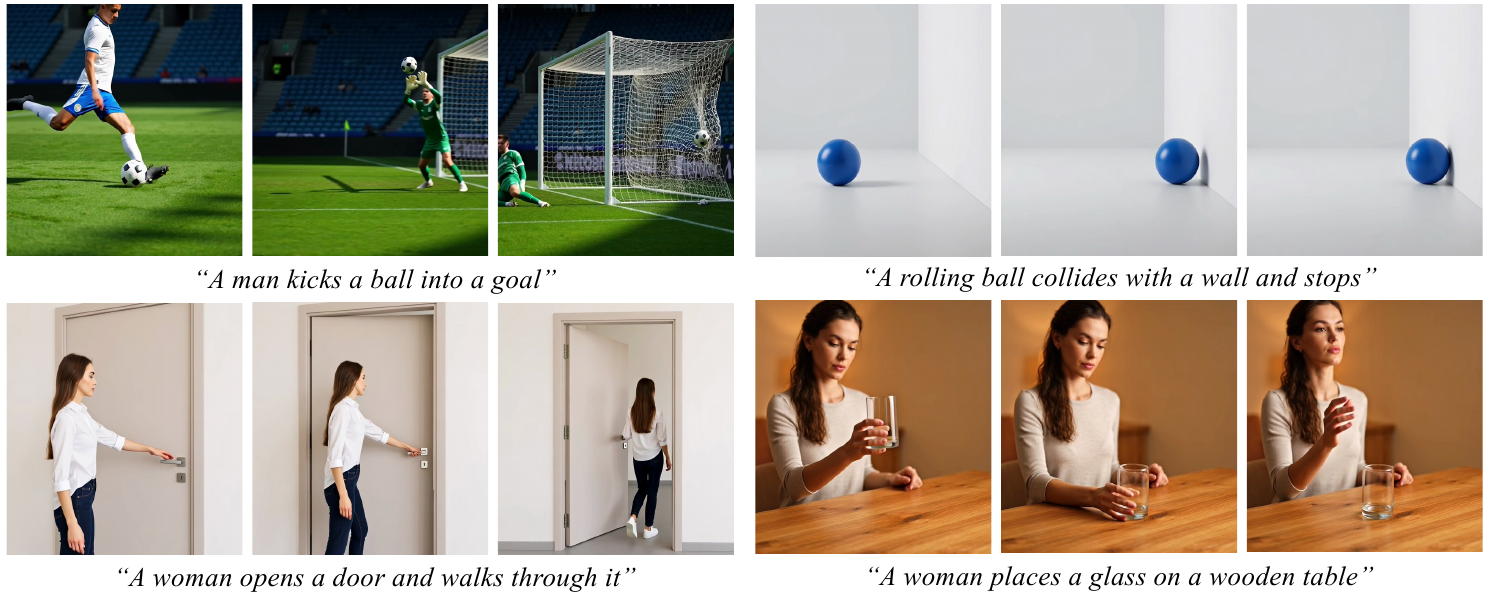}
    \captionof{figure}{\protect{\textbf{Representative text-conditioned video outputs produced by EVD.}} The samples highlight the intended effect of the event gate: latent updates are concentrated around prompt-relevant events, reducing pre-contact motion, missing interaction responses, and post-event drift.}
    \label{fig:teaser}
\end{center}

\input{sections/0_abstract}
\input{sections/1_introduction}

\input{sections/2_motivation}
\input{sections/3_related_work}
\input{sections/3_contributions}
\input{sections/4_evd}
\input{sections/5_experiments}

\FloatBarrier


%
%
\bibliographystyle{splncs04}
\bibliography{main}


\newpage
\appendix
\renewcommand{\theHsection}{appendix.\arabic{section}}
\renewcommand{\theHsubsection}{appendix.\arabic{section}.\arabic{subsection}}
\renewcommand{\theHsubsubsection}{appendix.\arabic{section}.\arabic{subsection}.\arabic{subsubsection}}
\onecolumn

\input{sections/appendix}

\end{document}

%% file: sections/0_abstract.tex
\begin{abstract}
Current text-to-video models can make individual frames look convincing while still getting simple interactions wrong: objects move before contact, an intended action is skipped, a placed object keeps drifting, or a support relation breaks. Our starting point is that standard \emph{frame-first} denoising updates every latent region at every step, even when the prompt implies that only a local interaction should be active. We introduce \emph{Event-Driven Video Generation (EVD)}, a small DiT-compatible intervention that gives the sampler an explicit event signal. A lightweight head predicts token-level event activity; training losses tie that activity to latent state change; and event-gated sampling, with hysteresis and an early-step schedule, applies the update field mainly where an interaction is forming. On EVD-Bench, EVD improves human preference and VBench dynamics for \emph{state persistence, spatial accuracy, support relations,} and \emph{contact stability}, while keeping appearance quality comparable to the base model. The results suggest that a modest amount of event structure can correct several interaction failures that otherwise remain hidden behind good frame-level appearance. Project webpage: \url{https://evd-project-website.pages.dev}

\keywords{Text-to-video generation \and Event grounding \and Video diffusion transformers}
\end{abstract}

%% file: sections/1_introduction.tex
\section{Introduction}
\label{sec:intro}

Text-to-video has improved quickly in frame realism, resolution, and clip length. Space--time diffusion models such as Lumiere~\cite{10.1145/3680528.3687614} can synthesize coherent clips in one pass, and large diffusion-transformer systems such as Movie~Gen~\cite{polyak2024movie} and Step-Video-T2V~\cite{ma2025stepvideot2vtechnicalreportpractice} scale latent representations and training recipes to longer, higher-fidelity videos. Open and semi-open systems, including HunyuanVideo~\cite{kong2025hunyuanvideosystematicframeworklarge}, Open-Sora~\cite{zheng2024opensorademocratizingefficientvideo}, and Wan~\cite{wan2025wanopenadvancedlargescale}, together with faster samplers such as Pyramidal Flow Matching~\cite{jin2025pyramidal} and T2V-Turbo/T2V-Turbo-v2~\cite{li2024tvturbo,li2025tvturbov}, make these capabilities more broadly available. As the frames become more convincing, the remaining errors are often no longer about texture or sharpness; they are about whether an interaction actually happened in the right place and in the right order.

%% file: sections/2_motivation.tex
\section{Motivation}
\label{sec:motivation}

\begin{figure*}[t!]
\centering
\includegraphics[width=0.98\textwidth]{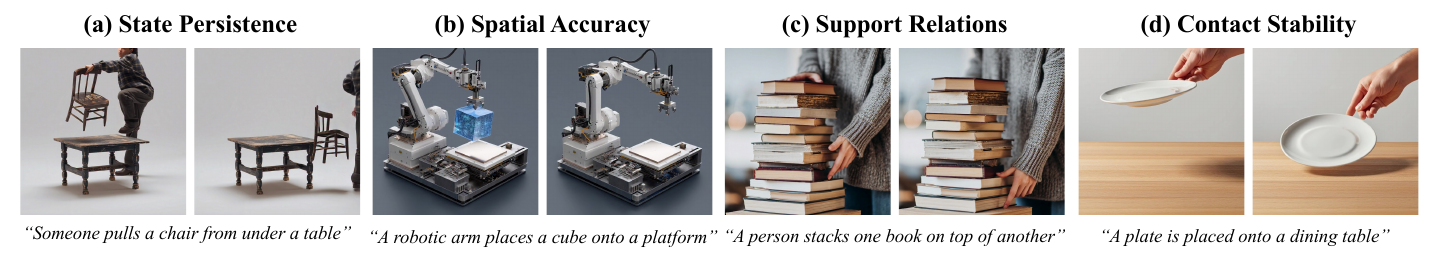}
\caption{\textbf{Four interaction failures we observe in DiT-30B.}
The examples isolate mistakes that are easy to miss when judging only single-frame quality:
(a) state persistence, where the chair keeps moving after the pull has ended;
(b) spatial accuracy, where the cube never aligns with the target platform;
(c) support relations, where the book appears stacked without a visible stacking action;
and (d) contact stability, where the plate begins moving before hand--object contact.
}
\label{fig:failures}
\end{figure*}

These gains do not remove a basic interaction problem. A clip can be smooth locally while the cause--effect story is wrong: effects appear before their causes, contact is missing, or the object keeps changing after the action should be over. Recent evaluation suites separate appearance from motion, physics, and temporal consistency, making the same point quantitatively: good-looking frames do not guarantee realistic dynamics (VBench~\cite{Huang_2024_CVPR}, VBench++~\cite{11250949}, VBench-2.0~\cite{zheng2025vbench20advancingvideogeneration}). In our experiments, this is also where large models remain fragile: they may score well on appearance and still fail on interaction grounding, which suggests that scaling alone is not a reliable fix~\cite{qin2025worldsimbench}.

Recent systems improve dynamics with stronger architectures and training/inference recipes (CogVideoX~\cite{yang2025cogvideox}, Open-Sora STDiT~\cite{zheng2024opensorademocratizingefficientvideo}, HunyuanVideo~\cite{kong2025hunyuanvideosystematicframeworklarge}, Step-Video-T2V~\cite{ma2025stepvideot2vtechnicalreportpractice}), and with motion-aware objectives or inference-time steering that reduce the appearance--motion imbalance (VideoJAM~\cite{pmlr-v267-chefer25a}). For the interaction failures in Fig.~\ref{fig:failures}, however, a common issue remains: the sampler is still mostly ``frame-first,'' updating the latent state everywhere at every step. That makes it easy for a locally plausible update to occur before the relevant contact, to miss the actual state change, or to keep drifting after the event. We group these errors into State Persistence, Spatial Accuracy, Support Relations, and Contact Stability. The mechanism we want is correspondingly simple: the model should know whether an interaction is active \emph{here} and \emph{now}, and it should use that signal to decide where the latent is allowed to change.

%% file: sections/3_related_work.tex
\section{Related Work}
\label{sec:related_work}

\paragraph{Video generation and evaluation.}
Recent text-to-video systems have advanced through space--time diffusion, large video DiT backbones, temporal autoencoders, and efficient flow/diffusion sampling~\cite{10.1145/3680528.3687614,polyak2024movie,ma2025stepvideot2vtechnicalreportpractice,kong2025hunyuanvideosystematicframeworklarge,zheng2024opensorademocratizingefficientvideo,wan2025wanopenadvancedlargescale,jin2025pyramidal,yang2025cogvideox}. The remaining interaction errors are often specific: motion starts too early, contact is weak, support relations are skipped, or the post-event state keeps moving. Evaluation suites such as VBench, VBench++, VBench-2.0, WorldSimBench, T2V-CompBench, and NeuS-V make a similar distinction between visual quality and temporally faithful or physically meaningful generation~\cite{Huang_2024_CVPR,11250949,zheng2025vbench20advancingvideogeneration,qin2025worldsimbench,sun2025t2vcompbench,sharan2025neusv}. EVD addresses these errors by coupling latent updates to prompt-relevant event activity.

\paragraph{Event structure, motion steering, and editing.}
The closest event-centric prior work is GEST, which represents visual/language stories as Graphs of Events in Space and Time~\cite{masala2023gest}, and GEST-Engine, which executes formal GEST specifications to synthesize controllable multi-actor videos with dense spatiotemporal annotations~\cite{cudlenco2026gestengine}. These works motivate events as an abstraction, but rely on symbolic/event-graph specifications or simulation-style control. EVD instead learns token-aligned event activity inside a pretrained video DiT and uses it to gate the sampled direction field. EVD also differs from motion-steering and editing/correction pipelines such as VideoJAM~\cite{pmlr-v267-chefer25a}, StreamDiffusion~\cite{kodaira2025streamdiffusion}, StreamV2V~\cite{liang2025streamv2v}, and ObjectAlign~\cite{munir2025objectalign}: it is a text-to-video generation-time mechanism that keeps the solver, decoder, NFE, and output-selection protocol unchanged while modifying only event-grounded training and the direction field.

%% file: sections/3_contributions.tex
\section{Contributions}
\label{sec:contributions}
We propose \textit{Event-Driven Video Generation (EVD)} as a small add-on for pretrained video DiTs rather than a replacement backbone. EVD makes three changes. First, it attaches a lightweight event head to DiT token features and predicts a token-aligned activity map. Second, it trains the model so latent updates are tied to that activity: inactive regions are discouraged from changing, while active interaction regions receive more stable updates. Third, at sampling time, it gates the solver direction field with a hysteresis-and-schedule rule, so early event formation is allowed but late spurious drift is damped. This keeps the solver family and decoder unchanged, which makes the method compatible with modern DiT video systems~\cite{polyak2024movie,ma2025stepvideot2vtechnicalreportpractice,yang2025cogvideox,zheng2024opensorademocratizingefficientvideo}. The main limitation we observe is also specific: if the event signal is weak at the operating resolution, as in small contacts, occlusion, or cluttered multi-object scenes, EVD can under-localize the interaction. This points to object-centric or contact-aware event cues as natural next steps.

%% file: sections/4_evd.tex
\section{EVD}
\label{sec:evd}

This section describes how EVD changes a pretrained video DiT without replacing its backbone. The DiT still predicts a latent update field, but we add a token-level event pathway that decides where that update should be trusted. We start with the latent-video interface (Sec.~\ref{sec:method_latent}), define the event head and gated update rule (Sec.~\ref{sec:method_event_core}), then specify the soft/hysteretic gate and the resulting direction field (Sec.~\ref{sec:method_gate}). The training objective adds event realization, consistency, ordering, and time-weighting terms on top of Flow Matching (Sec.~\ref{sec:method_training}), and inference uses the same idea under CFG with a scheduled gate (Sec.~\ref{sec:method_sampling}). Appendix~\ref{app:algorithms} gives the exact training and sampling pseudocode, and Sec.~\ref{sec:method_practical} lists the hyperparameters used in the experiments.

\begin{figure*}[t]
    \centering
    \includegraphics[width=\textwidth]{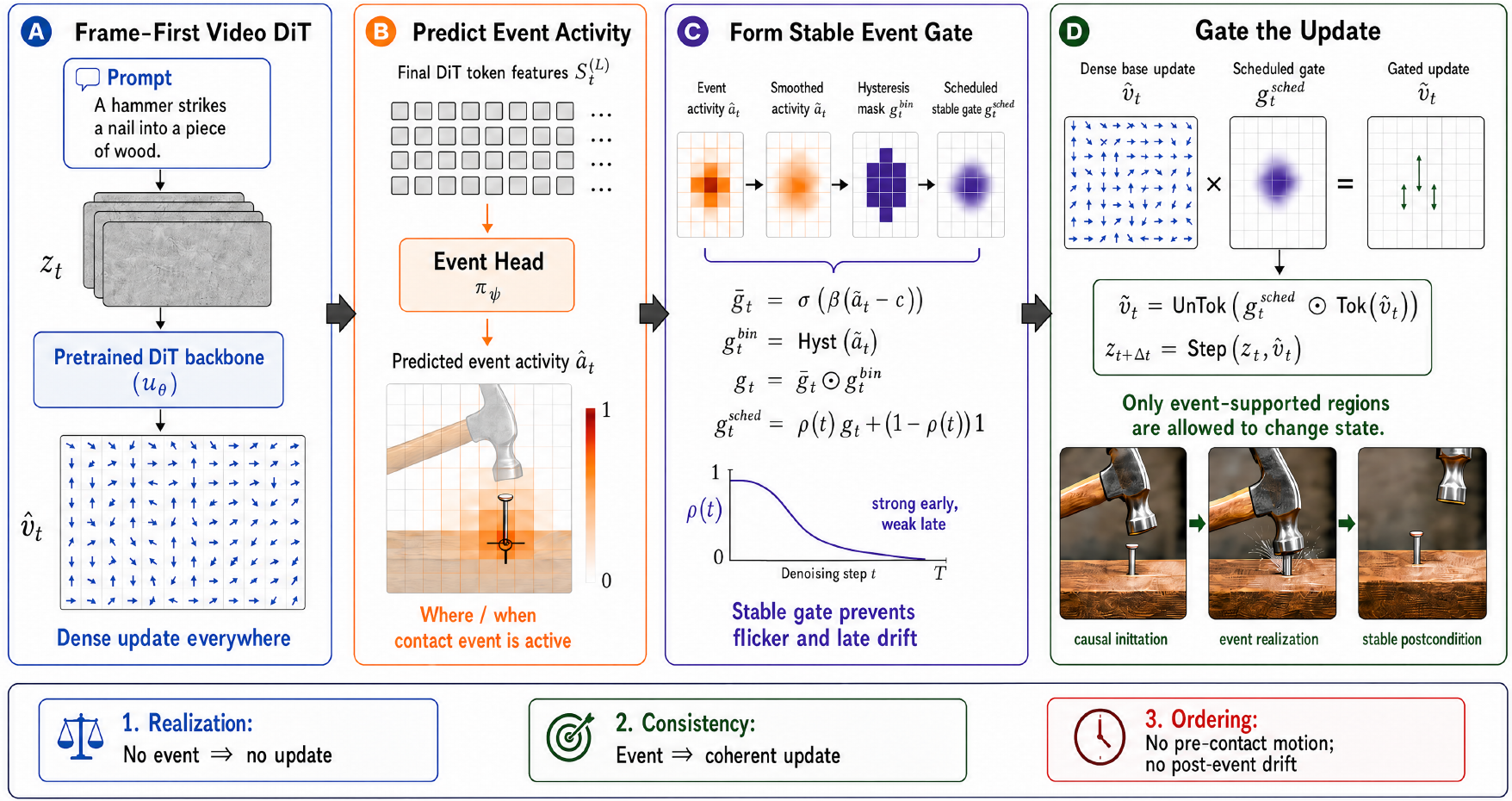}
    \caption{\textbf{Overview of Event-Driven Video Generation (EVD).}
    Given a noised latent video \(z_t\) and prompt \(y\), the DiT predicts its usual update field \(\widehat{v}_t\). A lightweight event head predicts token-aligned activity \(\hat{a}_t\), which we smooth and convert into a gate using soft activation, hysteresis, and an early-step schedule. The solver and decoder are unchanged; the intervention is that the update field is no longer applied uniformly everywhere. During training, the event losses tie active regions to state change and discourage unrelated regions from moving.}
    \label{fig:evd_overview}
\end{figure*}

\subsection{Method: Event-Driven Video Generation (EVD)}
\label{sec:method}

Figure~\ref{fig:evd_overview} summarizes EVD, from token-aligned event activity prediction and stable gate formation to gated latent updates during sampling.

\subsubsection{Latent video representation and notation}
\label{sec:method_latent}

We consider text-conditioned video generation. A video clip is denoted by
\(x \in \mathbb{R}^{T \times H \times W \times 3}\) with \(T\) frames.
Following standard practice in large-scale video diffusion/flow models, we generate in a compressed latent space using a temporal video autoencoder.
Let \(E(\cdot)\) and \(D(\cdot)\) denote the encoder and decoder, and define the latent video
\begin{equation}
    z = E(x), \qquad x \approx D(z),
\end{equation}
where \(z \in \mathbb{R}^{T' \times H' \times W' \times C}\) has reduced spatiotemporal resolution.

Let \(y\) denote the text prompt, encoded by a frozen text encoder into a sequence of embeddings. We denote the DiT backbone by
\(u_\theta(\cdot)\), parameterized by \(\theta\), which operates on noised latents and conditions on \((y,t)\), where
\(t \in [0,1]\) is the continuous diffusion/flow time.

\paragraph{Noise model and training target.}
We use a continuous-time Flow Matching formulation in latent space~\cite{lipman2023flow,maduabuchi2026temporalpairconsistencyvariancereduced,maduabuchi2026entropycontrolledflowmatching}. Given a clean latent \(z_1\) and noise \(z_0 \sim \mathcal{N}(0,I)\),
we sample \(t \sim \mathcal{U}[0,1]\) and form the interpolated latent
\begin{equation}
\label{eq:method_interp}
    z_t = t\, z_1 + (1-t)\, z_0,
\end{equation}
with velocity target
\begin{equation}
\label{eq:method_velocity}
    v_t = \frac{d z_t}{dt} = z_1 - z_0.
\end{equation}
The backbone predicts \(\widehat{v}_t = u_\theta(z_t,y,t)\). We keep this interface fixed and modify how the model represents
and applies \emph{event-driven} state changes in the subsequent subsections.

\subsubsection{DiT backbone and conditioning interface}
\label{sec:method_backbone}

EVD is built on a pretrained video Diffusion Transformer (DiT) operating in latent space.
Given \(z_t \in \mathbb{R}^{T' \times H' \times W' \times C}\), we patchify \(z_t\) into spatiotemporal tokens and map them to the model width \(d\),
yielding a token sequence \(\mathbf{s}_t \in \mathbb{R}^{N \times d}\), where \(N\) is the number of spatiotemporal patches.
The DiT applies \(L\) transformer blocks with spatiotemporal self-attention and text cross-attention, producing final features
\(\mathbf{s}_t^{(L)}\).

The DiT backbone outputs a prediction of the Flow Matching velocity (Eq.~\eqref{eq:method_velocity}):
\begin{equation}
\label{eq:method_backbone_pred}
    \widehat{v}_t \;=\; u_\theta(z_t, y, t).
\end{equation}
We do not modify the backbone architecture or its tokenization; EVD introduces an additional lightweight \emph{event pathway} that
predicts event activity aligned with the same token grid and uses it to gate updates during training and sampling (Secs.~\ref{sec:method_event}--\ref{sec:method_sampling}).

\paragraph{Sampling interface.}
Sampling evolves a latent trajectory \({\{z_{t_k}\}}_{k=0}^{K}\) along a monotone time grid \(0=t_0<\cdots<t_K=1\).
At each step, the sampler queries the backbone to obtain a direction field and updates \(z_{t_k}\) using a solver step
(Euler/Heun/DPM-style~\cite{zheng2023dpmsolverv}). EVD is compatible with any such solver because it only changes the direction field passed to the solver.

\subsubsection{Classifier-free guidance (CFG)}
\label{sec:method_cfg}

We use classifier-free guidance (CFG) to improve prompt adherence. At each sampling step \(t_k\), we evaluate the backbone
with the prompt \(y\) and with a null prompt \(\varnothing\), and form the guided direction field
\begin{equation}
\label{eq:method_cfg}
    \widehat{v}^{\mathrm{cfg}}(z_{t_k},y,t_k)
    \;=\;
    (1+w_{\mathrm{cfg}})\,u_\theta(z_{t_k},y,t_k)
    \;-\;
    w_{\mathrm{cfg}}\,u_\theta(z_{t_k},\varnothing,t_k),
\end{equation}
where \(w_{\mathrm{cfg}}\ge 0\) is the guidance scale. EVD applies event gating \emph{after} forming
\(\widehat{v}^{\mathrm{cfg}}\), so prompt adherence is preserved while spurious, event-inconsistent dynamics are suppressed
(Sec.~\ref{sec:method_sampling}).

\subsubsection{Why frame-first generation fails on interactions}
\label{sec:method_motivation}

Although modern video generators can produce locally smooth motion~\cite{10.1145/3680528.3687614,polyak2024movie,kong2025hunyuanvideosystematicframeworklarge,zheng2024opensorademocratizingefficientvideo,ma2025stepvideot2vtechnicalreportpractice,pmlr-v267-chefer25a,huang2025stepvideoti2vtechnicalreportstateoftheart,wu2025hunyuanvideo15technicalreport}, they often violate basic causal structure in simple interactions:
effects appear without causes (e.g., objects move before contact), causes occur without coherent effects (e.g., an interaction is implied but the state does not respond),
and post-interaction states drift instead of settling. These errors are particularly salient in prompts involving contact, support, constrained mechanisms,
or material transfer, and they map directly to the four failure categories used in our analysis:
\emph{State Persistence}, \emph{Spatial Accuracy}, \emph{Support Relations}, and \emph{Contact Stability}.

\subsection{Event-Driven Video Generation}
\label{sec:method_event}

\subsubsection{Core idea: event-gated state updates}
\label{sec:method_event_core}

EVD models a video as persistent latent state punctuated by discrete interaction events.
At diffusion/flow time \(t\), we introduce an \emph{event representation} \(e_t\) aligned with the DiT token grid, and derive from it an
event gate \(g_\psi(e_t,t)\in{[0,1]}^N\) (token-wise, broadcast across channels). The gate modulates the backbone direction field so that
state updates occur only when justified by an active interaction:
\begin{equation}
\label{eq:evd_core_update}
    \Delta z_t \;=\; \mathrm{UnTok}\!\Big(g_\psi(e_t,t)\odot \mathrm{Tok}\big(u_\theta(z_t,y,t)\big)\Big).
\end{equation}
When the event signal indicates ``no interaction'' (gate near zero), the update is suppressed and the state remains stable; when an event is active
(gate near one), the update proceeds normally. This single mechanism targets both common degeneracies: \emph{missing events}
(state changes without a visible interaction) and \emph{ghost events} (an implied interaction without a coherent state change).

In the remainder of this section, we specify (i) how \(e_t\) and \(g_\psi\) are instantiated, (ii) the event-grounded training objective, and
(iii) the event-driven sampling procedure with CFG.

\subsubsection{Event representation and event head}
\label{sec:method_event_rep}

EVD uses a token-aligned event field to localize interactions in space and time. Let \(\mathbf{s}_t^{(L)}\in\mathbb{R}^{N\times d}\) be the final
DiT token features at time \(t\). We attach a lightweight event head \(\pi_\psi\) that predicts an event field
\begin{equation}
\label{eq:method_event_head}
    \hat{e}_t \;=\; \pi_\psi(\mathbf{s}_t^{(L)},t) \;\in\; \mathbb{R}^{N\times C_e},
\end{equation}
with a small channel budget \(C_e\) (we use \(C_e=1\) in the main method). The first channel is interpreted as an \emph{event activity} logit,
and we obtain a token-wise activity probability via
\begin{equation}
\label{eq:method_event_activity}
    \hat{a}_t \;=\; \sigma(\hat{e}_t^{(1)}) \;\in\; {[0,1]}^N,
\end{equation}
where \(\sigma(\cdot)\) is the sigmoid and \(\hat{e}_t^{(1)}\) denotes the activity channel.

\paragraph{Zero-impact initialization.}
To preserve pretrained DiT behavior at the start of fine-tuning, we initialize \(\pi_\psi\) to near-zero output so that \(\hat{a}_t\approx 0\)
initially, and the model reduces to the base backbone before learning event structure.

\paragraph{Spatial smoothing.}
Event activity can be spatially fragmented due to noise. We apply a lightweight smoothing operator \(\mathcal{S}\) over the spatial patch grid
(per frame) and use \(\tilde{a}_t=\mathcal{S}(\hat{a}_t)\) in all gating computations. In our main setting, \(\mathcal{S}\) is a \(3\times 3\) average filter.

\subsubsection{Event gate: soft activation with hysteresis}
\label{sec:method_gate}

We convert the smoothed activity \(\tilde{a}_t\in{[0,1]}^N\) into a stable, token-wise gate \(g_t\in{[0,1]}^N\) that controls whether each token is
allowed to update. EVD uses \emph{soft activation} to avoid brittle thresholding and \emph{hysteresis} to prevent flickering event boundaries.

\paragraph{Soft activation.}
We first form a soft gate centered between the on/off thresholds:
\begin{equation}
\label{eq:method_soft_gate}
    \bar{g}_t \;=\; \sigma\!\Big(\beta\big(\tilde{a}_t - \tfrac{\tau_{\mathrm{on}}+\tau_{\mathrm{off}}}{2}\big)\Big),
\end{equation}
where \(\beta>0\) controls sharpness and \(\tau_{\mathrm{on}}>\tau_{\mathrm{off}}\) define the hysteresis band.

\paragraph{Hysteresis update.}
We maintain a binary state gate \(g_t^{\mathrm{bin}}\in{\{0,1\}}^N\) with token-wise update:
\begin{equation}
\label{eq:method_hysteresis}
    g^{\mathrm{bin}}_{t,i} \;=\;
    \begin{cases}
    1, & \tilde{a}_{t,i} \ge \tau_{\mathrm{on}},\\
    0, & \tilde{a}_{t,i} \le \tau_{\mathrm{off}},\\
    g^{\mathrm{bin}}_{t^{-},i}, & \text{otherwise},
    \end{cases}
\end{equation}
where \(t^{-}\) denotes the previous sampling step (or previous iteration in the discretized schedule) and \(i\) indexes tokens.
Finally, we combine soft and hysteresis gates to obtain the effective gate used for modulation:
\begin{equation}
\label{eq:method_gate_final}
g_t \;=\; \bar g_t \odot g_t^{\mathrm{bin}} ,
\end{equation}
\noindent
Here \(g_t^{\mathrm{bin}}\) provides stable on/off event activation (prevents flicker), while \(\bar g_t\) smoothly scales update magnitude within active regions; scheduled gating is applied afterward during sampling (Sec.~\ref{sec:method_schedule}).

In practice this reduces to using the hysteresis state for stability while retaining smooth transitions through \(\bar{g}_t\).
(Algorithm~\ref{alg:evd_sampling} provides the exact implementation used in our experiments.)

\subsubsection{Event-gated update field}
\label{sec:method_gated_field}

Given the backbone prediction \(\widehat{v}_t = u_\theta(z_t,y,t)\) and the event gate \(g_t\in{[0,1]}^N\), EVD forms an event-gated direction field
by modulating the patchified backbone output and unpatchifying back to latent space:
\begin{equation}
\label{eq:method_gated_field}
    \widetilde{v}_t \;=\; \mathrm{UnTok}\!\Big(g_t \odot \mathrm{Tok}\big(\widehat{v}_t\big)\Big).
\end{equation}
We then pass \(\widetilde{v}_t\) to the same solver used by the base model. Because EVD only changes the direction field, it is compatible with any
ODE sampler (Euler/Heun/DPM-style) used for DiT video generation.


\subsection{Training Objective}
\label{sec:method_training}

We keep the base Flow Matching objective (Eqs.~\eqref{eq:method_interp}--\eqref{eq:method_velocity}) and add event-grounded terms that couple
event activity to state evolution. Let \(\widehat{v}_t=u_\theta(z_t,y,t)\) be the predicted velocity and \(\Delta_t=\mathrm{Tok}(\widehat{v}_t)\) its token form.

\subsubsection{Base Flow Matching loss}
\label{sec:method_base_loss}

The base loss matches the predicted velocity to the target \(v_t=z_1-z_0\):
\begin{equation}
\label{eq:method_base_loss}
    \mathcal{L}_{\mathrm{base}}
    \;=\;
    \mathbb{E}_{z_1,z_0,t,y}\Big[\big\|u_\theta(z_t,y,t) - (z_1-z_0)\big\|_2^2\Big].
\end{equation}

\subsubsection{Event realization loss}
\label{sec:method_real_loss}

To prevent \emph{missing events} (state changes without an active interaction), we penalize update energy in tokens where the event activity is low:
\begin{equation}
\label{eq:method_real_loss}
    \mathcal{L}_{\mathrm{real}}
    \;=\;
    \mathbb{E}\Big[\big\|(1-\tilde{a}_t)\odot \Delta_t\big\|_2^2\Big],
\end{equation}
where \(\tilde{a}_t\) is the smoothed event activity (Sec.~\ref{sec:method_event_rep}).
This term forces the model to either (i) predict an active event where a state change is required, or (ii) suppress the state change when no event is present.

\subsubsection{Event consistency loss}
\label{sec:method_cons_loss}

To prevent \emph{ghost events} and reduce jitter during interactions, we enforce that state updates under active events are locally consistent across nearby
diffusion/flow times. For each training sample we draw a second time \(t'=\mathrm{clip}(t+\delta,0,1)\) with \(\delta\sim\mathcal{U}[-\Delta,\Delta]\),
construct \(z_{t'}=t' z_1+(1-t')z_0\), and compute
\(\Delta_t=\mathrm{Tok}(u_\theta(z_t,y,t))\), \(\Delta_{t'}=\mathrm{Tok}(u_\theta(z_{t'},y,t'))\),
with corresponding activities \(\tilde{a}_t\) and \(\tilde{a}_{t'}\).
We then minimize the event-masked discrepancy:
\begin{equation}
\label{eq:method_cons_loss}
    \mathcal{L}_{\mathrm{cons}}
    \;=\;
    \mathbb{E}\Big[\big\|\tilde{a}_t\odot \Delta_t - \tilde{a}_{t'}\odot \Delta_{t'}\big\|_2^2\Big].
\end{equation}
Intuitively, once an interaction is active, the model should not oscillate between incompatible update directions across infinitesimally close times;
\(\mathcal{L}_{\mathrm{cons}}\) encourages stable, directed state evolution during the event.

\subsubsection{Ordering and termination loss}
\label{sec:method_order_loss}

EVD additionally enforces a simple causal ordering: motion should not occur before event initiation and should decay after termination. Using the same
thresholds that define the hysteresis band (\(\tau_{\mathrm{on}}>\tau_{\mathrm{off}}\)), we suppress update energy in low-activity regions:
\begin{equation}
\label{eq:method_order_loss}
\mathcal{L}_{\mathrm{order}}
\;=\;
\mathbb{E}\Big[
\big\|\mathbf{1}[\tilde{a}_t<\tau_{\mathrm{on}}]\odot \Delta_t\big\|_2^2
+
\big\|\mathbf{1}[\tilde{a}_t<\tau_{\mathrm{off}}]\odot \Delta_t\big\|_2^2
\Big],
\end{equation}
where \(\mathbf{1}[\cdot]\) is applied token-wise. The first term discourages pre-event motion (improving \emph{Contact Stability});
the second term suppresses residual updates after the model indicates the event is off (improving \emph{State Persistence}).

\subsubsection{Time-weighted objective}
\label{sec:method_full_loss}

Event grounding matters most at early diffusion/flow times that determine coarse dynamics. We therefore apply a time weight
\begin{equation}
\label{eq:method_timeweight}
w(t)=\mathbf{1}[t\le t^\star_{\mathrm{loss}}]+\exp\!\big(-\kappa(t-t^\star_{\mathrm{loss}})\big)\mathbf{1}[t>t^\star_{\mathrm{loss}}],
\end{equation}
and optimize the total objective
\begin{equation}
\label{eq:method_total_loss}
\mathcal{L}
\;=\;
\mathcal{L}_{\mathrm{base}}
+
w(t)\Big(
\lambda_{\mathrm{real}}\mathcal{L}_{\mathrm{real}}
+
\lambda_{\mathrm{cons}}\mathcal{L}_{\mathrm{cons}}
+
\lambda_{\mathrm{order}}\mathcal{L}_{\mathrm{order}}
\Big).
\end{equation}
Appendix~\ref{app:algorithms} provides the complete training procedure.

\subsection{Inference: Event-Driven Sampling}
\label{sec:method_sampling}

At inference, EVD uses the same sampler and decoder as the base DiT model, but replaces the direction field passed to the solver with an
event-gated field. This change directly suppresses pre-contact motion and post-interaction drift while preserving prompt adherence via CFG.

\subsubsection{Scheduled event gating}
\label{sec:method_schedule}

Event grounding is most important early in sampling, where coarse motion and interaction structure are established. We therefore apply event gating
strongly for early steps and anneal it later. Given sampling time \(t_k\), we define
\begin{equation}
\label{eq:method_rho}
\rho(t_k)=
\begin{cases}
1, & t_k \le t^\star,\\
1-\dfrac{t_k-t^\star}{1-t^\star}, & t_k>t^\star,
\end{cases}
\end{equation}
and combine it with the gate \(g_k\) to obtain the scheduled gate
\begin{equation}
\label{eq:method_sched_gate}
g^{\mathrm{sched}}_k
\;=\;
\rho(t_k)\, g_k + (1-\rho(t_k))\,\mathbf{1},
\end{equation}
where \(\mathbf{1}\) is the all-ones gate (no gating). When \(\rho=1\), EVD applies full event gating; when \(\rho=0\), sampling reduces to the base model.

\subsubsection{Event-driven solver update}
\label{sec:method_solver}

At each step we compute the hysteresis gate \(g^{\mathrm{bin}}_k\) (Eq.~\eqref{eq:method_hysteresis}) and the soft gate \(\bar g_k\) (Eq.~\eqref{eq:method_soft_gate}), set \(g_k=\bar g_k\odot g^{\mathrm{bin}}_k\) (Eq.~\eqref{eq:method_gate_final}), and then apply the scheduled gate \(g^{\mathrm{sched}}_k\) (Eq.~\eqref{eq:method_sched_gate}).
We then pass the gated direction field to the base solver:
\begin{equation}
\label{eq:method_infer_field}
\widetilde{v}_{t_k}
\;=\;
\mathrm{UnTok}\!\Big(g^{\mathrm{sched}}_k \odot \mathrm{Tok}\big(\widehat{v}^{\mathrm{cfg}}(z_{t_k},y,t_k)\big)\Big),
\end{equation}
\begin{equation}
\label{eq:method_solver_step}
z_{t_{k+1}} \;=\; \mathrm{Step}(z_{t_k}, \widetilde{v}_{t_k}, t_k, t_{k+1}),
\end{equation}
where \(\mathrm{Step}(\cdot)\) is any solver step used by the base model (Euler/Heun/DPM-style). Algorithm~\ref{alg:evd_sampling} provides the complete
sampling procedure.

\subsection{Practical settings}
\label{sec:method_practical}

For all experiments, we use the same latent clip format and sampling interface described in Sec.~\ref{sec:method_latent}.
Unless otherwise stated, we use \(K=50\) sampling steps with CFG scale \(w_{\mathrm{cfg}}=4.0\).
Event gating uses \(\beta=12.0\) and hysteresis thresholds \(\tau_{\mathrm{on}}=0.62\), \(\tau_{\mathrm{off}}=0.38\), with \(3\times 3\) spatial smoothing on the patch grid.
We apply scheduled gating with cutoff \(t^\star=0.60\), i.e., full gating for early steps and linear annealing thereafter (Sec.~\ref{sec:method_schedule}).
For training, we set \(t^\star_{\mathrm{loss}}=0.60\) and \(\kappa=6\) in the time-weight \(w(t)\) (Eq.~\eqref{eq:method_timeweight}),
use event dropout \(p_e=0.25\), and optimize Eq.~\eqref{eq:method_total_loss} with \(\lambda_{\mathrm{real}}=0.12\),
\(\lambda_{\mathrm{cons}}=0.08\), and \(\lambda_{\mathrm{order}}=0.03\).
These values are sufficient to reproduce the qualitative behaviors in our figures and the quantitative gains on EVD-Bench; the appendix provides compute/data details, extended ablations and sensitivity analyses.

%% file: sections/5_experiments.tex
\section{Experiments}
\label{sec:experiments}
We evaluate EVD on prompts where the important question is not just whether the frames look plausible, but whether the interaction happens in the right order. The section covers the EVD-Bench setup, qualitative comparisons, human and automatic metrics, and the ablations needed to separate event grounding from a generic motion mask.

\subsection{Setup}
\label{sec:exp_setup}

We evaluate on EVD-Bench, a curated set of 150 short interaction-centric prompts that stress causal event realization, grouped into four failure categories used throughout (Fig.~\ref{fig:failures}): \emph{State Persistence}, \emph{Spatial Accuracy}, \emph{Support Relations}, and \emph{Contact Stability}. Our primary baselines are pretrained DiT-4B and DiT-30B, with DiT-4B+EVD and DiT-30B+EVD applying EVD as an additive modification (lightweight event head + event-driven training/sampling) without changing transformer blocks; Fig.~\ref{fig:comparisons} additionally includes strong external video generators for qualitative reference. All methods generate 128-frame clips at 24\,fps with a 256$\times$256 base generation resolution in the temporal-autoencoder latent/decoder space (upsampled to 720p for visualization), using a matched solver, step budget \(K\) (NFE), and CFG scale \(w_{\mathrm{cfg}}\); critically, EVD changes only the direction field passed to the sampler (event gating) and uses identical decoding with no post-hoc filtering. We report automatic VBench \emph{Appearance}/\emph{Dynamics} and human 2AFC preferences over \emph{Text Faithfulness}, \emph{Quality}, and \emph{Dynamics}, running each model once per prompt with a fixed seed and evaluating the first sample (no cherry-picking). Human-eval details, EVD-Bench construction/leakage safeguards, and closed-source normalization are provided in Appendices~\ref{app:human_eval},~\ref{app:evd-bench_construction}, and~\ref{app:external_fairness}.

\vspace{-6px}
\subsection{Qualitative Results}
\label{sec:exp_qual}

\begin{figure*}[ht!]
\centering
\includegraphics[width=\textwidth]{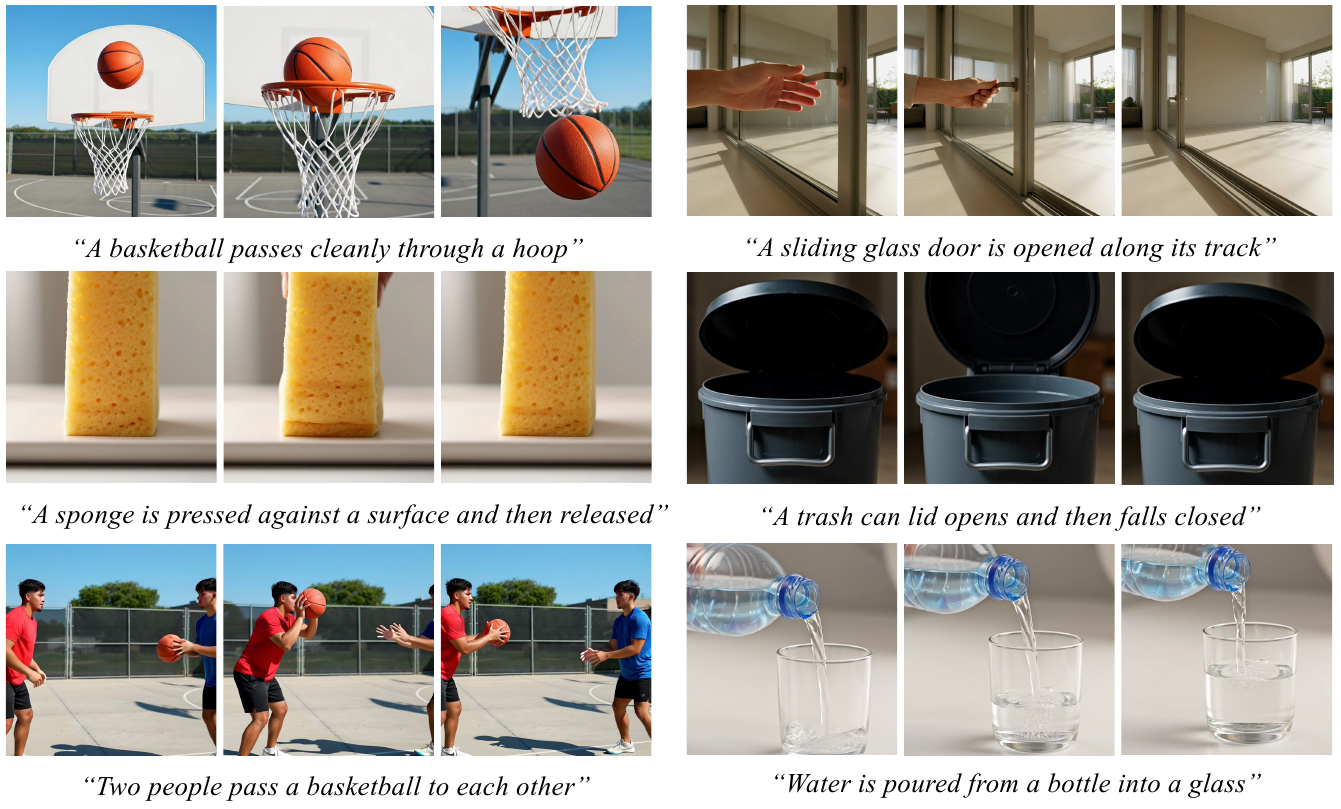}

\caption{\textbf{Representative EVD samples.}
The six prompts cover target-directed motion, a constrained mechanism, deformation and recovery, gravity-mediated closure, a two-person handoff, and liquid transfer. The examples are meant to show the local effect of the event gate: motion is concentrated near the active interaction and reduced once the intended state change has happened.}
\label{fig:qualitative}
\end{figure*}

Fig.~\ref{fig:qualitative} shows six EVD samples: ball-through-hoop, sliding door, sponge press/release, trash-can lid open/close, two-person pass, and liquid pouring.\footnote{These are exactly the six prompts shown in Fig.~\ref{fig:qualitative}.} We use these examples to check the concrete behavior of the method. The motion starts after the triggering action, contact regions stay spatially plausible, and the scene usually settles instead of continuing to drift after the event.

\subsection{Quantitative Results on EVD-Bench}
\label{sec:exp_quant}

Table~\ref{tab:evd_main_results} combines the human 2AFC study with VBench scores on EVD-Bench. Human raters prefer DiT+EVD over the corresponding DiT baseline for \emph{Text Faithfulness}, \emph{Overall Quality}, and most strongly \emph{Dynamics}. The automatic metrics show the same pattern we want to see: VBench \emph{Dynamics} increases, while \emph{Appearance} stays close to the base model. This matters because many failures in Fig.~\ref{fig:failures} are not low-level visual artifacts; they are incorrect or unstable state updates inside otherwise realistic-looking clips.

\begin{table}[t!]
\centering
\caption{\textbf{Comparison of EVD with video generation baselines on EVD-Bench.}
Human evaluation reports the percentage of pairwise votes favoring EVD; automatic metrics are computed using VBench.
\emph{TF} denotes Text Faithfulness, \emph{Qual.} denotes Overall Quality, \emph{Dyn.} denotes Dynamics, and \emph{App.} denotes Appearance; higher is better for all columns.}
\label{tab:evd_main_results}

\scriptsize
\setlength{\tabcolsep}{2.5pt}
\renewcommand{\arraystretch}{0.92}

\begin{minipage}[t]{0.485\linewidth}
\centering
\textbf{(a) Prior video generation baselines}
\resizebox{\linewidth}{!}{%
\begin{tabular}{@{}lccccc@{}}
\toprule
& \multicolumn{3}{c}{\textbf{Human Eval}} & \multicolumn{2}{c}{\textbf{Auto. Metrics}} \\
\cmidrule(lr){2-4}\cmidrule(lr){5-6}
\textbf{Method} & \textbf{TF} & \textbf{Qual.} & \textbf{Dyn.} & \textbf{App.} & \textbf{Dyn.} \\
\midrule
CogVideo2B      & 80.2 & 88.1 & 89.7 & 69.8 & 87.6 \\
CogVideo5B      & 65.4 & 73.8 & 72.5 & 72.3 & 89.2 \\
PyramidFlow     & 75.8 & 82.4 & 81.1 & 73.9 & 88.5 \\
DiT-4B          & 70.6 & 76.8 & 80.3 & 75.4 & 78.9 \\
\textbf{DiT-4B+EVD} & \textbf{88.9} & \textbf{91.3} & \textbf{96.4} & \textbf{76.2} & \textbf{94.8} \\
\bottomrule
\end{tabular}}
\end{minipage}
\hfill
\begin{minipage}[t]{0.485\linewidth}
\centering
\textbf{(b) Large-scale video generators}
\resizebox{\linewidth}{!}{%
\begin{tabular}{@{}lccccc@{}}
\toprule
& \multicolumn{3}{c}{\textbf{Human Eval}} & \multicolumn{2}{c}{\textbf{Auto. Metrics}} \\
\cmidrule(lr){2-4}\cmidrule(lr){5-6}
\textbf{Method} & \textbf{TF} & \textbf{Qual.} & \textbf{Dyn.} & \textbf{App.} & \textbf{Dyn.} \\
\midrule
Kling 3.0 Pro   & 61.8 & 67.4 & 72.9 & 78.6 & 92.6 \\
Runway Gen-4.5  & 64.7 & 71.2 & 76.8 & 76.9 & 91.4 \\
Veo 3.1         & 66.9 & 73.5 & 78.4 & 77.2 & 91.8 \\
Sora 2 Pro      & 63.5 & 69.8 & 74.1 & 77.8 & 90.9 \\
Mochi 1         & 58.2 & 63.7 & 70.3 & 72.6 & 88.8 \\
DiT-30B         & 72.4 & 76.1 & 79.5 & 73.8 & 88.7 \\
\textbf{DiT-30B+EVD} & \textbf{89.7} & \textbf{92.4} & \textbf{97.1} & \textbf{78.1} & \textbf{95.7} \\
\bottomrule
\end{tabular}}
\end{minipage}
\end{table}

\paragraph{Stress tests and diagnostics.}
Beyond EVD-Bench, we evaluate fixed-seed compositional/temporal and simultaneous-event subsets drawn from T2V-CompBench~\cite{sun2025t2vcompbench} and NeuS-V~\cite{sharan2025neusv}, compare against recent open-source generators (Wan/Hunyuan)~\cite{wan2025wanopenadvancedlargescale,kong2025hunyuanvideosystematicframeworklarge}, and report confidence intervals and motion-mask controls in Appendix~\ref{app:additional_checks}. Fig.~\ref{fig:pseudotarget_diagnostics} further validates that EVD activity localizes to prompt-relevant interaction regions rather than diffuse background motion: the pseudo-target, learned activity, and final gate concentrate around placement or contact and material-transfer events while suppressing inactive regions. This supports the central claim that EVD learns an event-grounded update signal, not merely a generic motion mask.

\begin{figure}[ht!]
\centering
\includegraphics[width=0.98\textwidth]{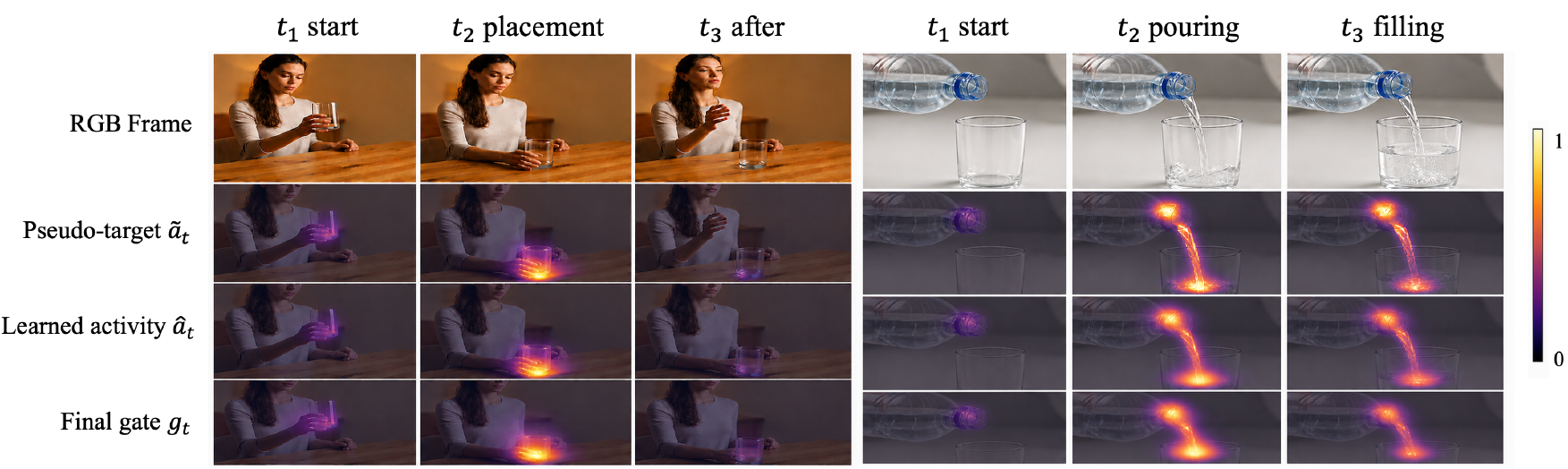}
\vspace{-8pt}
\caption{\textbf{Pseudo-target and gate localization diagnostics.}
We visualize placement/contact (left) and pouring/material transfer (right) at three time points. The first row shows RGB frames; lower rows overlay event signals on the corresponding frames. The pseudo-target \(\tilde{a}_t\) is the self-supervised event target extracted from localized change, the learned activity \(\hat{a}_t\) is the event-head prediction, and the final gate \(g_t\) is the smoothed, hysteretic, scheduled gate used to modulate latent updates. The three signals are aligned but not identical because they correspond to different stages of the EVD pipeline: supervision, prediction, and update gating. Activity concentrates at the glass--table contact and along the bottle-mouth, stream, and receiving-glass transfer path, while inactive/background regions remain suppressed.}
\label{fig:pseudotarget_diagnostics}
\vspace{-10pt}
\end{figure}

\subsection{Baseline Comparisons}
\begin{figure}[H]
\centering
\includegraphics[width=\textwidth]{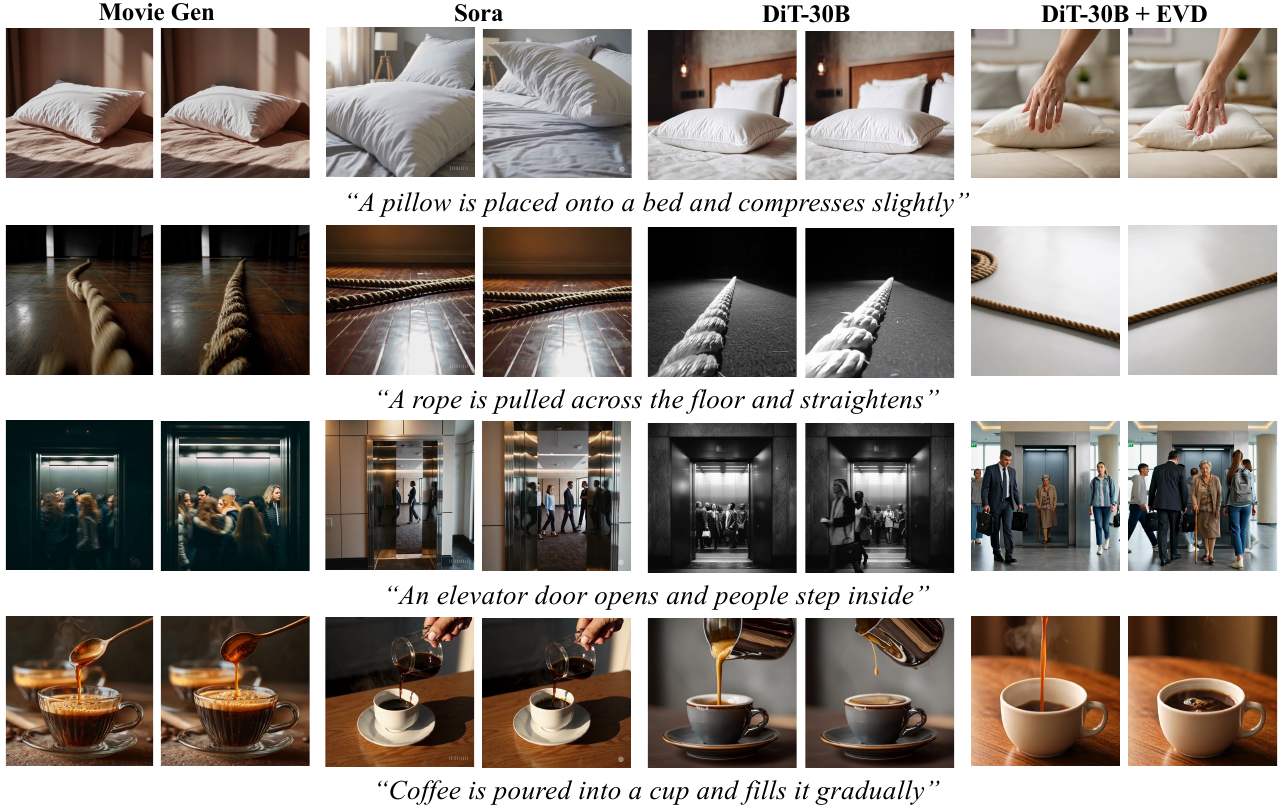}
\caption{\textbf{Qualitative comparison with video generation baselines.}
We compare EVD against Movie Gen, Sora, and DiT-30B on representative prompts involving soft-body deformation, flexible-object dynamics, structured scene interactions, and liquid transfer.
The comparison highlights where the event gate changes the result: the baseline clips often miss the contact response or keep changing state after the intended interaction, while EVD better localizes the state change around the active event.}
\label{fig:comparisons}
\end{figure}

Fig.~\ref{fig:comparisons} compares four interaction prompts. The baselines often look realistic in isolated frames, but the event itself is misplaced: an object begins changing before contact, the deformation or constraint response is too weak, or the scene keeps drifting after the interaction should have ended. We see this for material transfer (coffee filling), compliance (pillow compression), constraint enforcement (rope straightening), and multi-agent transitions (elevator door opening and people stepping inside). EVD does not solve every detail, but in these examples the main state change is better aligned with the trigger and has a more stable postcondition.

\subsection{Ablations and Sensitivity}
\label{sec:exp_ablate}

\paragraph{Component ablations.}
Table~\ref{tab:evd_ablations} breaks down the EVD components on EVD-Bench. Removing \emph{realization} mostly brings back non-causal initiation (\emph{Contact Stability}); removing \emph{consistency} makes interactions noisier and less stable (\emph{State Persistence}); and removing \emph{ordering} weakens the settling behavior after an event. Training-only EVD and inference-only EVD recover only part of the gain. The controls show that motion masking alone is not enough; the useful part is training the model to align its own event signal with the latent update. We also find that constant gating without annealing hurts preference, which is why the full model uses the scheduled gate. Details of the motion-mask audit are in Appendix~\ref{app:motionmask_audit}.
\vspace{-5px}
\paragraph{Efficiency and overhead.}
EVD keeps the backbone, solver, decoder, and sampling budget (\(K\)/NFE) fixed; the extra work is only the event head and direction-field gate.
Table~\ref{tab:evd_efficiency} shows that this adds negligible parameters and about \(1.02\times\) inference overhead.

\begin{table}[t!]
\caption{\textbf{Efficiency and overhead.}
EVD adds a lightweight event head and gating logic while keeping the same sampler, steps (\(K{=}50\)), and CFG structure (2 DiT evaluations per step).}
\label{tab:evd_efficiency}
\centering
\setlength{\tabcolsep}{3.6pt}
\renewcommand{\arraystretch}{0.92}
\resizebox{\linewidth}{!}{%
\begin{tabular}{@{}lccccccc@{}}
\toprule
\textbf{Model} &
\makecell{\textbf{Total}\\\textbf{Params}} &
\makecell{\textbf{Added}\\\textbf{Params}} &
\makecell{\textbf{Added}\\\textbf{(\%)}} &
\makecell{\textbf{Train}\\\textbf{Throughput}} &
\makecell{\textbf{Inference}\\\textbf{Overhead}} &
\makecell{\textbf{DiT}\\\textbf{evals/step}} &
\makecell{\textbf{Steps}\\\textbf{(K)}} \\
\midrule
DiT-4B        & 4.0B  & --    & --     & 1.00$\times$ & 1.00$\times$ & 2 (CFG) & 50 \\
DiT-4B+EVD    & 4.0B  & 6.5M  & 0.16\% & 0.98$\times$ & 1.02$\times$ & 2 (CFG) & 50 \\
\midrule
DiT-30B       & 30.0B & --    & --     & 1.00$\times$ & 1.00$\times$ & 2 (CFG) & 50 \\
DiT-30B+EVD   & 30.0B & 12.0M & 0.04\% & 0.97$\times$ & 1.02$\times$ & 2 (CFG) & 50 \\
\bottomrule
\end{tabular}}

\end{table}

\vspace{-5px}
\paragraph{Hyperparameter sensitivity.}
EVD is broadly robust to \(K\) (NFE), CFG scale \(w_{\mathrm{cfg}}\), and gating hyperparameters \((\beta,\tau_{\mathrm{on}},\tau_{\mathrm{off}},t^\star)\); the full sensitivity sweep is provided in Appendix~\ref{app:ablations} (Table~\ref{tab:evd_sensitivity}).

\begin{table*}[t!]
\vspace{-6px}
\caption{\textbf{EVD ablations on EVD-Bench (DiT-4B backbone).}
\textbf{Human Eval} reports the \emph{percentage of 2AFC votes favoring the full DiT-4B+EVD model} over each ablated variant (higher is better for EVD).
\textbf{Auto. Metrics} are computed using VBench.
All variants use identical sampling settings (solver, NFE, CFG scale) and the same prompt set.}
\label{tab:evd_ablations}
\centering
\resizebox{\textwidth}{!}{%
\begin{tabular}{@{}lccccc ccc cc@{}}
\toprule
& \multicolumn{5}{c}{\textbf{Settings}} & \multicolumn{3}{c}{\textbf{Human Eval (EVD wins \%)}} & \multicolumn{2}{c}{\textbf{Auto. Metrics}} \\
\cmidrule(r){2-6}\cmidrule(r){7-9}\cmidrule(r){10-11}
\textbf{Variant} &
\small{Real} & \small{Cons} & \small{Order} & \small{Gate} & \small{Sched.} &
\small{Text Faith.} & \small{Quality} & \textbf{\small{Dynamics}} &
\small{Appearance} & \textbf{\small{Dynamics}} \\
\midrule
\small{DiT-4B (no EVD)}                 & No  & No  & No  & No   & --         & 70.6 & 76.8 & 80.3 & 75.4 & 78.9 \\
\small{w/o event realization}           & No  & Yes & Yes & Yes  & Anneal     & 65.2 & 69.4 & 78.8 & 75.9 & 91.2 \\
\small{w/o event consistency}           & Yes & No  & Yes & Yes  & Anneal     & 61.3 & 65.0 & 74.2 & 76.0 & 92.1 \\
\small{Training-only (no gating)}       & Yes & Yes & Yes & No   & Off        & 63.0 & 66.8 & 77.1 & 76.1 & 90.5 \\
\small{Inference-only (no event losses)}& No  & No  & No  & Ext. & Anneal     & 70.5 & 74.9 & 86.0 & 75.6 & 84.0 \\
\small{Disable gating \& event use at inference} & Yes& Yes& Yes & No   & --         & 67.8 & 71.6 & 82.4 & 75.5 & 86.7 \\
\small{No schedule (const.\ gate)}      & Yes & Yes & Yes & Yes  & Const. (1.0) & 55.7 & 58.9 & 60.5 & 75.7 & 94.1 \\
\small{No schedule (weak const.\ gate)} & Yes & Yes & Yes & Yes  & Const. (0.5) & 57.9 & 61.0 & 65.8 & 76.1 & 93.6 \\
\midrule
\textbf{DiT-4B + EVD (full)}            & Yes & Yes & Yes & Yes  & Anneal     & \textbf{88.9}   & \textbf{91.3}   & \textbf{96.4}   & \textbf{76.2} & \textbf{94.8} \\
\bottomrule
\end{tabular}}
\vspace{-6px}
\end{table*}

\vspace{-6px}
\section{Conclusion}
\label{sec:conclusion}

We introduced Event-Driven Video Generation (EVD), a small modification to pretrained video DiTs that ties latent updates to predicted event activity. The main empirical takeaway is that the model does not need a new solver or decoder to reduce several interaction errors: when the event head is trained together with the latent update and used during sampling, EVD improves human preference and automatic dynamics metrics on EVD-Bench while leaving appearance largely unchanged.

\paragraph{Limitations.}
EVD is least reliable when the event is hard to localize at the model's operating resolution: small or occluded contacts, cluttered multi-object interactions, thin fluid effects, or dominant camera motion. In those cases, both the pseudo-targets and the learned event head can become ambiguous, and the gate may either suppress useful motion or allow unrelated motion through. Future work should make the event signal more object- and contact-aware, and may also benefit from stronger motion disentanglement or higher-resolution latent/video backbones.

\section*{Acknowledgements}

The authors thank NSF and NCSA for computational support.
This paper used generative AI tools for language proof.

%% file: sections/appendix.tex
\clearpage

\section{Appendix}
\label{appendix}

\etocarticlestyle
\etocsettocstyle{\subsection*{Table of Contents}}{}

\etocsetstyle{section}
  {\par\noindent}
  {\par\noindent}
  {\etocname\dotfill\etocpage\par}
  {}

\etocsetstyle{subsection}
  {\par\noindent\hspace{0em}}
  {\par\noindent\hspace{0em}}
  {\makebox[3.2em][l]{\etocnumber}\etocname\dotfill\etocpage\par}
  {}

\etocsetstyle{subsubsection}
  {\par\noindent\hspace{1.5em}}
  {\par\noindent\hspace{1em}}
  {\makebox[4.2em][l]{\etocnumber}\etocname\dotfill\etocpage\par}
  {}

\begingroup
\providecommand\authcount[1]{}

\etocsetlevel{title}{6}
\etocsetlevel{author}{6}
\etocsetlevel{institute}{6}
\etocsetlevel{email}{6}
\etocsetstyle{title}{}{}{}{}
\etocsetstyle{author}{}{}{}{}
\etocsetstyle{institute}{}{}{}{}
\etocsetstyle{email}{}{}{}{}
\etocsetnexttocdepth{subsubsection}
\localtableofcontents
\endgroup

\subsection{Abbreviations and symbols}
\label{app:glossary}

\paragraph{Abbreviations.}
\begin{itemize}
    \item \textbf{EVD:} Event-Driven Video Generation.
    \item \textbf{DiT:} Diffusion Transformer (video DiT backbone used as the base model).
    \item \textbf{CFG:} Classifier-Free Guidance.
    \item \textbf{FM:} Flow Matching.
    \item \textbf{ODE:} Ordinary Differential Equation (sampling view for rectified-flow / FM samplers).
    \item \textbf{NFE:} Number of Function Evaluations (sampling compute proxy).
    \item \textbf{TAE:} Temporal Autoencoder (video encoder/decoder used to map \(x \leftrightarrow z\)).
    \item \textbf{2AFC:} Two-Alternative Forced Choice (human preference protocol).
\end{itemize}

\paragraph{Core variables.}
\begin{itemize}
    \item \(x \in \mathbb{R}^{T \times H \times W \times 3}\): video clip in pixel space.
    \item \(z = E(x) \in \mathbb{R}^{T' \times H' \times W' \times C}\): latent video produced by the temporal autoencoder encoder \(E\).
    \item \(D(\cdot)\): temporal autoencoder decoder mapping latents back to pixels.
    \item \(t \in [0,1]\): continuous diffusion/flow time; \({\{t_k\}}_{k=0}^{K}\) is the discretized sampling grid.
    \item \(z_0 \sim \mathcal{N}(0,I)\): Gaussian noise latent; \(z_1\): clean latent.
    \item \(z_t\): noised/interpolated latent at time \(t\) (see Sec.~\ref{app:backbone}).
    \item \(y\): text prompt; \(\varnothing\): null prompt used for CFG.
\end{itemize}

\paragraph{Backbone and tokenization.}
\begin{itemize}
    \item \(u_\theta(z_t,y,t)\): DiT backbone prediction of the base target at time \(t\) (parameters \(\theta\)).
    \item \(\mathbf{s}_t \in \mathbb{R}^{N\times d}\): patchified token sequence; \(N\) tokens, width \(d\).
    \item \(\mathrm{Tok}(\cdot)\), \(\mathrm{UnTok}(\cdot)\): patchify/unpatchify operators aligned with DiT tokenization.
\end{itemize}

\paragraph{Flow Matching and guidance.}
\begin{itemize}
    \item \(\tau(z_1,z_0,t)\): base training target; in FM, this is the velocity \(v_t\).
    \item \(v_t = z_1 - z_0\): FM velocity target under linear interpolation.
    \item \(w_{\mathrm{cfg}}\): CFG scale.
    \item \(\widehat{\tau}^{\mathrm{cfg}}\): CFG-combined backbone prediction.
\end{itemize}

\paragraph{Events and gating.}
\begin{itemize}
    \item \(e_t \in \mathbb{R}^{N\times C_e}\): token-aligned event field with \(C_e\) channels.
    \item \(\hat{e}_t\): predicted event field; \(\hat{a}_t=\sigma(\hat{e}_t^{(1)})\in{[0,1]}^N\) is the activity channel.
    \item \(g_\psi(\cdot)\): event gate derived from event predictions (parameters \(\psi\)).
    \item \(g_k\): gate on sampling step \(k\); \(\tau_{\mathrm{on}},\tau_{\mathrm{off}}\): hysteresis thresholds; \(\beta\): gate sharpness.
    \item \(\rho(t)\): inference annealing schedule; \(t^\star\): cutoff controlling ``strong-early'' gating.
\end{itemize}

\paragraph{Losses.}
\begin{itemize}
    \item \(\mathcal{L}_{\mathrm{base}}\): backbone base loss (FM regression).
    \item \(\mathcal{L}_{\mathrm{real}}\): event realization loss (no event \(\Rightarrow\) no update).
    \item \(\mathcal{L}_{\mathrm{cons}}\): event consistency loss (event \(\Rightarrow\) coherent update).
    \item \(\mathcal{L}_{\mathrm{order}}\): ordering/termination loss (suppresses pre-event motion and post-event drift).
    \item \(\lambda_{\mathrm{real}},\lambda_{\mathrm{cons}},\lambda_{\mathrm{order}}\): corresponding loss weights.
    \item \(w(t)\): time-weighting function emphasizing early timesteps.
\end{itemize}

\subsection{Backbone and Notation}
\label{app:backbone}

\subsubsection{Latent video representation and notation}
\label{app:latent_notation}

We consider text-conditioned video generation where a clip is represented as a tensor
\(x \in \mathbb{R}^{T \times H \times W \times 3}\) with \(T\) frames. Following standard practice in large-scale video diffusion/flow models,
we operate in a compressed latent space using a temporal video autoencoder (e.g., VAE/TAE).
Let \(E(\cdot)\) and \(D(\cdot)\) denote the encoder and decoder, respectively, and define the latent video
\begin{equation}
    z = E(x), \qquad x \approx D(z),
\end{equation}
where \(z \in \mathbb{R}^{T' \times H' \times W' \times C}\) has reduced spatial/temporal resolution.

We write the text prompt as \(y\), encoded by a frozen text encoder into a sequence of text embeddings.
The generative backbone is a Diffusion Transformer (DiT) operating on latent videos; we denote the network by
\(\,u_\theta(\cdot)\), parameterized by \(\theta\).

\paragraph{Noising and training target.}
EVD is compatible with common continuous-time formulations used in modern DiT video models.
We adopt a generic continuous-time notation: a clean latent \(z_1\) is interpolated with noise \(z_0 \sim \mathcal{N}(0,I)\) to obtain
a noised latent \(z_t\) at time \(t \in [0,1]\),
\begin{equation}
    z_t = \alpha(t)\, z_1 + \sigma(t)\, z_0,
\end{equation}
for a chosen schedule \(\alpha,\sigma\).
The corresponding training target depends on the base model (diffusion-score, velocity/flow-matching, or rectified flow);
we denote it abstractly as \(\tau(z_1,z_0,t)\).
The DiT backbone is trained to predict \(\tau\) from \((z_t, y, t)\),
\begin{equation}
    \mathcal{L}_{\text{base}}(\theta)
    = \mathbb{E}_{z_1,z_0,t,y}\Big[\big\|u_\theta(z_t, y, t) - \tau(z_1,z_0,t)\big\|_2^2\Big].
\end{equation}
All EVD components introduced in later subsections are built on top of this backbone objective and sampling procedure.

\subsubsection{DiT-30B backbone summary}
\label{app:dit_backbone}

EVD is built on a large video Diffusion Transformer (DiT) operating in latent space.
Given a noised latent video \(z_t \in \mathbb{R}^{T' \times H' \times W' \times C}\), the backbone maps \((z_t,y,t)\) to a prediction
\(u_\theta(z_t,y,t)\) of the base training target (Sec.~\ref{app:latent_notation}).

\paragraph{Patchification and token embeddings.}
We partition \(z_t\) into non-overlapping spatiotemporal patches of size
\(p_t \times p_h \times p_w\), flatten each patch, and linearly project it into the model width \(d\),
yielding a token sequence
\begin{equation}
    \mathbf{s}_t = \mathrm{PatchEmbed}(z_t) \in \mathbb{R}^{N \times d},
\end{equation}
where \(N = (T'/p_t)(H'/p_h)(W'/p_w)\).
We add a learned spatiotemporal positional encoding \(\mathrm{PE} \in \mathbb{R}^{N \times d}\) and a time embedding
\(\gamma(t) \in \mathbb{R}^{d}\) (broadcast to all tokens) to obtain the transformer input.

\paragraph{Spatiotemporal transformer blocks.}
The backbone consists of \(L\) blocks of multi-head self-attention and MLP layers applied to \(\mathbf{s}_t\),
with residual connections and normalization. We write the block update abstractly as
\begin{equation}
    \mathbf{s}_t^{(\ell+1)} = \mathrm{Block}^{(\ell)}\!\left(\mathbf{s}_t^{(\ell)},\, \gamma(t),\, \phi(y)\right),
    \qquad \ell = 0,\dots,L-1,
\end{equation}
where \(\phi(y)\) denotes the text conditioning (described below).
The self-attention mixes information across both space and time by attending over the full spatiotemporal token set.

\paragraph{Text conditioning.}
Text \(y\) is encoded into a sequence of text embeddings \(\phi(y) \in \mathbb{R}^{M \times d_y}\) using a frozen text encoder.
The DiT conditions on \(\phi(y)\) via cross-attention (or equivalently, attention over a concatenated key/value memory),
so that each video token can attend to the prompt representation while preserving spatiotemporal structure.

\paragraph{Unpatchification and output projection.}
After the final block, tokens are linearly projected back to patch space and unpatchified to recover a latent-shaped tensor
\(\widehat{u} \in \mathbb{R}^{T' \times H' \times W' \times C}\). This \(\widehat{u}\) is interpreted according to the base objective
(e.g., velocity/flow target, score, or noise prediction) and is used for training and sampling.

\paragraph{Sampling interface (for later EVD modifications).}
At inference, the backbone is evaluated repeatedly along a discretized time grid \(\{t_k\}\), and an ODE/SDE solver (or discrete update)
uses \(u_\theta(z_{t_k},y,t_k)\) to update the latent trajectory \(\{z_{t_k}\}\).
EVD will modify \emph{what} the backbone is trained to represent (event grounding) and \emph{how} its predictions are used during sampling
(event-driven updates), while keeping the DiT backbone unchanged.

\subsubsection{Base training objective (Flow Matching)}
\label{app:base_objective}

In our implementation, the DiT-30B backbone is trained with a continuous-time Flow Matching objective in latent space.
Let \(z_1 = E(x)\) denote the clean latent video and \(z_0 \sim \mathcal{N}(0,I)\) denote Gaussian noise.
For a timestep \(t \sim \mathcal{U}[0,1]\), we form the interpolated latent
\begin{equation}
    z_t = t\, z_1 + (1-t)\, z_0.
\end{equation}
The corresponding velocity target is the time derivative of the interpolation,
\begin{equation}
    v_t \;=\; \frac{d z_t}{dt} \;=\; z_1 - z_0,
\end{equation}
which is constant with respect to \(t\) under the linear interpolation above.
The DiT backbone \(u_\theta(\cdot)\) is optimized to predict \(v_t\) from \((z_t, y, t)\) using an \(\ell_2\) regression loss:
\begin{equation}
\label{eq:fm_loss}
    \mathcal{L}_{\mathrm{FM}}(\theta)
    = \mathbb{E}_{z_1,\,z_0,\,t,\,y}\Big[\big\|u_\theta(z_t, y, t) - v_t\big\|_2^2\Big].
\end{equation}

\paragraph{Compatibility.}
While we describe EVD using Flow Matching notation for concreteness, the EVD components introduced in the sequel
(event representation, event-grounded losses, and event-driven sampling) apply to other parameterizations
(e.g., noise prediction or score prediction) by replacing the target in~\eqref{eq:fm_loss} with the corresponding base objective.

\subsubsection{Sampling, guidance, and evaluation conventions}
\label{app:sampling_conventions}

We summarize the sampling interface of the DiT-30B backbone and the conventions we use for guidance and evaluation.
Unless otherwise stated, all models generate fixed-length clips (128 frames at 24\,fps) in the latent space of a temporal autoencoder (TAE),
followed by decoding to pixel space~\cite{pmlr-v267-chefer25a}. The TAE design follows the Movie~Gen temporal autoencoder specification~\cite{polyak2024movie}.

\paragraph{Time discretization and number of function evaluations (NFE).}
Sampling proceeds by evolving a latent trajectory \({\{z_{t_k}\}}_{k=0}^{K}\) along a monotone time grid
\(0 = t_0 < t_1 < \cdots < t_K = 1\). Each step queries the backbone once (or more, depending on guidance batching),
so the total NFE scales with \(K\) times the number of model evaluations per step. In practice, we report results at fixed \(K\)
to ensure fair comparisons across methods.

\paragraph{Classifier-free guidance (CFG).}
We follow standard CFG conventions: at each step \(t_k\), we evaluate the model with the prompt \(y\) and with a null prompt \(\varnothing\),
and combine the predictions with a guidance scale \(w \ge 0\):
\begin{equation}
\label{eq:cfg}
u_{\mathrm{cfg}}(z_{t_k}, y, t_k)
= (1+w)\,u_\theta(z_{t_k}, y, t_k) \;-\; w\,u_\theta(z_{t_k}, \varnothing, t_k).
\end{equation}
We use the same \(w\) across all compared methods unless noted.

\paragraph{Batching for multi-condition guidance.}
When multiple conditioning signals are used at inference (e.g., conditional/unconditional CFG and additional auxiliary conditions),
the corresponding model evaluations can be executed as a single batched forward pass for efficiency.
This follows standard CFG-style implementations and their multi-condition extensions (e.g., composable guidance and IP2P-style formulations)~\cite{ho2021classifierfree,10.1007/978-3-031-19790-1_26,10204579,pmlr-v267-chefer25a}.

\paragraph{Guidance scheduling over timesteps.}
For guidance signals that primarily shape coarse spatiotemporal structure, concentrating guidance in early denoising steps is often beneficial,
since these steps largely determine global dynamics.
For example, VideoJAM applies its motion guidance only during the first half of generation (50 steps), motivated by the observation that coarse motion is set early~\cite{pmlr-v267-chefer25a}.
EVD adopts the same principle: event-focused guidance is applied strongly in early steps and annealed thereafter (Sec.~\ref{app:inf_schedule}),
while text CFG is applied throughout~\cite{ho2021classifierfree}.

\paragraph{Evaluation protocol.}
Unless explicitly stated, we generate one sample per prompt per model under identical sampling settings and a fixed random seed, and report both
automatic scores (VBench) and human preference results; we use the first obtained sample for each prompt (no cherry-picking)~\cite{pmlr-v267-chefer25a,Huang_2024_CVPR,maduabuchi2026corruptionawaretraininglatentvideo}.
For human evaluation, we follow a standard two-alternative forced-choice (2AFC) setup in which raters compare our output against a baseline
and select the better video along text faithfulness, overall quality, and dynamics~\cite{9878449,Blattmann2023StableVD}.

\subsection{EVD}
\label{app:evd_onepage}

\subsubsection{Core claim: event-driven state transitions}
\label{app:evd_core_claim}

Modern video generators often produce locally smooth frame-to-frame motion while violating basic causal structure:
objects may move without contact, effects may precede causes, and post-interaction states may drift.
EVD addresses this by enforcing a simple modeling principle:

\begin{quote}
\emph{A video is generated as a sequence of event-driven state transitions: persistent state evolves only when an event occurs,
and events must be realized as consistent state changes.}
\end{quote}

\paragraph{State and event variables.}
Let \(z_t \in \mathbb{R}^{T' \times H' \times W' \times C}\) denote the latent video state at diffusion/flow time \(t\),
and let \(e_t\) denote an \emph{event representation} aligned to the same time (Sec.~\ref{app:event_def}).
Intuitively, \(z_t\) carries appearance and scene configuration, while \(e_t\) encodes the presence and phase of an interaction
(e.g., initiation, continuation, termination) that should drive changes in \(z_t\).

\paragraph{Event grounding principle.}
EVD implements two coupled constraints during learning and sampling:
\begin{align}
\textbf{(i) No-event $\Rightarrow$ no-update:}\quad & e_t \approx \mathbf{0} \;\Longrightarrow\; \Delta z_t \approx \mathbf{0}, \label{eq:no_event_no_update}\\
\textbf{(ii) Event $\Rightarrow$ realized update:}\quad & e_t \not\approx \mathbf{0} \;\Longrightarrow\; \Delta z_t \;\text{reflects the event semantics.}\label{eq:event_realized_update}
\end{align}
We use these two constraints for the failure cases that show up most often in frame-first generation:
\emph{missing events}, where the state changes without a visible initiating interaction, and \emph{ghost events}, where an apparent action has no coherent outcome.

\paragraph{Event-driven update rule (high-level).}
Let \(u_\theta(z_t,y,t)\) denote the backbone prediction (Sec.~\ref{app:backbone}).
EVD uses a gated update interface of the form
\begin{equation}
\label{eq:evd_update_highlevel}
\Delta z_t \;=\; g_\psi(e_t,t)\,\odot\,u_\theta(z_t,y,t),
\end{equation}
where \(g_\psi(\cdot)\in[0,1]\) is an event gate (scalar, per-token, or per-patch) derived from the event representation,
and \(\odot\) denotes elementwise modulation.
When \(e_t\) says that no interaction is active, the gate reduces the update in that region. When an event is active,
the same interface lets the backbone update pass through for the corresponding state change.
The exact forms of \(e_t\) and \(g_\psi\) are specified in Secs.~\ref{app:event_rep}--\ref{app:event_confidence} and
Secs.~\ref{app:inference}--\ref{app:inf_schedule}.
\paragraph{Connection to observed failure modes.}
The constraints in~\eqref{eq:no_event_no_update}--\eqref{eq:event_realized_update} give a common explanation for the four failure categories used in our analysis:
\emph{State Persistence} (termination without drift), \emph{Spatial Accuracy} (event outcome aligns with target),
\emph{Support Relations} (valid load-bearing configurations), and \emph{Contact Stability} (cause precedes motion and settling is stable).
We use the same mechanism for all four cases: identify where the interaction is active, gate the latent update there, and reduce updates elsewhere.

\subsubsection{What changes vs.\ vanilla DiT}
\label{app:evd_whatchanges}

The DiT-30B tokenization, attention blocks, and text conditioning are left in place. EVD changes the
\emph{representation}, \emph{objective}, and \emph{sampling interface} around that backbone.

\paragraph{(1) Add an event representation.}
In addition to the latent video state \(z_t\), EVD introduces an event variable \(e_t\) aligned with the same diffusion/flow time.
Depending on the variant, \(e_t\) can be implemented as (i) a dense event map in latent patch space or (ii) a compact set of event tokens.
The role of \(e_t\) is to explicitly encode whether an interaction is active and its phase (initiation/progression/termination).

\paragraph{(2) Train with event-grounded losses.}
Vanilla DiT minimizes the base objective \(\mathcal{L}_{\mathrm{FM}}\) (Sec.~\ref{app:base_objective}).
EVD augments this with two lightweight constraints:
(i) \emph{event realization} to discourage state changes when no event is present, and
(ii) \emph{event consistency} to ensure that predicted events correspond to coherent state evolution.
These terms are meant to reduce both ``missing events'' and ``ghost events'' (Sec.~\ref{app:evd_core_claim}).

\paragraph{(3) Use event-driven sampling.}
At inference, vanilla DiT applies text guidance (CFG) and updates \(z_t\) using the backbone prediction.
EVD modifies the update using an event gate \(g_\psi(e_t,t)\) (Eq.~\eqref{eq:evd_update_highlevel}) so that
\emph{event confidence controls when and where the latent state is allowed to change}.
This is the intended source of the behavior we measure later: less post-interaction drift, fewer pre-contact updates,
and more reliable completion of placement or alignment events.

\paragraph{Net effect.}
EVD is not just a smoother. It changes which latent updates are allowed to occur under a predicted event signal.
For that reason, the largest changes should appear in dynamics-related metrics and human judgments of interaction quality,
while appearance scores should move much less.

\subsubsection{Mapping EVD components to failure modes}
\label{app:evd_mapping}

We use four failure categories throughout the paper---\emph{State Persistence}, \emph{Spatial Accuracy}, \emph{Support Relations}, and
\emph{Contact Stability}---and design EVD so that each category is addressed by an explicit mechanism rather than emergent smoothing.
Below we summarize the correspondence between the observed failures (Fig.~\ref{fig:failures}) and EVD's components
(Secs.~\ref{app:event_rep}--\ref{app:inference}).

\paragraph{State Persistence (terminate $\Rightarrow$ rest).}
Failure signature: objects continue drifting or jittering after an interaction ends (e.g., residual chair motion).
EVD targets this via \textbf{event termination} in the event representation \(e_t\) and the \textbf{no-event $\Rightarrow$ no-update} constraint
(Eq.~\eqref{eq:no_event_no_update}). Concretely, once \(e_t\) indicates the event has ended, the event gate suppresses further latent updates,
preventing post-interaction drift. The event-realization loss further discourages nonzero updates when \(e_t\) is near zero.

\paragraph{Spatial Accuracy (outcome aligns with target).}
Failure signature: misaligned outcomes (e.g., placement misses a platform, offsets accumulate).
EVD encourages \textbf{outcome-conditioned updates} by coupling \(e_t\) to state changes through gated modulation
(Eq.~\eqref{eq:evd_update_highlevel}). During training, the event-consistency loss penalizes event predictions that do not produce the
corresponding state change toward a stable, target-consistent configuration, sharpening alignment at event completion.

\paragraph{Support Relations (valid load-bearing configurations).}
Failure signature: stacked objects appear without a placing action, or settle into physically inconsistent configurations.
EVD addresses this in two ways: (i) \textbf{event realization} discourages ``teleportation'' of support relationships (stacking without a placement event);
(ii) \textbf{event consistency} ties the predicted event phase to a coherent progression of the state (approach, contact, and release),
encouraging stable postconditions rather than instantaneous, unsupported transitions.

\paragraph{Contact Stability (cause precedes motion; stable settling).}
Failure signature: motion begins before contact, contact is visually absent, or settling remains unstable (sliding/drifting).
EVD explicitly models \textbf{causal initiation} by requiring \(e_t\) to activate before allowing the corresponding update
(\(g_\psi(e_t,t)\) increases only when an initiating interaction is present). This reduces pre-contact motion.
In addition, the gate is annealed after event completion to promote stable settling, aligning with the ``contact then rest'' structure of many prompts.

\paragraph{Summary.}
The four categories differ visually, but the correction we apply is the same: learn an event representation \(e_t\), use it to decide which latent updates are
permitted, and penalize mismatches between the event signal and the resulting state evolution.

\subsection{Event Representation and Supervision}
\label{app:event_rep}

\subsubsection{What is an ``event'' in EVD?}
\label{app:event_def}

EVD models a video as \emph{persistent latent state} punctuated by \emph{events} that induce structured state changes.
Informally, an event is the minimal interaction that explains a meaningful transition in the scene---e.g., \emph{contact is made},
\emph{an object is released}, \emph{a constraint becomes active} (hinge/track), or \emph{material is transferred} (pouring).

\paragraph{Event as a typed, phased interaction.}
We represent events with two ingredients:
(i) an \emph{activity} signal indicating whether an interaction is currently active, and
(ii) a \emph{phase} signal indicating where the interaction lies along an initiation-to-termination progression.
Concretely, for each diffusion/flow time \(t\), we define an event representation
\begin{equation}
    e_t \;\triangleq\; \big(a_t,\; p_t,\; \kappa_t\big),
\end{equation}
where \(a_t \in [0,1]\) is an event activity score, \(p_t \in [0,1]\) is an event phase (from early to late),
and \(\kappa_t\) encodes event type (e.g., contact/transfer/deformation/constraint) when multiple interaction modes may occur.
In the simplest variant, \(\kappa_t\) is omitted and \(e_t=(a_t,p_t)\).

\paragraph{Spatial localization (where the event acts).}
Because events are typically localized (e.g., hand--object contact, placement region, hinge boundary), EVD attaches event variables
to the same spatiotemporal tokenization used by the DiT backbone.
Let \(\mathrm{PatchEmbed}(z_t)\) produce \(N\) tokens (Sec.~\ref{app:dit_backbone}). We define a token-aligned event field
\begin{equation}
    e_t \in \mathbb{R}^{N \times C_e},
\end{equation}
where \(C_e\) is small (e.g., \(C_e \in \{1,2,4\}\)).
The activity component \(a_t\) can be interpreted as a per-token gate, while \(p_t\) captures local event progress.
This alignment allows EVD to modulate updates \emph{where} an interaction occurs, rather than globally smoothing the entire video.

\paragraph{Event-grounded state transitions.}
EVD uses \(e_t\) to constrain state evolution in latent space:
event activity determines whether updates are permitted, and event phase shapes when updates should begin and terminate.
This is the local form of the two failure modes discussed earlier:
(i) \emph{missing events} (state changes without an interaction) and
(ii) \emph{ghost events} (an apparent interaction with no coherent state change).

\subsubsection{Event signals used in EVD}
\label{app:event_signals}

EVD supports multiple instantiations of the event representation \(e_t\), trading off expressivity, overhead, and ease of integration with a
pretrained DiT-30B backbone. We describe three practical variants; in all cases, \(e_t\) is aligned with the DiT tokenization so it can
modulate latent updates at the appropriate spatiotemporal locations.

\paragraph{(i) Dense event field (per-token event map).}
The default EVD representation is a dense event field
\begin{equation}
    e_t \in \mathbb{R}^{N \times C_e},
\end{equation}
where \(N\) is the number of spatiotemporal tokens and \(C_e\) is small (typically 1--4).
A minimal choice is \(C_e=1\), where \(e_t\) encodes activity only.
A slightly richer choice is \(C_e=2\), where \(e_t = (a_t,p_t)\) encodes both event activity \(a_t\in[0,1]\) and phase \(p_t\in[0,1]\).
This variant is lightweight, fully local, and directly supports event gating \(g_\psi(e_t,t)\) (Sec.~\ref{app:inf_update}). 

\paragraph{(ii) Sparse event tokens (compact event memory).}
For interactions that are semantically global but spatially sparse (e.g., a single handoff or a single placement), EVD can instead represent
events as a small set of learned tokens
\begin{equation}
    e_t^{\mathrm{tok}} \in \mathbb{R}^{K_e \times d}, \qquad K_e \ll N,
\end{equation}
and inject them via cross-attention into the DiT backbone.
This yields a compact ``event memory'' that conditions the video tokens, while a separate projection produces a token-aligned activity gate
\(\tilde{a}_t \in {[0,1]}^N\) used for update modulation. In practice, this variant is useful when compute is tight and events are few.

\paragraph{(iii) Scalar event progress (global activity/phase).}
For ablations and the simplest deployments, EVD can use a global event descriptor
\begin{equation}
    e_t^{\mathrm{glob}} = (a_t^{\mathrm{glob}}, p_t^{\mathrm{glob}}) \in {[0,1]}^2,
\end{equation}
shared across all tokens. This captures ``whether something is happening'' and ``how far along it is'' but cannot localize interactions.
We include this variant mainly as a diagnostic baseline; it improves gross temporal ordering but is weaker on spatially localized contacts.

\paragraph{Which variant we use.}
Unless noted otherwise, our 30B results use the dense event field \(e_t \in \mathbb{R}^{N \times C_e}\) with a small channel budget.
This choice provides a direct interface for (a) suppressing spurious updates outside interaction regions and (b) enforcing event termination
to stabilize the post-interaction state, while adding negligible overhead relative to the DiT-30B backbone.

\subsubsection{How event targets are obtained (self-supervised extraction)}
\label{app:event_targets}

EVD does not require manual event annotations. Instead, we derive \emph{pseudo-targets} for event activity and phase directly from the
training videos using lightweight, off-the-shelf signals that capture \emph{when} and \emph{where} meaningful change occurs.

\paragraph{Inputs and alignment.}
Given a training clip \(x\) and its latent \(z_1 = E(x)\), we compute event pseudo-targets at the same spatiotemporal granularity as the DiT tokens.
In practice, we operate either (i) on decoded frames \(\tilde{x}=D(z_1)\) at the autoencoder resolution or (ii) directly on latent differences,
and then downsample/aggregate to the patch grid, yielding a token-aligned event field \(e_t^\star \in \mathbb{R}^{N\times C_e}\).

\paragraph{Activity target (``is an interaction happening''?).}
We estimate a dense motion/change magnitude signal and convert it into an activity map.
A simple and robust choice is to use optical flow magnitude between consecutive frames (or a latent-space proxy).
Let \(F_\tau\) denote an off-the-shelf flow estimator (e.g., RAFT is a common choice for large-scale pipelines) and let
\(f_\tau = F_\tau(\tilde{x}_\tau,\tilde{x}_{\tau+1})\).
Define the per-pixel magnitude \(m_\tau = \|f_\tau\|_2\).
We then map this to an activity probability via a soft threshold:
\begin{equation}
a_\tau^\star(u) = \sigma\!\left(\frac{m_\tau(u)-\mu}{s}\right),
\end{equation}
where \(\sigma(\cdot)\) is the logistic sigmoid, \(\mu\) is a robust scale (e.g., median magnitude), and \(s\) controls softness.
Finally, we aggregate \(a_\tau^\star\) to the DiT patch grid (average pooling over pixels and frames within each patch) to obtain
\(a_\tau^\star \in {[0,1]}^N\).

\paragraph{Phase target (``where are we within the interaction''?).}
For many prompts, a single dominant interaction admits a canonical progression from initiation to termination.
We construct a normalized phase signal from the cumulative activity:
\begin{equation}
p_\tau^\star = 
\frac{\sum_{j=0}^{\tau} \langle a_j^\star, \mathbf{1}\rangle}{\sum_{j=0}^{T-1} \langle a_j^\star, \mathbf{1}\rangle + \varepsilon},
\end{equation}
where \(\langle a_j^\star,\mathbf{1}\rangle\) sums activity over tokens and \(\varepsilon\) avoids division by zero.
This yields \(p_\tau^\star \in [0,1]\) that increases smoothly over the event and saturates afterward.
When multiple disjoint interactions exist, we compute phase locally per-token (by normalizing within spatial neighborhoods) to avoid
forcing unrelated regions to share a global phase.

\paragraph{Event-type cues.}
When we instantiate a multi-channel event type \(\kappa_t\) (Sec.~\ref{app:event_def}), we assign coarse types using simple diagnostics on the same signals:
e.g., large localized activity near boundaries suggests \emph{contact/impact}, sustained constrained motion suggests \emph{mechanism/track},
and spatially diffuse change suggests \emph{material transfer}. These cues remain self-supervised and are used only as weak targets.

\paragraph{Summary.}
The pseudo-targets \(e_t^\star=(a_t^\star,p_t^\star,\kappa_t^\star)\) provide a lightweight supervisory signal that teaches the model
\emph{when and where} changes should occur, without requiring any additional human labeling.

\subsubsection{Event confidence and uncertainty (and how it is used)}
\label{app:event_confidence}

In practice, event pseudo-targets are noisy: flow-based change can be triggered by camera motion, texture flicker, or small background dynamics.
EVD therefore associates each event estimate with a \emph{confidence} signal that controls how strongly event grounding is enforced during
training and sampling.

\paragraph{Confidence score.}
Given a token-aligned activity target \(a_t^\star \in {[0,1]}^N\) (Sec.~\ref{app:event_targets}), we define a scalar confidence
\(\,c_t^\star \in {[0,1]}\) as a robust summary of activity concentration:
\begin{equation}
\label{eq:event_confidence}
c_t^\star \;=\; \mathrm{clip}\!\Big(\mathrm{mean}_{i\in\mathcal{I}_t}\, a_t^\star(i),\, 0,\, 1\Big),
\qquad
\mathcal{I}_t \;=\; \{ i : a_t^\star(i) \ge \tau_a \},
\end{equation}
where \(\tau_a\) is a small threshold (0.3) and \(\mathcal{I}_t\) selects the most active tokens.
Intuitively, \(c_t^\star\) is high when the interaction is spatially localized and unambiguous, and low when the activity is diffuse or weak.

\paragraph{Training usage: loss weighting.}
We weight event-grounded auxiliary losses by confidence to avoid over-regularizing ambiguous regions:
\begin{equation}
\label{eq:event_weighted_losses}
\mathcal{L} \;=\; \mathcal{L}_{\mathrm{FM}}
\;+\; \lambda_{\mathrm{real}}\, c_t^\star\, \mathcal{L}_{\mathrm{real}}
\;+\; \lambda_{\mathrm{cons}}\, c_t^\star\, \mathcal{L}_{\mathrm{cons}},
\end{equation}
so that event grounding is emphasized when the extracted event signal is reliable.

\paragraph{Inference usage: adaptive gating and guidance.}
At sampling time, we similarly modulate the strength of event-driven updates by a confidence-weighted schedule:
\begin{equation}
\label{eq:confidence_guidance}
g_\psi(e_t,t) \;=\; \underbrace{\rho(t)}_{\text{early-step emphasis}}
\cdot \underbrace{\tilde{c}_t}_{\text{model/event confidence}}
\cdot \underbrace{\sigma\!\big(\beta\, a_t\big)}_{\text{activity gate}},
\end{equation}
where \(\rho(t)\) is a monotonically decreasing schedule (strong early, weak late),
\(\tilde{c}_t\) is the model's predicted confidence (trained to match \(c_t^\star\)),
and \(\beta\) controls gate sharpness.
This ensures that strong event gating is applied primarily when the model is confident an interaction is occurring, reducing the risk of
suppressing legitimate motion in challenging scenes.

\paragraph{Calibration.}
We calibrate \(\tilde{c}_t\) using a temperature parameter on a held-out set (or simple clipping),
so that confidence reflects the empirical reliability of event predictions and remains comparable across prompts.

\subsection{Architecture: Adding Events to DiT-30B}
\label{app:arch}

\subsubsection{Conditioning pathway: injecting events into the DiT backbone}
\label{app:arch_cond}

EVD leaves the DiT-30B transformer blocks intact and introduces an \emph{event pathway} that (i) produces a token-aligned event representation
\(e_t\) and (ii) injects it into the backbone as an additional conditioning signal.
We design the injection to be lightweight and to preserve the pretrained behavior at initialization via zero-impact conditioning
(e.g., zero-initialized projections / appended zero-rows), following the same stability principle used in DiT-based video adaptations~\cite{10377858,pmlr-v267-chefer25a} and in conditional-control networks for diffusion models~\cite{10377881}.
When a parameter-efficient variant is desired, the same event pathway can be implemented with low-rank adapters on attention/MLP projections~\cite{hu2022lora}.

\paragraph{Event module.}
Given the noised latent \(z_t\), prompt \(y\), and time \(t\), an event module \(h_\psi\) produces a token-aligned event field
\begin{equation}
    e_t \;=\; h_\psi(z_t, y, t)\in\mathbb{R}^{N\times C_e},
\end{equation}
where \(N\) is the number of spatiotemporal tokens induced by patchification and \(C_e\) is small (Sec.~\ref{app:event_rep}).

\paragraph{Input-level event injection (default).}
Let \(\mathbf{s}_t=\mathrm{PatchEmbed}(z_t)\in\mathbb{R}^{N\times d}\) be the DiT token sequence (Sec.~\ref{app:dit_backbone}).
We embed the event field into the same width \(d\) via a linear map \(W_e\in\mathbb{R}^{C_e\times d}\) and add it as a token-wise bias:
\begin{equation}
\label{eq:event_add}
    \mathbf{s}_t^{(0)} \;=\; \mathbf{s}_t \;+\; \eta(t)\,\big(e_t W_e\big),
\end{equation}
where \(\eta(t)\) is a (possibly scheduled) scalar controlling event-conditioning strength.
We initialize \(W_e\) to zero (or initialize \(\eta(t)=0\)) so that the model reduces exactly to the pretrained backbone at step 0,
and event conditioning is learned during fine-tuning.

\paragraph{Alternative: channel concatenation with zero-init projection.}
In an equivalent implementation, we concatenate a projected event tensor to the latent channels prior to patchification
(Sec.~\ref{app:dit_backbone}), \(\tilde{z}_t=[z_t,\;\Pi(e_t)]\), and extend the input projection with zero-initialized rows so that the
pretrained mapping is preserved at initialization~\cite{pmlr-v267-chefer25a,Zhang_2023_ICCV}.
This variant is convenient when the codebase already supports multi-channel latent inputs.

\paragraph{Mid-block modulation.}
For tighter control of where/when events affect computation, we use FiLM-style modulation inside each transformer block:
\begin{equation}
\label{eq:event_film}
    \mathrm{Norm}(\mathbf{s}) \;\mapsto\; \gamma_\ell(e_t,t)\odot \mathrm{Norm}(\mathbf{s}) \;+\; \beta_\ell(e_t,t),
\end{equation}
where \(\gamma_\ell,\beta_\ell\) are shallow MLPs applied token-wise to \([e_t;\gamma(t)]\).
We use this only in the 30B setting when the qualitative gains justify the additional parameters; otherwise the input-level injection
(Eq.~\eqref{eq:event_add}) suffices.

\paragraph{Text conditioning unchanged.}
Text embeddings \(\phi(y)\) are consumed by the DiT via the existing cross-attention pathway; EVD does not alter the text encoder or
the prompt-conditioning interface.

\paragraph{Summary.}
EVD adds an event pathway that produces \(e_t\) and injects it into the DiT token stream with zero-initialized conditioning,
ensuring stable fine-tuning of a pretrained 30B backbone while enabling event-aware computation.

\subsubsection{Output parameterization: state update and event prediction}
\label{app:arch_output}

EVD keeps the DiT backbone prediction interface for the \emph{state update} (the base training target), and adds a lightweight
\emph{event head} that predicts the event representation used for grounding and gating.

\paragraph{State prediction (unchanged).}
Let \(\widehat{\tau}_t = u_\theta(z_t,y,t)\) denote the DiT prediction of the base target \(\tau(z_1,z_0,t)\)
(e.g., the Flow Matching velocity \(v_t\); Sec.~\ref{app:base_objective}).
This prediction is used exactly as in the pretrained backbone during both training and sampling, except that EVD modulates its effect on
the latent update via an event gate (Sec.~\ref{app:inf_update}).

\paragraph{Event prediction.}
To obtain an explicit event variable aligned with the DiT tokenization, we attach a small projection head to the final DiT token features.
Let \(\mathbf{s}_t^{(L)} \in \mathbb{R}^{N\times d}\) be the final token sequence produced by the transformer.
We predict a token-aligned event field \(\hat{e}_t \in \mathbb{R}^{N\times C_e}\) using a linear layer (or a 2-layer MLP):
\begin{equation}
\label{eq:event_head}
    \hat{e}_t \;=\; \pi_\psi\!\left(\mathbf{s}_t^{(L)}, t\right),
    \qquad
    \pi_\psi(\mathbf{s},t) \;=\; \mathrm{MLP}_\psi\!\big([\mathbf{s};\gamma(t)]\big)\,,
\end{equation}
where \(\gamma(t)\) is the standard time embedding.
We typically interpret the first channel as an \emph{activity} logit and additional channels as phase/type descriptors
(Sec.~\ref{app:event_def}).

\paragraph{Joint prediction view.}
It is often convenient to view the model as producing a joint output
\begin{equation}
\label{eq:joint_output}
    u_{\theta,\psi}(z_t,y,t) \;=\; \big(\widehat{\tau}_t,\; \hat{e}_t\big),
\end{equation}
where \(\widehat{\tau}_t\) drives the state evolution and \(\hat{e}_t\) provides the event grounding signal used for both auxiliary losses
(Sec.~\ref{app:losses}) and event-driven sampling (Sec.~\ref{app:inference}).

\paragraph{Initialization and stability.}
To preserve the pretrained DiT behavior at the beginning of fine-tuning, we initialize the event head \(\pi_\psi\) to near-zero output
(e.g., small weights), so that the event pathway does not perturb the backbone prediction initially.
This follows a common ``no-op at initialization'' design used when attaching new conditioning/residual branches to large pretrained generators,
including zero-initialized control branches in diffusion models~\cite{Zhang_2023_ICCV}, lightweight DiT-video adaptations that preserve the
pretrained mapping at initialization~\cite{pmlr-v267-chefer25a}, and parameter-efficient adapter layers that are initialized to behave close to
the identity~\cite{pmlr-v97-houlsby19a}.

\paragraph{When to omit the event head.}
For ablations or strict minimalism, one may compute \(e_t\) purely from an external extractor on decoded frames (Sec.~\ref{app:event_targets}).
However, we find that predicting \(\hat{e}_t\) directly from the DiT features yields the strongest gains, since it lets the model learn
an event representation aligned with its own latent geometry and sampling trajectory.

\subsubsection{Parameter budget options (and what we use at 30B)}
\label{app:arch_budget}

We implement EVD under three adaptation budgets for large pretrained DiT backbones.
We summarize three practical configurations, ordered from minimal overhead to maximal flexibility.

\paragraph{EVD-lite (lowest overhead).}
This variant keeps the DiT backbone frozen (or lightly tuned) and adds only:
(i) the event injection parameters \(W_e\) (Eq.~\eqref{eq:event_add}),
(ii) the event head \(\pi_\psi\) (Eq.~\eqref{eq:event_head}), and
(iii) a small gating module \(g_\psi\) used at sampling time (Sec.~\ref{app:inference}).
The additional parameters are \(O(C_e d)\) for \(W_e\) and \(O(d C_e)\) for the event head (often a single linear layer),
which is negligible relative to a 30B backbone.
EVD-lite is useful for rapid iteration and ablations, and already yields visible improvements on event fidelity.

\paragraph{EVD-adapter (moderate overhead).}
Here we add lightweight adapters (e.g., LoRA or small bottleneck MLPs) to a subset of transformer blocks while keeping the base weights fixed.
Event injection and the event head remain as in EVD-lite.
This typically improves the alignment between the learned event representation and the backbone dynamics without the cost of full fine-tuning.
In our experience, adapting attention projections in later blocks provides most of the benefit.

\paragraph{EVD-full (highest performance).}
This variant fine-tunes the full DiT-30B weights jointly with the event modules.
Although this is the most expensive option, it produces the strongest and most reliable gains on EVD-Bench, particularly for:
(i) precise interaction outcomes (spatial accuracy),
(ii) stable post-contact settling (contact stability),
and (iii) ``event realization'' failures where the baseline skips the visible interaction.

\paragraph{What we report at 30B.}
Unless otherwise stated, our main 30B results use \textbf{EVD-full} with:
(a) dense event field \(e_t \in \mathbb{R}^{N\times C_e}\),
(b) input-level event injection (Eq.~\eqref{eq:event_add}),
(c) a lightweight event head (Eq.~\eqref{eq:event_head}),
and (d) event-driven sampling with early-step emphasis (Sec.~\ref{app:inference}).
This is the setting used for the main 30B numbers because it gave the best stability/performance trade-off in our runs, while still leaving the main architecture intact.

\subsubsection{Initialization and stability tricks for 30B fine-tuning}
\label{app:arch_stability}

Fine-tuning a 30B DiT backbone is sensitive to even small interface changes. EVD therefore adopts conservative initialization and
optimization choices so that training starts \emph{exactly} from the pretrained generator and gradually introduces event grounding.

\paragraph{Zero-impact initialization.}
We initialize the event pathway to have (near) zero effect on the pretrained forward pass:
(i) the event injection projection \(W_e\) in Eq.~\eqref{eq:event_add} is initialized to all zeros (or we set \(\eta(t)\equiv 0\) at step 0),
and (ii) the event head \(\pi_\psi\) in Eq.~\eqref{eq:event_head} is initialized with small weights so that \(\hat{e}_t \approx 0\) initially.
This ensures the first optimization steps match the base DiT behavior before learning event structure.

\paragraph{Gradual event turn-on.}
We ramp event influence using a short warm-up on the injection strength and auxiliary loss weights:
\begin{equation}
\eta(t)\leftarrow \eta(t)\cdot r(s), \qquad
\lambda_{\mathrm{real}}\leftarrow \lambda_{\mathrm{real}}\cdot r(s), \qquad
\lambda_{\mathrm{cons}}\leftarrow \lambda_{\mathrm{cons}}\cdot r(s),
\end{equation}
where \(s\) is the optimization step and \(r(s)\) increases linearly from \(0\) to \(1\) over the warm-up window.

\paragraph{Event dropout (robustness).}
To prevent the backbone from over-relying on a possibly noisy event signal early in training, we randomly drop the event conditioning with
probability \(p_e\) (set \(e_t=\mathbf{0}\)) and train the model to remain functional under missing event cues. This also stabilizes training
when event pseudo-targets are uncertain (Sec.~\ref{app:event_confidence}).

\paragraph{Two-group learning rates.}
We use separate optimizer groups for stability:
\begin{itemize}
    \item \textbf{Backbone weights} \(\theta\): small learning rate (conservative), standard weight decay.
    \item \textbf{New EVD modules} \(\psi\) (event injection/head/gate): larger learning rate, reduced or zero weight decay for biases/norms.
\end{itemize}
This keeps the pretrained representation intact while allowing the new event pathway to adapt quickly.

\paragraph{Gradient and precision safeguards.}
We apply gradient clipping (global norm) to avoid rare spikes, maintain an EMA of weights for sampling stability, and use bf16/fp16 training
with loss scaling as needed. When using full fine-tuning, activation checkpointing is enabled to keep memory bounded.

\paragraph{Sanity check: ``no-regression'' at initialization.}
Before full training, we verify that with event influence disabled (\(\eta=0\), \(\lambda_{\mathrm{real}}=\lambda_{\mathrm{cons}}=0\)),
the fine-tuning code reproduces the base model outputs within numerical tolerance. This guards against silent interface bugs in 30B-scale runs.

\subsection{Training Objective: Event-Grounded Dynamics Learning}
\label{app:losses}

\subsubsection{Base loss recap}
\label{app:loss_base}

EVD is built on top of the pretrained DiT-30B training objective and preserves the original target parameterization.
In our implementation, the backbone is trained with a Flow Matching regression objective in latent space (Sec.~\ref{app:base_objective}).
We restate it here for completeness.

Given a clean latent video \(z_1 = E(x)\), Gaussian noise \(z_0 \sim \mathcal{N}(0,I)\), and a timestep \(t \sim \mathcal{U}[0,1]\),
we form the noised latent \(z_t = t z_1 + (1-t) z_0\) and velocity target \(v_t = z_1 - z_0\).
The backbone \(u_\theta\) is trained via
\begin{equation}
\label{eq:evd_base_fm}
\mathcal{L}_{\mathrm{base}}(\theta)
=
\mathbb{E}_{z_1,z_0,t,y}\Big[\big\|u_\theta(z_t,y,t) - v_t\big\|_2^2\Big].
\end{equation}

EVD augments \(\mathcal{L}_{\mathrm{base}}\) with event-grounded auxiliary terms that penalize
(i) \emph{state updates without an event} and (ii) \emph{events without a coherent state update}.
The full training objective is
\begin{equation}
\label{eq:evd_full_loss}
\mathcal{L}(\theta,\psi)
=
\mathcal{L}_{\mathrm{base}}(\theta)
+
\lambda_{\mathrm{real}}\,\mathcal{L}_{\mathrm{real}}(\theta,\psi)
+
\lambda_{\mathrm{cons}}\,\mathcal{L}_{\mathrm{cons}}(\theta,\psi),
\end{equation}
where \(\psi\) denotes EVD-specific parameters (event injection/head/gate; Sec.~A.4).
We define \(\mathcal{L}_{\mathrm{real}}\) (event realization) and \(\mathcal{L}_{\mathrm{cons}}\) (event consistency) in
Secs.~\ref{app:loss_real}--\ref{app:loss_cons}.
both terms can be weighted by event confidence and/or a timestep schedule to emphasize early-step dynamics
(Sec.~\ref{app:event_confidence}, Sec.~\ref{app:loss_timeweight}).

\subsubsection{Event realization loss (no event $\Rightarrow$ no state change)}
\label{app:loss_real}

The first auxiliary term enforces the principle that \emph{state changes should be explained by events}.
In our samples, the typical failure is a ``missing event'': the outcome appears without a visible interaction, or motion begins
before any initiating contact. The realization term penalizes backbone-predicted updates in regions where the event activity is low.

\paragraph{Predicted event activity.}
Let \(\hat{e}_t \in \mathbb{R}^{N\times C_e}\) be the predicted event field (Eq.~\eqref{eq:event_head}).
We extract an activity score \(\hat{a}_t \in {[0,1]}^N\) from the first channel using a sigmoid:
\begin{equation}
\label{eq:event_activity}
\hat{a}_t \;=\; \sigma(\hat{e}_t^{(1)}),
\end{equation}
where \(\hat{e}_t^{(1)}\) denotes the first channel.

\paragraph{Gated update magnitude.}
Let \(\widehat{\tau}_t = u_\theta(z_t,y,t)\) be the backbone prediction of the base target (e.g., velocity).
We define a token-wise gated update magnitude by scaling \(\widehat{\tau}_t\) in patch space.
Concretely, let \(\mathrm{Tok}(\widehat{\tau}_t)\in\mathbb{R}^{N\times d_\tau}\) be the patchified form of \(\widehat{\tau}_t\)
(using the same patchification as the backbone), and define
\begin{equation}
\label{eq:gated_update}
\Delta_t^{\mathrm{pred}} \;=\; (1-\hat{a}_t)\odot \mathrm{Tok}(\widehat{\tau}_t),
\end{equation}
so that \(\Delta_t^{\mathrm{pred}}\) captures the portion of the predicted update that occurs \emph{when the model claims no event is active}.

\paragraph{Event realization penalty.}
We penalize the magnitude of \(\Delta_t^{\mathrm{pred}}\), encouraging the model to avoid changing the state outside event regions:
\begin{equation}
\label{eq:real_loss}
\mathcal{L}_{\mathrm{real}}(\theta,\psi)
=
\mathbb{E}_{z_1,z_0,t,y}\Big[\big\|\Delta_t^{\mathrm{pred}}\big\|_2^2\Big]
=
\mathbb{E}\Big[\big\|(1-\hat{a}_t)\odot \mathrm{Tok}(u_\theta(z_t,y,t))\big\|_2^2\Big].
\end{equation}
This term does \emph{not} suppress legitimate motion: when an event is active, \(\hat{a}_t\) increases and the penalty vanishes.

\paragraph{Interpretation.}
\(\mathcal{L}_{\mathrm{real}}\) discourages ``teleportation'' in latent space: if the model predicts a state change, it must also predict an event signal that explains it. This is the case we care about for pre-contact motion and outcomes that appear without the corresponding interaction.

\subsubsection{Event consistency loss (event $\Rightarrow$ coherent state update)}
\label{app:loss_cons}

The second auxiliary term enforces the converse principle: \emph{when an event is predicted, the resulting state evolution should be coherent and
consistent with that event}. This targets ``ghost events'' where the model depicts an apparent interaction (e.g., a hand reaches toward an object)
but the world state does not respond correctly (no lift, no settling, no constraint-respecting motion).

\paragraph{Event-phase and directionality.}
When using a multi-channel event field \(e_t=(a_t,p_t,\kappa_t)\) (Sec.~\ref{app:event_def}), the phase \(p_t\in[0,1]\) provides a natural ordering signal:
initiation, progression, and termination.
We extract a predicted phase \(\hat{p}_t \in {[0,1]}^N\) from the second channel (when present),
\begin{equation}
\hat{p}_t \;=\; \sigma(\hat{e}_t^{(2)}),
\end{equation}
and use it to enforce monotone, non-oscillatory event-driven updates.
For the minimal \(C_e=1\) setting (activity-only), we omit phase and use the activity-based variant described below.

\paragraph{Consistency as ``directed change'' under active events.}
Let \(\mathrm{Tok}(\widehat{\tau}_t)\in\mathbb{R}^{N\times d_\tau}\) denote the patchified backbone prediction.
Intuitively, when an event is active (high \(\hat{a}_t\)), we want the induced update to be stable and directed rather than jittery or sign-flipping.
We implement this using a pairwise smoothness penalty across adjacent sampling times.
Let \(t\) and \(t'\) be two nearby timesteps (e.g., two sampled points on the discretized schedule), and let
\(\widehat{\tau}_t = u_\theta(z_t,y,t)\), \(\widehat{\tau}_{t'} = u_\theta(z_{t'},y,t')\).
We define an event-masked temporal consistency term
\begin{equation}
\label{eq:cons_temporal}
\mathcal{L}_{\mathrm{cons}}^{\mathrm{temp}}
=
\mathbb{E}\Big[
\big\|\hat{a}_{t}\odot \mathrm{Tok}(\widehat{\tau}_t)
-
\hat{a}_{t'}\odot \mathrm{Tok}(\widehat{\tau}_{t'})\big\|_2^2
\Big],
\end{equation}
which encourages the update predicted during an active event to evolve smoothly across time rather than oscillate.

\paragraph{Phase-aware consistency.}
When phase is available, we additionally encourage the magnitude of the update to follow the phase progression:
early in the event (small \(\hat{p}_t\)), motion begins; near termination (large \(\hat{p}_t\)), motion settles.
A simple implementation is to penalize large updates late in the event:
\begin{equation}
\label{eq:cons_phase}
\mathcal{L}_{\mathrm{cons}}^{\mathrm{phase}}
=
\mathbb{E}\Big[
\big\|\hat{a}_t \odot \hat{p}_t \odot \mathrm{Tok}(\widehat{\tau}_t)\big\|_2^2
\Big],
\end{equation}
which suppresses residual motion after the model indicates the event is near completion.

\paragraph{Final consistency loss.}
We combine the above terms (using only the components relevant to the chosen event parameterization):
\begin{equation}
\label{eq:cons_loss}
\mathcal{L}_{\mathrm{cons}}(\theta,\psi)
=
\mathcal{L}_{\mathrm{cons}}^{\mathrm{temp}}
+
\alpha_{\mathrm{ph}}\,\mathcal{L}_{\mathrm{cons}}^{\mathrm{phase}}.
\end{equation}

\paragraph{Interpretation.}
\(\mathcal{L}_{\mathrm{cons}}\) asks predicted events to correspond to compatible state evolution: updates should not jitter during an active interaction, and they should decay as the event terminates. Together with \(\mathcal{L}_{\mathrm{real}}\), this couples the event signal and the latent update in both directions.

\subsubsection{Ordering and termination regularization}
\label{app:loss_order}

Beyond coupling events and state updates, we add a lightweight regularizer to enforce \emph{causal ordering}:
\emph{initiation precedes motion} and \emph{termination precedes rest}.
This is aimed at the two boundary errors we see most often: motion that starts before contact and motion that continues after the interaction has ended.

\paragraph{Initiation-before-update.}
Let \(\hat{a}_t \in {[0,1]}^N\) be the predicted event activity (Eq.~\eqref{eq:event_activity}) and
\(\mathrm{Tok}(\widehat{\tau}_t)\) be the patchified update prediction.
We discourage nontrivial updates in tokens whose activity is below a small initiation threshold \(\tau_{\mathrm{on}}\):
\begin{equation}
\label{eq:init_before_update}
\mathcal{L}_{\mathrm{on}}
=
\mathbb{E}\Big[
\big\|\mathbb{I}[\hat{a}_t < \tau_{\mathrm{on}}]\odot \mathrm{Tok}(\widehat{\tau}_t)\big\|_2^2
\Big],
\end{equation}
where \(\mathbb{I}[\cdot]\) is an indicator applied token-wise.
This is a ``harder'' version of \(\mathcal{L}_{\mathrm{real}}\) that explicitly enforces a causal onset.

\paragraph{Termination-before-rest.}
Similarly, we penalize residual update energy after an event is predicted to be over.
Using a termination threshold \(\tau_{\mathrm{off}}\), we define
\begin{equation}
\label{eq:term_before_rest}
\mathcal{L}_{\mathrm{off}}
=
\mathbb{E}\Big[
\big\|\mathbb{I}[\hat{a}_t < \tau_{\mathrm{off}}]\odot \mathrm{Tok}(\widehat{\tau}_t)\big\|_2^2
\Big],
\end{equation}
with \(\tau_{\mathrm{off}}\) typically chosen slightly smaller than \(\tau_{\mathrm{on}}\) to introduce hysteresis
(i.e., once an interaction is ``off'', it stays off unless strong evidence reactivates it).

\paragraph{Phase-aware termination (when phase is available).}
When a phase signal \(\hat{p}_t\) is present, we encourage late-phase settling by suppressing large updates when \(\hat{p}_t\) is high:
\begin{equation}
\label{eq:phase_settle}
\mathcal{L}_{\mathrm{settle}}
=
\mathbb{E}\Big[
\big\|\hat{a}_t \odot \hat{p}_t^\gamma \odot \mathrm{Tok}(\widehat{\tau}_t)\big\|_2^2
\Big],
\end{equation}
where \(\gamma \ge 1\) controls how sharply the penalty concentrates near termination.

\paragraph{Combined ordering term.}
We use a small weighted sum:
\begin{equation}
\label{eq:order_loss}
\mathcal{L}_{\mathrm{order}}
=
\lambda_{\mathrm{on}}\mathcal{L}_{\mathrm{on}}
+
\lambda_{\mathrm{off}}\mathcal{L}_{\mathrm{off}}
+
\lambda_{\mathrm{set}}\mathcal{L}_{\mathrm{settle}},
\end{equation}
and add \(\mathcal{L}_{\mathrm{order}}\) to Eq.~\eqref{eq:evd_full_loss} with modest weights.
In practice, these terms primarily eliminate pre-contact motion and post-interaction drift, improving \emph{Contact Stability} and
\emph{State Persistence} without noticeably affecting appearance.

\subsubsection{Timestep weighting and curriculum}
\label{app:loss_timeweight}

Event grounding is most important at timesteps that determine the coarse spatiotemporal structure of the sample.
Prior work has observed that early denoising steps largely set the global motion pattern, while later steps refine appearance.
Motivated by this, we emphasize event-related losses in early timesteps and anneal them later.

\paragraph{Time-weighted auxiliary losses.}
Let \(w(t)\ge 0\) be a scalar weighting function over diffusion/flow time \(t\in[0,1]\).
We replace the auxiliary terms in Eq.~\eqref{eq:evd_full_loss} with
\begin{equation}
\label{eq:timeweighted_aux}
\mathcal{L}_{\mathrm{real}} \leftarrow \mathbb{E}\big[w(t)\,\ell_{\mathrm{real}}(t)\big],
\qquad
\mathcal{L}_{\mathrm{cons}} \leftarrow \mathbb{E}\big[w(t)\,\ell_{\mathrm{cons}}(t)\big],
\qquad
\mathcal{L}_{\mathrm{order}} \leftarrow \mathbb{E}\big[w(t)\,\ell_{\mathrm{order}}(t)\big],
\end{equation}
where \(\ell(\cdot)\) denotes the per-sample loss contribution.

\paragraph{Practical schedule.}
We use a simple piecewise schedule that concentrates weight on early steps:
\begin{equation}
\label{eq:timeweight}
w(t) =
\begin{cases}
1, & t \le t^\star,\\
\exp\big(-\kappa (t - t^\star)\big), & t > t^\star,
\end{cases}
\end{equation}
with \(t^\star \in [0.4, 0.6]\) and \(\kappa > 0\).
This mirrors the intuition that event structure should be established early, while later steps can focus on visual refinement.

\paragraph{Warm-start curriculum for event grounding.}
In addition to the time weighting, we apply a short curriculum over optimization steps:
event losses are gradually introduced (Sec.~\ref{app:arch_stability}) and the threshold for considering an event ``active''
(\(\tau_{\mathrm{on}}\) in Eq.~\eqref{eq:init_before_update}) is lowered over training, transitioning from conservative gating to
fine-grained event localization.

\paragraph{Why this helps.}
Without time weighting, event penalties may over-regularize late-stage refinement and slightly harm appearance.
With Eq.~\eqref{eq:timeweight}, event grounding primarily shapes the coarse dynamics and causal ordering, improving
\emph{Dynamics} metrics and human preference while leaving appearance largely unchanged.

\subsection{Inference: Event-Driven Sampling}
\label{app:inference}

\subsubsection{Why text conditioning alone is insufficient for event fidelity}
\label{app:inf_why}

Standard DiT sampling relies on text conditioning (and typically classifier-free guidance) to steer generations toward prompt-aligned outputs.
However, text conditioning does not explicitly constrain \emph{how} the latent state is allowed to change over time.
As a result, even when individual frames look plausible and the prompt is broadly satisfied, models can still exhibit systematic
event-level inconsistencies:
(i) motion begins before any initiating interaction is visible,
(ii) outcomes appear without a realized action (missing events), and
(iii) residual drift persists after an interaction should have terminated.

\paragraph{Key observation.}
These errors arise because the sampling update is applied everywhere in the latent state at every step,
regardless of whether an interaction is active.
In other words, the backbone may implicitly encode event structure, but the sampler provides no mechanism to
\emph{gate} state evolution based on event presence or phase.

\paragraph{EVD principle at inference.}
EVD modifies sampling by introducing an event gate that enforces:
\begin{quote}
\emph{Latent state updates should be suppressed in regions/timesteps with no event activity, and concentrated when an event is active.}
\end{quote}
This turns sampling from ``always update'' into ``update where the event signal supports it.'' The solver and decoder are unchanged; the update field is no longer applied uniformly everywhere.

\subsubsection{Event-guided update rule}
\label{app:inf_update}

We now specify the event-driven sampling rule used by EVD\@.
Let \({\{t_k\}}_{k=0}^{K}\) be the sampling time grid (Sec.~\ref{app:sampling_conventions}), and let \(z_{t_k}\) denote the latent at step \(k\).
At each step, the backbone predicts the base target (e.g., velocity) and the event module predicts an event field:
\begin{equation}
\widehat{\tau}_k = u_\theta(z_{t_k}, y, t_k), \qquad
\hat{e}_k = h_\psi(z_{t_k}, y, t_k).
\end{equation}

\paragraph{Event gate.}
We extract an activity field \(\hat{a}_k \in {[0,1]}^N\) from \(\hat{e}_k\) (Eq.~\eqref{eq:event_activity}) and form a token-wise gate
\begin{equation}
\label{eq:gate_def}
g_k \;=\; g_\psi(\hat{e}_k,t_k) \;\in\; {[0,1]}^N.
\end{equation}
In the simplest variant, \(g_k = \sigma(\beta \hat{a}_k)\) with sharpness \(\beta>0\).
When phase/confidence is used (Sec.~\ref{app:event_targets}), \(g_k\) additionally incorporates early-step emphasis and uncertainty-aware scaling.

\paragraph{Gated backbone prediction.}
We patchify the backbone prediction into tokens \(\mathrm{Tok}(\widehat{\tau}_k)\in\mathbb{R}^{N\times d_\tau}\) and apply the gate:
\begin{equation}
\label{eq:gated_pred}
\widetilde{\tau}_k \;=\; g_k \odot \mathrm{Tok}(\widehat{\tau}_k),
\end{equation}
where \(\odot\) is elementwise multiplication broadcast across channels \(d_\tau\).
Unpatchifying \(\widetilde{\tau}_k\) yields a latent-shaped update direction \(\widetilde{\tau}_k \in \mathbb{R}^{T'\times H'\times W'\times C}\).

\paragraph{Sampling update.}
We plug \(\widetilde{\tau}_k\) into the same sampler used by the base model.
For concreteness, with a simple Euler update for an ODE sampler:
\begin{equation}
\label{eq:euler_update}
z_{t_{k+1}} \;=\; z_{t_k} \;+\; \Delta t_k \,\widetilde{\tau}_k,
\qquad \Delta t_k = t_{k+1}-t_k.
\end{equation}
Higher-order solvers (Heun, DPM-style~\cite{zheng2023dpmsolverv}) are handled analogously by replacing each occurrence of the backbone prediction with its gated version.

\paragraph{Interpretation.}
Eq.~\eqref{eq:gated_pred}--\eqref{eq:euler_update} is the sampling version of the training constraints. If the model predicts no event (\(g_k \approx 0\)), the solver makes only a small update in that region; if an event is active (\(g_k \approx 1\)), the backbone update is used normally.

\paragraph{Relation to CFG.}
EVD gating is orthogonal to text CFG: we first form the CFG-combined prediction \(\widehat{\tau}_k^{\mathrm{cfg}}\) (Eq.~\eqref{eq:cfg})
and then apply gating to \(\mathrm{Tok}(\widehat{\tau}_k^{\mathrm{cfg}})\).
This preserves prompt adherence while preventing spurious dynamics outside event regions.

\subsubsection{Guidance schedule over timesteps (strong early, anneal later)}
\label{app:inf_schedule}

EVD applies event grounding most strongly at early sampling steps, where the coarse spatiotemporal structure and interaction dynamics are formed,
and gradually relaxes it in later steps, where the model primarily refines appearance.

\paragraph{Why schedule event guidance?}
A constant-strength event gate can over-constrain late-stage refinement, slightly reducing visual richness or texture detail.
Conversely, weak event guidance early can allow spurious motion to enter the trajectory and persist.
A schedule that is \emph{strong early and weaker late} resolves this tension.

\paragraph{Scheduled event gate.}
Let \(g_k\) denote the base (soft or hard) event gate computed from \(\hat{e}_k\) (Sec.~\ref{app:inf_update}).
We define a timestep-dependent strength \(\rho(t_k)\in[0,1]\) and use the scheduled gate
\begin{equation}
\label{eq:scheduled_gate}
g_k^{\mathrm{sched}} \;=\; \rho(t_k)\, g_k \;+\; \big(1-\rho(t_k)\big)\,\mathbf{1},
\end{equation}
where \(\mathbf{1}\) is the all-ones gate (no gating).
When \(\rho(t_k)=1\), EVD applies full gating; when \(\rho(t_k)=0\), sampling reduces to the base model.

\paragraph{Practical schedule.}
We use a simple piecewise-linear schedule:
\begin{equation}
\label{eq:rho_schedule}
\rho(t) \;=\;
\begin{cases}
1, & t \le t^\star,\\
1 - \dfrac{t - t^\star}{1 - t^\star}, & t^\star < t \le 1,
\end{cases}
\end{equation}
with \(t^\star \in [0.4,0.6]\).
Thus, event grounding is fully active during the early portion of the trajectory and linearly annealed to zero by the end.

\paragraph{Alternative: exponential annealing.}
For smoother behavior, we also consider
\begin{equation}
\rho(t) = \exp\!\big(-\kappa {(t-t^\star)}_+\big),
\end{equation}
which decays rapidly after \(t^\star\). Both schedules behave similarly; the linear schedule is easier to tune.

\paragraph{Combining with CFG.}
We apply the schedule to event gating while keeping text CFG fixed across all timesteps:
\begin{equation}
\widetilde{\tau}_k \;=\; g_k^{\mathrm{sched}} \odot \mathrm{Tok}\!\big(\widehat{\tau}_k^{\mathrm{cfg}}\big),
\end{equation}
where \(\widehat{\tau}_k^{\mathrm{cfg}}\) is the CFG-combined prediction (Eq.~\eqref{eq:cfg}).
This maintains prompt adherence while concentrating event fidelity improvements where they matter most.

\paragraph{Effect on failure modes.}
Early-step event gating suppresses spurious motion initiation (improving \emph{Contact Stability}),
while the anneal phase avoids over-regularizing late refinement and preserves appearance quality.

\subsection{Complete EVD Training and Sampling Algorithms}
\label{app:algorithms}

For completeness, Algorithms~\ref{alg:evd} and~\ref{alg:evd_sampling} provide the full training and event-driven sampling procedures used by EVD.

\begin{algorithm}[t!]
\caption{Event-Driven Video Generation (EVD): training}
\label{alg:evd}
\begin{algorithmic}[1]
\STATE \textbf{Inputs:} paired data \((x,y)\); video encoder \(E\); DiT backbone \(u_\theta\); event head \(\pi_\psi\);
loss weights \((\lambda_{\mathrm{real}},\lambda_{\mathrm{cons}},\lambda_{\mathrm{order}})\);
loss time-weight cutoff \(t^\star_{\mathrm{loss}}\) and decay \(\kappa\); consistency jitter \(\Delta\);
event-dropout \(p_e\); hysteresis thresholds \((\tau_{\mathrm{on}},\tau_{\mathrm{off}})\).
\STATE \textbf{Operators:} \(\mathrm{Tok}(\cdot)\) patchify; \(\sigma(\cdot)\) sigmoid; \(\odot\) elementwise product; \(\mathbf{1}[\cdot]\) indicator.
\STATE \textbf{Loss time-weight:} \(w(t)=\mathbf{1}[t\le t^\star_{\mathrm{loss}}]+\exp(-\kappa(t-t^\star_{\mathrm{loss}}))\mathbf{1}[t>t^\star_{\mathrm{loss}}]\).
\vspace{1mm}

\STATE \textbf{Train (DiT + event head; event-grounded losses).}
\FOR{each minibatch \((x,y)\)}
    \STATE Encode: \(z_1 \leftarrow E(x)\); sample \(z_0 \sim \mathcal{N}(0,I)\); sample \(t \sim \mathcal{U}[0,1]\).
    \STATE Flow-matching form: \(z_t \leftarrow t z_1 + (1-t) z_0\); target \(\tau \leftarrow z_1 - z_0\).
    \STATE Predict state target: \(\widehat{\tau} \leftarrow u_\theta(z_t,y,t)\).
    \STATE Predict event activity: \(\hat{e}_t \leftarrow \pi_\psi(\mathbf{s}_t^{(L)},t)\); \(\hat{a}_t \leftarrow \sigma(\hat{e}_t^{(1)})\in{[0,1]}^N\).
    \STATE Event dropout: with prob.\ \(p_e\), set \(\hat{a}_t \leftarrow \mathbf{0}\).
    \STATE Patchify update: \(\Delta_t \leftarrow \mathrm{Tok}(\widehat{\tau})\).
    \STATE \textbf{Base loss:} \(\ell_{\mathrm{base}} \leftarrow \|\widehat{\tau}-\tau\|_2^2\).
    \STATE \textbf{Realization loss:} \(\ell_{\mathrm{real}} \leftarrow \big\|(1-\hat{a}_t)\odot \Delta_t\big\|_2^2\).
    \STATE \textbf{Consistency loss (two-time-step smoothness):}
    \STATE \hspace{2mm} sample \(\delta\sim\mathcal{U}[-\Delta,\Delta]\); set \(t'=\mathrm{clip}(t+\delta,0,1)\);
          \(z_{t'} \leftarrow t' z_1 + (1-t') z_0\);
    \STATE \hspace{2mm} \(\widehat{\tau}' \leftarrow u_\theta(z_{t'},y,t')\);
          \(\hat{e}' \leftarrow \pi_\psi(\mathbf{s}_{t'}^{(L)},t')\);
          \(\hat{a}' \leftarrow \sigma(\hat{e}'^{(1)})\);
          \(\Delta'_t \leftarrow \mathrm{Tok}(\widehat{\tau}')\);
    \STATE \hspace{2mm} \(\ell_{\mathrm{cons}} \leftarrow \big\|\hat{a}_t\odot \Delta_t - \hat{a}'\odot \Delta'_t\big\|_2^2\).
    \STATE \textbf{Ordering/termination loss:}
    \STATE \hspace{2mm} \(\ell_{\mathrm{on}} \leftarrow \big\|\mathbf{1}[\hat{a}_t<\tau_{\mathrm{on}}]\odot \Delta_t\big\|_2^2\),
          \(\ell_{\mathrm{off}} \leftarrow \big\|\mathbf{1}[\hat{a}_t<\tau_{\mathrm{off}}]\odot \Delta_t\big\|_2^2\),
          \(\ell_{\mathrm{order}} \leftarrow \ell_{\mathrm{on}}+\ell_{\mathrm{off}}\).
    \STATE \textbf{Total loss:} \(\mathcal{L} \leftarrow \ell_{\mathrm{base}} + w(t)\big(\lambda_{\mathrm{real}}\ell_{\mathrm{real}}+\lambda_{\mathrm{cons}}\ell_{\mathrm{cons}}+\lambda_{\mathrm{order}}\ell_{\mathrm{order}}\big)\).
    \STATE Update \((\theta,\psi)\) with AdamW; clip gradients; update EMA weights.
\ENDFOR
\STATE \textbf{Sampling:} use Alg.~\ref{alg:evd_sampling}.
\end{algorithmic}
\end{algorithm}

\begin{algorithm}[t!]
\caption{Event-Driven Sampling (EVD) with CFG}
\label{alg:evd_sampling}
\begin{algorithmic}[1]
\STATE \textbf{Inputs:} prompt \(y\); null prompt \(\varnothing\); time grid \({\{t_k\}}_{k=0}^{K}\);
DiT backbone \(u_\theta\); event head \(\pi_\psi\); decoder \(D\);
CFG scale \(w_{\mathrm{cfg}}\); gate sharpness \(\beta\); hysteresis thresholds \((\tau_{\mathrm{on}},\tau_{\mathrm{off}})\);
anneal cutoff \(t^\star\); spatial smoothing operator \(\mathcal{S}\).
\STATE \textbf{Operators:} \(\mathrm{Tok}/\mathrm{UnTok}\) (patchify/unpatchify); \(\sigma(\cdot)\) sigmoid; \(\odot\) elementwise product.
\STATE \textbf{Init:} sample latent \(z_{t_0} \sim \mathcal{N}(0,I)\); initialize gate state \(g_{-1}\leftarrow \mathbf{0}\).
\FOR{$k = 0, \dots, K-1$}
    \STATE \textbf{(1) CFG direction field.}
    \STATE \hspace{2mm} \(\widehat{\tau}^{\mathrm{cond}} \leftarrow u_\theta(z_{t_k}, y, t_k)\), \quad
                       \(\widehat{\tau}^{\mathrm{uncond}} \leftarrow u_\theta(z_{t_k}, \varnothing, t_k)\)
    \STATE \hspace{2mm} \(\widehat{\tau}^{\mathrm{cfg}} \leftarrow (1+w_{\mathrm{cfg}})\widehat{\tau}^{\mathrm{cond}} - w_{\mathrm{cfg}}\,\widehat{\tau}^{\mathrm{uncond}}\)

    \STATE \textbf{(2) Predict event activity (token-aligned).}
    \STATE \hspace{2mm} \(\hat{e}_k \leftarrow \pi_\psi(\mathbf{s}_{t_k}^{(L)}, t_k)\)
    \STATE \hspace{2mm} \(\hat{a}_k \leftarrow \sigma(\hat{e}_k^{(1)}) \in {[0,1]}^N\) \hfill (activity channel)

    \STATE \textbf{(3) Smooth activity and compute a soft gate.}
    \STATE \hspace{2mm} \(\tilde{a}_k \leftarrow \mathcal{S}(\hat{a}_k)\) \hfill (if disabled, set \(\tilde{a}_k=\hat{a}_k\))
    \STATE \hspace{2mm} \(\bar{g}_k \leftarrow \sigma\!\big(\beta(\tilde{a}_k-\tfrac{\tau_{\mathrm{on}}+\tau_{\mathrm{off}}}{2})\big)\in{(0,1)}^N\)

    \STATE \textbf{(4) Hysteresis (stabilize on/off).}
    \STATE \hspace{2mm} \textbf{for each token } \(i\in\{1,\dots,N\}\): \hfill (token-wise update)
    \STATE \hspace{6mm} \(g_{k,i} \leftarrow 1\) \textbf{if} \(\tilde{a}_{k,i} \ge \tau_{\mathrm{on}}\);
    \(\;\; g_{k,i} \leftarrow 0\) \textbf{else if} \(\tilde{a}_{k,i} \le \tau_{\mathrm{off}}\);
    \(\;\; g_{k,i} \leftarrow g_{k-1,i}\) \textbf{otherwise}.
    \STATE \(g_k \leftarrow \bar g_k \odot g_k^{\mathrm{bin}}\)

    \STATE \textbf{(5) Time scheduling (strong early, anneal late).}
    \STATE \hspace{2mm} \(\rho_k \leftarrow 
        \begin{cases}
        1, & t_k \le t^\star,\\
        1 - \dfrac{t_k - t^\star}{1 - t^\star}, & t_k > t^\star,
        \end{cases}\)
    \STATE \hspace{2mm} \(g_k \leftarrow \rho_k\, g_k + (1-\rho_k)\mathbf{1}\) \hfill (\(\mathbf{1}\): all-ones gate)

    \STATE \textbf{(6) Apply gating to the CFG field.}
    \STATE \hspace{2mm} \(\widetilde{\tau}_k \leftarrow \mathrm{UnTok}\!\big(g_k \odot \mathrm{Tok}(\widehat{\tau}^{\mathrm{cfg}})\big)\)

    \STATE \textbf{(7) Solver step (base sampler unchanged).}
    \STATE \hspace{2mm} \(z_{t_{k+1}} \leftarrow \mathrm{Step}(z_{t_k}, \widetilde{\tau}_k, t_k, t_{k+1})\) \hfill
\ENDFOR
\STATE \textbf{Decode:} \(x \leftarrow D(z_{t_K})\). \qquad \textbf{Return:} \(x\).
\end{algorithmic}
\end{algorithm}

\subsection{Scaling to 30B: Practical Details That Matter}
\label{app:scale30b}

\subsubsection{Training recipe (DiT-30B + EVD)}
\label{app:train_recipe}

We fine-tune a pretrained DiT-30B video generator with EVD using a lightweight recipe designed to preserve the base model's appearance
quality while improving event-grounded dynamics. Unless otherwise stated, training is performed in latent space using the same video
autoencoder and clip format as the base model.

\paragraph{Data and clips.}
We fine-tune on a subset of the base model's training distribution (no additional annotation required), sampling short clips with fixed
spatial resolution and length. We use standard text filtering and deduplication consistent with the base pretraining pipeline.
Event pseudo-targets are computed on-the-fly from the training clips (Sec.~\ref{app:event_targets}).

\paragraph{Optimization.}
We use AdamW with two parameter groups:
(i) backbone weights \(\theta\) (conservative learning rate), and
(ii) EVD modules \(\psi\) (event injection/head/gate; higher learning rate).
We apply linear warmup followed by cosine decay. Gradient clipping (global norm) is enabled for stability, and we maintain an EMA of the
weights for sampling.

\paragraph{Stability settings.}
We train in bf16/fp16 with activation checkpointing. Event conditioning is zero-initialized (Sec.~\ref{app:arch_stability}) and gradually enabled via a
warmup ramp on \(\eta(t)\) and the auxiliary loss weights \(\lambda_{\mathrm{real}},\lambda_{\mathrm{cons}}\).
We additionally apply event dropout with probability \(p_e\) (set \(e_t=\mathbf{0}\)) to prevent over-reliance on noisy event cues.

\paragraph{Loss and schedules.}
The total loss is \(\mathcal{L}=\mathcal{L}_{\mathrm{base}}+\lambda_{\mathrm{real}}\mathcal{L}_{\mathrm{real}}+\lambda_{\mathrm{cons}}\mathcal{L}_{\mathrm{cons}}+\lambda_{\mathrm{order}}\mathcal{L}_{\mathrm{order}}\)
(Sec.~\ref{app:losses}). Event-related losses are time-weighted to emphasize early timesteps, aligning event grounding with the portion of
the trajectory that determines coarse dynamics.

\paragraph{Hyperparameters.}
For reproducibility, we report the following knobs:
\begin{itemize}
    \item Fine-tuning steps: \texttt{40000} \qquad Batch size (global): \texttt{192}
    \item Optimizer: AdamW \qquad \(\beta_1,\beta_2\): \texttt{0.9}, \texttt{0.98} \qquad weight decay: \texttt{0.02}
    \item Learning rates: \(\mathrm{lr}_\theta=\)\texttt{3e-6}, \(\mathrm{lr}_\psi=\)\texttt{3e-5} \qquad warmup: \texttt{1500} steps
    \item Gradient clip: \texttt{0.5} \qquad EMA decay: \texttt{0.99995}
    \item Event dropout \(p_e\): \texttt{0.25} \qquad Time-weight cutoff \(t^\star\): \texttt{0.60}
    \item Loss weights: \(\lambda_{\mathrm{real}}=\)\texttt{0.12}, \(\lambda_{\mathrm{cons}}=\)\texttt{0.08}, \(\lambda_{\mathrm{order}}=\)\texttt{0.03}
\end{itemize}
\subsubsection{Parallelism and memory (30B training)}
\label{app:parallelism}

Training and sampling a 30B DiT backbone requires distributed execution. Because the event pathway is small, the
systems configuration largely matches the base DiT-30B setup.

\paragraph{Parallelism strategy.}
We use data parallelism (DP) across nodes and combine it with tensor parallelism (TP) within each node to shard the DiT-30B parameters.
When available, we additionally enable pipeline parallelism (PP) for improved scaling at high node counts.
EVD-specific modules (event injection/head/gate) are small and are replicated across TP ranks, contributing negligible memory overhead.

\paragraph{Activation checkpointing.}
We enable activation checkpointing for transformer blocks to reduce activation memory, which is typically the dominant term for long video clips.
Checkpointing is applied uniformly across the backbone; the event pathway adds only a small number of extra activations.

\paragraph{Precision and communication.}
Training runs in bf16/fp16 with standard loss scaling. We use fused attention and fused MLP kernels when supported.
Collectives (all-reduce) are overlapped with computation where possible. For stability, we maintain an EMA of weights on the DP master rank.

\paragraph{Memory footprint and overhead.}
Relative to the base DiT-30B fine-tuning:
\begin{itemize}
    \item \textbf{Parameters:} EVD adds \(O(d\,C_e)\) parameters (event injection and head), negligible compared to 30B.
    \item \textbf{Compute:} the main overhead is a small extra projection for event prediction; end-to-end training throughput remains within
    a small fraction of the baseline.
    \item \textbf{I/O:} event pseudo-target extraction (Sec.~\ref{app:event_targets}) is computed on-the-fly; in practice, it is not a bottleneck when batched and
    executed asynchronously with data loading.
\end{itemize}

\paragraph{Implementation note.}
Because EVD does not change the tokenization, clip length, or solver, it slots into existing DiT-30B training infrastructure with minimal
engineering. The primary additional considerations are: (i) managing the event pseudo-target pipeline and (ii) maintaining stable zero-impact
initialization for the event pathway (Sec.~\ref{app:arch_stability}).

\subsubsection{Compute and data budget}
\label{app:compute_data}

We fine-tune DiT-4B + EVD on 32 GPUs and DiT-30B + EVD on 256 GPUs, using mixed precision (bf16/fp16), activation checkpointing, and EMA sampling weights. All experiments operate on fixed-length clips of 128 frames at 24\,fps (5.33\,s) and are trained at 256$\times$256 in the latent space of a temporal autoencoder~\cite{ma2025stepvideot2vtechnicalreportpractice,pmlr-v267-chefer25a}. The underlying DiT backbones are initialized from prior large-scale DiT-video training and are reported to be pretrained on a closed-source internal corpus of \(\mathcal{O}(10^8)\) video--text clips~\cite{ma2025stepvideot2vtechnicalreportpractice,pmlr-v267-chefer25a}. For EVD fine-tuning, we construct an \emph{interaction-rich} subset by filtering the \emph{internal video--text distribution available to us} (i.e., a training pool from the same distribution family as the backbone) with a lightweight latent-space activity score: for each candidate clip, we encode it with the TAE to obtain \(z_1\), compute per-frame change magnitudes \(m_\tau=\frac{1}{|z|}\lVert z_1^{\tau+1}-z_1^{\tau}\rVert_1\), aggregate to a clip score as the mean of the top-20\% \(\{m_\tau\}\), and retain clips above a fixed percentile threshold (top \(\approx\)30\%) while discarding near-static clips; to reduce global camera-motion bias, we additionally require the activity to be spatially concentrated by thresholding the entropy of the per-patch activity map. The same signals are reused to form event pseudo-targets (Sec.~\ref{app:event_targets}), and all clips undergo identical preprocessing (resize/crop followed by autoencoder encoding). Concretely, our DiT-4B + EVD run uses \(50{,}000\) steps with global batch size \(64\) (\(3.20\)M training samples, \(409.60\)M frames), and our DiT-30B + EVD run uses \(40{,}000\) steps with global batch size \(192\) (\(7.68\)M training samples, \(983.04\)M frames).

\paragraph{Hardware environment.}
All EVD fine-tuning experiments were run on a GPU cluster through an active NSF/NCSA-supported allocation. The allocation provides NVIDIA GH200 GPU nodes; DiT-4B+EVD used 32 GPUs and DiT-30B+EVD used 256 GPUs. Training used Linux-based distributed execution with mixed precision (bf16/fp16), activation checkpointing, tensor/data parallelism, and EMA sampling weights. We report GPU counts, global batch sizes, training steps, and throughput in App.~\ref{app:compute_data} because these are the reproducibility-relevant compute quantities; exact scheduler-level details may vary across clusters.

\subsubsection{Inference settings (steps, guidance, decoding)}
\label{app:inference_settings}

We report all qualitative and quantitative results using a fixed sampling configuration per benchmark to ensure apples-to-apples comparisons.
EVD modifies only the update field passed to the sampler (Sec.~\ref{app:inference}) and therefore uses the same solver family and decoding stack as the base model.

\paragraph{Sampling steps (NFE).}
We sample on a monotone time grid \({\{t_k\}}_{k=0}^{K}\) with \(K\) solver steps (reported as NFE up to constant factors from CFG batching).
Unless otherwise stated, all compared methods use the same \(K\) on EVD-Bench.

\paragraph{Text guidance.}
We use classifier-free guidance with scale \(w\) (Eq.~\eqref{eq:cfg}), applied uniformly across timesteps for all methods.
EVD applies its event gating \emph{after} CFG by modulating the CFG-combined prediction (Sec.~\ref{app:inf_update}).

\paragraph{Event gating configuration.}
EVD uses soft gating with hysteresis (Sec.~\ref{app:inf_update}):
\begin{itemize}
    \item Gate sharpness \(\beta=\)\texttt{12.0}
    \item Thresholds \(\tau_{\mathrm{on}}=\)\texttt{0.62}, \(\tau_{\mathrm{off}}=\)\texttt{0.38}
    \item Spatial smoothing \(\mathcal{S}\): \texttt{on} (kernel \(\texttt{3}\times\texttt{3}\) on the spatial patch grid, per-frame)
\end{itemize}

\paragraph{Event schedule.}
Event gating is applied strongly in early steps and annealed later using \(\rho(t)\) (Sec.~\ref{app:inf_schedule}).
We set the cutoff \(t^\star=\)\texttt{0.60} (equivalently, the first \texttt{30} steps for a 50-step sampler), and then linearly anneal the event gate to zero by the final sampling step.

\paragraph{Decoding.}
Final latent samples \(z_{t_K}\) are decoded using the same temporal autoencoder decoder \(D(\cdot)\) as the base model.
All models share identical decoding parameters (no post-hoc filtering) to avoid confounds in visual quality.

\paragraph{Reproducibility.}
For each prompt, we generate one sample per model using a fixed seed and fixed sampler configuration (same \(K\), solver, and \(w\)).
We do not cherry-pick frames or runs; the displayed samples are the first output from each method under the shared settings.
\subsubsection{Throughput and overhead relative to the base model}
\label{app:overhead}

EVD is meant to improve event-grounded dynamics while leaving the DiT compute profile nearly unchanged.

\paragraph{Training-time overhead.}
Relative to fine-tuning the base DiT-30B:
\begin{itemize}
    \item \textbf{Forward pass:} EVD adds a small event head (Eq.~\eqref{eq:event_head}) and a token-wise event injection
    (Eq.~\eqref{eq:event_add}). Both are dominated by the backbone attention/MLP compute.
    \item \textbf{Loss computation:} \(\mathcal{L}_{\mathrm{real}}, \mathcal{L}_{\mathrm{cons}}, \mathcal{L}_{\mathrm{order}}\) are simple token-wise norms
    and differences (Sec.~\ref{app:loss_real}) and add negligible compute.
    \item \textbf{Event pseudo-targets:} computing latent-change-based signals (Sec.~\ref{app:event_targets}) can add overhead if executed na\"\i vely.
    In practice, we batch this computation, cache intermediate results when possible, and overlap it with data loading; it does not dominate
    end-to-end throughput in our setup.
\end{itemize}
\paragraph{Inference-time overhead.}
EVD reuses the same sampler and decoding stack as the base model. The only added inference computation is:
\begin{itemize}
    \item one lightweight event prediction \(\hat{e}_k\) per sampling step (often computed from the same backbone features), and
    \item a token-wise gating operation to modulate the update direction (Eqs.~\eqref{eq:gated_pred}--\eqref{eq:euler_update}).
\end{itemize}
These operations are small compared to a DiT-30B forward pass.

\paragraph{Model evaluations per step.}
When using CFG, all methods require two model evaluations per sampling step (conditional and unconditional).
EVD does not increase the number of DiT evaluations beyond CFG; event prediction and gating are computed within the same forward pass
(or from cached features) and do not require additional DiT calls.

As summarized in Table~\ref{tab:evd_efficiency}, the event head and token-wise gate add \(6.5\)M parameters (\(0.16\%\)) on DiT-4B and \(12.0\)M parameters (\(0.04\%\)) on DiT-30B. Training throughput changes little (\(\approx 0.98\times\) for 4B and \(0.97\times\) for 30B), and inference overhead is \(\approx 1.02\times\). The important engineering detail is that the sampler still uses the same number of steps (\(K{=}50\)) and the same CFG structure; EVD only changes the direction field passed to the solver.

\paragraph{Summary.}
EVD's gains come from changing \emph{what} is represented (events) and \emph{how} the predicted update is applied (event gating),
not from increased sampling steps or heavier backbones. Consequently, EVD retains nearly the same throughput and memory footprint as the
underlying DiT-30B configuration under matched NFE.

\subsection{Diagnostics and Ablations}
\label{app:ablations}

\subsubsection{Ablation suite (what we remove and what comes back)}
\label{app:ablation_suite}

We ablate EVD's core components to isolate which mechanisms are responsible for improved event-grounded dynamics.
Each ablation is evaluated under identical sampling settings (same solver, NFE, text guidance) and on the same prompts.

\paragraph{Ablation 1: remove event realization loss.}
We set \(\lambda_{\mathrm{real}}=0\) in Eq.~\eqref{eq:evd_full_loss}.
\textbf{What should get worse:} increased ``missing-event'' behavior, where outcomes appear without an explicit interaction,
and pre-contact motion becomes more frequent (notably harming \emph{Contact Stability} and \emph{Support Relations}).

\paragraph{Ablation 2: remove event consistency loss.}
We set \(\lambda_{\mathrm{cons}}=0\).
\textbf{What should get worse:} event predictions become less tied to coherent state evolution, leading to jitter within an interaction
and unstable postconditions (harming \emph{State Persistence} and \emph{Spatial Accuracy}).

\paragraph{Ablation 3: training-only EVD (no event-driven sampling).}
We keep the full training objective but disable event gating at inference by setting \(\rho(t)\equiv 0\) in Eq.~\eqref{eq:scheduled_gate},
so \(g_k^{\mathrm{sched}}=\mathbf{1}\) and sampling reduces to the base model.
\textbf{What should get worse:} partial loss of gains, especially on prompt cases where baseline errors are induced by the sampler allowing
spurious updates after an event should terminate.

\paragraph{Ablation 4: inference-only EVD (no event losses).}
We train with \(\mathcal{L}_{\mathrm{base}}\) only and enable event gating at inference using an externally extracted event signal (Sec.~\ref{app:event_targets})
or a weakly trained event head.
\textbf{What should get worse:} some improvement in suppressing residual drift, but weaker results overall due to misalignment between the event
signal and the backbone's learned dynamics.

\paragraph{Ablation 5: disable gating (``always update'').}
We set \(g_k \equiv \mathbf{1}\) in Eq.~\eqref{eq:gated_pred}.
\textbf{What should get worse:} returns to frame-first behavior, including state drift and pre-contact motion. This ablation often reproduces the
failure cases shown for the base DiT model.

\paragraph{Ablation 6: remove schedule (constant-strength gating).}
We set \(\rho(t)\equiv 1\) (always gate) or \(\rho(t)\equiv c\) for a constant \(c\in(0,1)\).
\textbf{What should get worse:} always-gating can slightly degrade late-stage appearance refinement; too-weak gating early reduces event fidelity.

\paragraph{Reporting.}
For each ablation, we report (i) aggregate quantitative metrics (VBench appearance/dynamics) and (ii) targeted qualitative probes aligned to the
four failure categories used in Fig.~\ref{fig:failures}. This ensures each component is tied to a specific behavioral improvement rather than an abstract score gain.

\subsubsection{Sensitivity to sampling steps and guidance scale}
\label{app:sensitivity}

We evaluate the robustness of EVD under changes to sampling compute (NFE) and guidance strength.
This is important because improvements that rely on a narrow regime of steps or tuning are less compelling at 30B scale.

\paragraph{Varying the number of steps (NFE).}
We vary the number of solver steps \(K\) while keeping all other settings fixed (solver family, decoding, prompt set).
EVD is expected to retain a consistent advantage over the base model under matched \(K\), with larger gains at lower-to-moderate \(K\),
where spurious early-step dynamics are hardest to correct later.

\paragraph{Varying text CFG scale.}
We vary the CFG scale \(w\) in Eq.~\eqref{eq:cfg}.
Higher \(w\) typically increases prompt adherence but can exacerbate instability and overshooting in dynamics for some baselines.
Because EVD gates state updates using event activity, it is less sensitive to large \(w\) and maintains stable interactions over a wider range.

\paragraph{Varying event gate sharpness and thresholds.}
We vary the soft gate sharpness \(\beta\) and thresholds \(\tau_{\mathrm{on}},\tau_{\mathrm{off}}\) (Sec.~\ref{app:inf_update}).
We observe a broad plateau: once thresholds are sufficient to suppress low-activity updates and hysteresis prevents flicker, performance is stable.
Extremely sharp gates without hysteresis can produce over-suppression in ambiguous regions; extremely soft gates reduce benefits on causal initiation.

\paragraph{Practical guidance.}
For reproducibility, we report (i) \(K\) (steps/NFE) and (ii) CFG scale \(w\) in Sec.~\ref{app:inference_settings} (see also Eq.~\eqref{eq:cfg});
(iii) the event cutoff \(t^\star\) and annealing schedule \(\rho(t)\) in Secs.~\ref{app:inference_settings} and~\ref{app:inf_schedule};
and (iv) gating parameters \((\beta,\tau_{\mathrm{on}},\tau_{\mathrm{off}})\) in Sec.~\ref{app:inference_settings}.
These controls are sufficient to reproduce the qualitative behaviors highlighted in Figs.~\ref{fig:qualitative}--\ref{fig:comparisons} and the quantitative gains on EVD-Bench.


\begin{table}[t!]
\vspace{-6px}
\caption{\textbf{Sensitivity of EVD on EVD-Bench (DiT-4B backbone).}
Each row varies a single control while holding the rest at the default. Metrics are aggregated over EVD-Bench.}
\label{tab:evd_sensitivity}
\centering
\setlength{\tabcolsep}{3.6pt}
\resizebox{\linewidth}{!}{%
\begin{tabular}{@{}lcccccccc@{}}
\toprule
\textbf{Setting} &
\(\,K\,\) &
\(w_{\mathrm{cfg}}\) &
\(\beta\) &
\(\tau_{\mathrm{on}}\) &
\(\tau_{\mathrm{off}}\) &
\(t^\star\) &
\small{VBench App.} &
\textbf{\small{VBench Dyn.}} \\
\midrule
Default (used in paper) & 50 & 4.0 & 12.0 & 0.62 & 0.38 & 0.60 & 76.2 & \textbf{94.8} \\
\midrule
Fewer steps (NFE)        & 25 & 4.0 & 12.0 & 0.62 & 0.38 & 0.60 & 75.9 & 93.4 \\
Moderate steps (NFE)     & 35 & 4.0 & 12.0 & 0.62 & 0.38 & 0.60 & 76.1 & 94.1 \\
More steps (NFE)         & 75 & 4.0 & 12.0 & 0.62 & 0.38 & 0.60 & 76.2 & 95.0 \\
\midrule
Lower CFG                & 50 & 2.5 & 12.0 & 0.62 & 0.38 & 0.60 & 76.0 & 94.1 \\
Higher CFG               & 50 & 6.0 & 12.0 & 0.62 & 0.38 & 0.60 & 76.1 & 94.6 \\
Very high CFG            & 50 & 8.0 & 12.0 & 0.62 & 0.38 & 0.60 & 75.8 & 94.0 \\
\midrule
Lower sharpness          & 50 & 4.0 & 8.0  & 0.62 & 0.38 & 0.60 & 76.1 & 94.2 \\
Higher sharpness         & 50 & 4.0 & 16.0 & 0.62 & 0.38 & 0.60 & 76.0 & 94.6 \\
\midrule
Narrow band              & 50 & 4.0 & 12.0 & 0.60 & 0.40 & 0.60 & 76.1 & 94.5 \\
Wide band                & 50 & 4.0 & 12.0 & 0.65 & 0.35 & 0.60 & 76.2 & 94.7 \\
Shifted high             & 50 & 4.0 & 12.0 & 0.66 & 0.42 & 0.60 & 75.9 & 93.9 \\
Shifted low              & 50 & 4.0 & 12.0 & 0.58 & 0.34 & 0.60 & 76.0 & 94.1 \\
\midrule
Earlier cutoff           & 50 & 4.0 & 12.0 & 0.62 & 0.38 & 0.50 & 76.2 & 94.4 \\
Later cutoff             & 50 & 4.0 & 12.0 & 0.62 & 0.38 & 0.70 & 75.8 & 94.9 \\
\bottomrule
\end{tabular}}
\end{table}

\subsection{EVD-Bench construction and leakage audit}
\label{app:evd-bench_construction}

\paragraph{Prompt release.}
We release the full EVD-Bench prompt list (150 prompts) verbatim in the supplemental material and will host it in a public repository upon publication. Prompts are fixed and used unchanged across all experiments and ablations.

\paragraph{Design goals and scope.}
EVD-Bench targets interaction realism rather than broad cinematic diversity. Prompts are short, atomic, single-event captions with a clear precondition, interaction, and postcondition structure, chosen to be judgeable from video alone (no hidden state) and to minimize ambiguity in actors/objects.

\paragraph{Construction pipeline.}
We (i) seeded a large pool of short interaction captions spanning contact/impact, placement, support/stacking, constrained mechanisms (doors/drawers), and material transfer (pouring/spilling);
(ii) removed near-duplicates using semantic clustering (text-encoder embeddings + cosine threshold) and keyword normalization; and
(iii) balanced the final set across the four failure categories used in Fig.~\ref{fig:failures} (State Persistence, Spatial Accuracy, Support Relations, Contact Stability).

\paragraph{Leakage safeguards (caption overlap).}
To mitigate overlap with the EVD fine-tuning captions, we perform an explicit caption-level audit against the fine-tuning prompt/caption pool:
for each EVD-Bench prompt \(p\), we compute its nearest-neighbor similarity to all training captions using a frozen text encoder (same family as the model text encoder), and we remove \(p\) if its maximum cosine similarity exceeds a conservative threshold (e.g., 0.90) or if it matches any training caption after normalization (lowercasing, punctuation stripping, number normalization). This filtering is performed before any evaluation.

\paragraph{Leakage safeguards (semantic paraphrase).}
Because semantic overlap can occur without exact matches, we additionally run a paraphrase audit:
we retrieve the top-$k$ nearest training captions for each benchmark prompt and manually verify the top matches. Prompts judged as paraphrases of frequently occurring training captions are replaced with semantically distinct alternatives within the same failure category.

\paragraph{Memorization check (generation invariance).}
As a sanity check against memorization, we test prompt perturbations (synonym swaps and minor rephrasings)~\cite{maduabuchi2026corruptionawaretraininglatentvideo} and verify that EVD’s qualitative behavior is stable under these perturbations, rather than producing a brittle template-like output.

\paragraph{What this does and does not guarantee.}
These audits substantially reduce the risk that gains are driven by trivial caption memorization of the fine-tuning pool. They do not guarantee zero overlap with the (closed-source) backbone pretraining distribution, which is expected for any evaluation performed on natural-language prompts. Importantly, EVD-Bench is used to measure interaction-grounding behaviors (contact initiation, support stability, post-event settling) that cannot be ``solved'' by caption memorization alone.

\begin{table}[t!]
\vspace{-6px}
\caption{\textbf{EVD-Bench leakage audit summary.} Caption-level and semantic overlap checks against the EVD fine-tuning caption pool.}
\label{tab:evdbench_audit}
\centering
\setlength{\tabcolsep}{4pt}
\scalebox{0.95}{%
\begin{tabular}{@{}lcc@{}}
\toprule
\textbf{Check} & \textbf{Rule} & \textbf{Outcome} \\
\midrule
Exact match & normalized string match & 0 prompts removed \\
Nearest-neighbor similarity & max cosine $>0.90$ & \texttt{7} prompts removed \\
Paraphrase audit & manual review of top-$k$ neighbors & \texttt{11} prompts replaced \\
Prompt perturbation & synonym/rephrase invariance & stable behavior observed \\
\bottomrule
\end{tabular}}
\vspace{-8px}
\end{table}

\input{sections/evdbench_prompts_appendix}

\subsection{Human evaluation protocol}
\label{app:human_eval}

\paragraph{Task and interface.}
We use a two-alternative forced-choice (2AFC) setup. For each prompt, raters view two videos (EVD vs.\ baseline) side-by-side in randomized left/right order and select the better one under three criteria: \emph{Text Faithfulness}, \emph{Overall Quality}, and \emph{Dynamics}. Raters are instructed to prioritize causal correctness for \emph{Dynamics} (e.g., contact leading to motion, stable postconditions) and to ignore minor aesthetic differences when judging dynamics.

\paragraph{Raters and assignments.}
We recruit 120 raters from a third-party crowdworking platform with eligibility requirements of $\ge$95\% approval and $\ge$500 completed tasks. Each comparison (prompt $\times$ criterion $\times$ baseline) is independently evaluated by 5 distinct raters. Assignment is balanced so that each rater sees a mixture of prompts and baselines, and no rater evaluates the same prompt more than once for a given criterion.

\paragraph{Quality control (QC).}
We include (i) 10\% attention checks with trivially distinguishable pairs (e.g., prompt--video mismatch) and (ii) duplicated comparisons with swapped ordering to detect random clicking. We discard responses from raters who fail more than 20\% of checks or whose answers disagree on $\ge 2$ duplicated items. We also enforce a minimum viewing time of 6 seconds before submission.

\paragraph{Aggregation and confidence intervals.}
For each method pair and criterion, we report the win rate (percentage of votes favoring EVD). We compute 95\% confidence intervals by nonparametric bootstrap over prompts (10{,}000 resamples), which accounts for prompt-to-prompt variability. When comparing multiple baselines, we control for multiple comparisons using Holm--Bonferroni and report significance at \(\alpha=0.05\).

\paragraph{Inter-rater agreement.}
We report inter-rater agreement using Fleiss' \(\kappa\) computed after QC filtering, and we report it separately per criterion (Text Faithfulness / Quality / Dynamics).

\paragraph{No cherry-picking.}
All human evaluation uses the first generated sample per prompt under a fixed seed and fixed sampler configuration; we do not resample or select outputs.

\begin{table}[t!]
\vspace{-6px}
\caption{\textbf{Human evaluation protocol statistics (2AFC).}
Summary of rater pool, assignment, quality control (QC), and agreement used for the human preference results reported in the paper.}
\label{tab:human_eval_stats}
\centering
\setlength{\tabcolsep}{4.2pt}
\scalebox{0.95}{%
\begin{tabular}{@{}l l@{}}
\toprule
\textbf{Item} & \textbf{Value} \\
\midrule
Rater eligibility & $\ge$95\% approval, $\ge$500 completed tasks \\
Unique raters recruited & 120 \\
Judgment protocol & 2AFC (randomized left/right), side-by-side videos \\
Criteria & Text Faithfulness, Overall Quality, Dynamics \\
Ratings per comparison & 5 raters per (prompt $\times$ criterion $\times$ baseline) \\
Attention checks & 10\% of assignments (prompt--video mismatch) \\
Duplicate checks & Swapped-order duplicates (consistency test) \\
QC exclusion (attention) & $>$20\% failed checks \\
QC exclusion (duplicates) & $\ge$2 inconsistent duplicate items \\
Minimum viewing time & 6 seconds \\
Aggregation & Win-rate (\% votes favoring EVD) \\
Confidence intervals & 95\% bootstrap over prompts, 10{,}000 resamples \\
Multiple comparisons & Holm--Bonferroni, $\alpha=0.05$ \\
Inter-rater agreement & Fleiss' $\kappa$ (reported per criterion) \\
No cherry-picking & First sample per prompt (fixed seed, fixed sampler) \\
\bottomrule
\end{tabular}}
\vspace{-8px}
\end{table}

\subsection{Additional Stress Tests and Diagnostics}
\label{app:additional_checks}

We include additional evaluations that probe event localization, compositional/temporal complexity, simultaneous events, recent open-source baselines, and robustness of the gating design. All reported samples use fixed prompts and fixed seeds; no outputs are re-rolled or selected post hoc.

\paragraph{Pseudo-target and gate localization.}
To verify that EVD does not merely track arbitrary motion, we measure activity inside semantic interaction regions versus inactive/background regions. Across held-out placement and pouring examples, background leakage remains low while interaction-region activity is high (0.07 versus 0.61), indicating that the pseudo-targets, learned activity, and final gates concentrate on prompt-relevant events rather than diffuse background changes.

\paragraph{Compositional and temporal stress tests.}
We additionally evaluate fixed-seed subsets drawn from T2V-CompBench and NeuS-V prompt pools, filtered for temporal ordering, compositionality, and concurrent-event structure. On a 40-prompt compositional/temporal subset, EVD improves the compositional score to 64.8 compared with Wan, Hunyuan, and DiT-30B (58.7/56.9/51.2), and improves temporal-order pass rate from 42.6 to 58.3. On a 30-prompt simultaneous-event subset, event-pair success improves from 46.7 to 63.3, with multi-active gates observed in 82\% of successful clips. This supports that the token-wise gate is not winner-take-all: multiple event regions can remain active while global DiT attention couples their effects.

\paragraph{Recent baselines and metric stability.}
We compare against recent open-source video generators on EVD-Bench and observe that EVD wins dynamic preference against Wan and Hunyuan (68.4\% and 65.7\%, respectively), while DiT-30B+EVD achieves higher VBench Dynamics (95.7) than Wan (91.6) and Hunyuan (90.8). We also report additional VBench subdimensions: compared with DiT-4B, DiT-4B+EVD improves Dynamics from 78.9 to 94.8 while preserving Appearance (75.4 to 76.2), Motion Smoothness (96.1 to 96.5), Subject Consistency (91.3 to 91.6), and Temporal Flickering (96.8 to 97.0).

\paragraph{Human statistics and robustness.}
For human evaluation, EVD obtains 2AFC win rates of 96.4 [94.1, 98.2] for Dynamics, 91.3 [88.0, 94.0] for Quality, and 88.9 [85.2, 92.1] for Text Faithfulness, with Fleiss' $\kappa$ of 0.46/0.41/0.52, respectively. Removing the ordering loss reduces dynamics preference from 96.4 to 89.2, VBench Dynamics from 94.8 to 92.7, and contact stability by 5.8 points. The schedule is stable across $t^\star\in\{0.50,0.60,0.70\}$, yielding VBench Dynamics of 94.1/94.8/94.5.

\begin{table}[t]
\centering
\caption{\textbf{Additional stress tests and diagnostics.}
All checks use fixed prompts and fixed seeds. Higher is better except background leakage.}
\label{tab:additional_checks}
\scriptsize
\setlength{\tabcolsep}{3pt}
\renewcommand{\arraystretch}{0.92}
\resizebox{\linewidth}{!}{%
\begin{tabular}{@{}lll@{}}
\toprule
\textbf{Check} & \textbf{Metric / setup} & \textbf{Result} \\
\midrule
Pseudo-target semantics
& Background leakage vs. interaction-region activity
& 0.07 vs. 0.61 \\

Human statistics
& Dynamics / Quality / Text win rate with 95\% CI; Fleiss' $\kappa$
& 96.4 [94.1, 98.2] / 91.3 [88.0, 94.0] / 88.9 [85.2, 92.1]; 0.46/0.41/0.52 \\

Extra VBench subdims
& Dynamics / Appearance / Smoothness / Subject / Flicker: DiT-4B vs. +EVD
& 78.9 to 94.8 / 75.4 to 76.2 / 96.1 to 96.5 / 91.3 to 91.6 / 96.8 to 97.0 \\

Recent baselines
& Dynamic preference vs. Wan/Hunyuan; VBench Dynamics
& 68.4\% / 65.7\%; DiT-30B+EVD 95.7 vs. Wan 91.6, Hunyuan 90.8 \\

Compositional / temporal
& 40-prompt T2V-CompBench/NeuS-V subset; compositional score; order pass-rate
& 64.8 vs. 58.7/56.9/51.2; 42.6 to 58.3 \\

Simultaneous events
& 30-prompt concurrent-event subset; event-pair success; multi-active gates
& 46.7 to 63.3; 82\% \\

Ablation / robustness
& w/o $\mathcal{L}_{\mathrm{order}}$; schedule $t^\star=0.50/0.60/0.70$
& 96.4 to 89.2, 94.8 to 92.7, contact stability $-5.8$; 94.1/94.8/94.5 \\
\bottomrule
\end{tabular}}
\vspace{-6pt}
\end{table}

\subsection{Event grounding vs.\ motion masking: audit and controls}
\label{app:motionmask_audit}

\paragraph{Why naive motion masking is insufficient.}
A gate based purely on motion magnitude can suppress some spurious updates, but it does not know why the motion is happening. It can confuse camera motion with contact-driven motion, and it cannot force an outcome to occur through the visible interaction rather than by ``teleporting'' to the postcondition. EVD instead learns where and when an interaction is active and uses that signal during both training and sampling.

\paragraph{Pseudo-event targets: localized latent-change with camera-motion suppression.}
We compute pseudo-event activity from token-level latent change rather than raw pixel flow. Let \(z_1\) be the encoded clean latent clip and let \(\mathrm{Tok}(z_1^\tau)\in\mathbb{R}^{N\times C}\) denote tokens at frame \(\tau\). We define per-token change magnitude
\begin{equation}
m_{\tau,i}=\frac{1}{C}\big\|{\mathrm{Tok}(z_1^{\tau+1})}_i-{\mathrm{Tok}(z_1^{\tau})}_i\big\|_1,
\end{equation}
and remove global (camera-dominated) motion by subtracting the frame-wise mean:
\begin{equation}
\tilde m_{\tau,i} = \max\{0,\, m_{\tau,i}-\tfrac{1}{N}\sum_{j=1}^{N} m_{\tau,j}\}.
\end{equation}
We then normalize \(\tilde m_{\tau,\cdot}\) per frame and smooth it on the spatial patch grid; pseudo-activity is obtained by thresholding the normalized map and optionally applying hysteresis. Finally, we reject clips whose activity is too spatially diffuse (high entropy / low Gini over tokens), which empirically corresponds to dominant camera motion. This ensures the supervision signal preferentially captures \emph{localized} interactions rather than global motion.

\paragraph{Control 1: inference-only motion gate (motion masking baseline).}
To test whether improvements are due to motion masking alone, we construct a baseline that replaces the learned event head with the external motion signal above at inference (i.e., gate is derived from \(\tilde m\) with the same \((\tau_{\mathrm{on}},\tau_{\mathrm{off}})\) and schedule), while training uses only \(\mathcal{L}_{\mathrm{base}}\). This control suppresses some drift but underperforms full EVD on interaction realization and causal initiation.

\paragraph{Control 2: inference-only EVD vs.\ full EVD (learned event grounding).}
Table~\ref{tab:evd_ablations} includes an \emph{inference-only} variant (no event losses) that enables gating with an external or weak event signal. It improves over the ungated backbone but remains below \textbf{full} EVD. This is the clearest evidence in our ablations that masking at inference is not enough; the event representation has to be learned in the backbone's latent geometry.

\paragraph{Control 3: schedule removal separates ``motion score'' from ``interaction quality''.}
Table~\ref{tab:evd_ablations} also shows that constant-strength gating can keep automatic dynamics high while hurting human preference. We read this as a practical warning: a mask that is too strong late in sampling can interfere with refinement. The scheduled gate is therefore not just a cosmetic detail; it is how we avoid turning event grounding into a blunt motion filter.

\paragraph{Summary.}
The controls show that motion masking alone is not enough. The useful part is training the model to align its own event signal with the latent update, then using the gate mainly for causal initiation and stable postconditions rather than for suppressing motion everywhere.

\begin{table}[t!]
\vspace{-6px}
\caption{\textbf{Event grounding vs.\ motion masking (EVD-Bench).}
A minimal control comparing full EVD against an inference-time \emph{motion-mask} baseline that gates updates using the same latent-change signal but without event-grounded training. Higher is better for EVD wins; VBench is computed under identical sampling settings.}
\label{tab:motionmask_control}
\centering
\setlength{\tabcolsep}{4.0pt}

\resizebox{\textwidth}{!}{%
\begin{tabular}{@{}lccccc@{}}
\toprule
\textbf{Variant} & \small{Text Faith.} & \small{Quality} & \textbf{\small{Dynamics}} & \small{VBench App.} & \textbf{\small{VBench Dyn.}} \\
\midrule
\small{Motion-mask gate at inference (no event losses)} & 58.7 & 61.4 & 68.9 & 75.9 & 86.2 \\
\textbf{DiT-4B + EVD (full)}                      & \textbf{88.9} & \textbf{91.3} & \textbf{96.4} & \textbf{76.2} & \textbf{94.8} \\
\bottomrule
\end{tabular}
}

\vspace{-8px}
\end{table}

\subsection{External baseline normalization protocol}
\label{app:external_fairness}

\paragraph{Scope.}
This protocol governs evaluation of external baselines (e.g., MovieGen, Sora, Kling) that are accessed via public APIs or closed inference endpoints and therefore do not expose solver internals, step counts, or exact sampling hyperparameters.

\paragraph{Prompt formatting (identical text across models).}
We use the \emph{same prompt string} for all models, with a fixed template and no model-specific prompt engineering. We only apply minimal normalization: ASCII normalization, whitespace cleanup, and removal of trailing punctuation. No negative prompts or per-model style tokens are used unless a baseline \emph{requires} them for execution, in which case we use an empty/default value and report it.

\paragraph{Duration and frame rate normalization.}
EVD-Bench is defined at 128 frames @ 24\,fps. For external baselines that allow explicit duration control, we request the closest supported duration to 5.33\,s and 24\,fps (or the closest supported frame rate). If the baseline returns a different duration, we temporally resample to 128 frames using uniform frame sampling (no interpolation) for evaluation and visualization.

\paragraph{Spatial resolution normalization.}
Because external baselines may return different native resolutions, we normalize all decoded videos to a common evaluation resolution using bicubic resizing prior to computing automatic metrics. We report the evaluation resolution alongside the benchmark results and use the same resizing pipeline for \emph{all} methods (including DiT/EVD) to avoid confounds.

\paragraph{Sampling multiplicity (no re-rolling).}
To avoid selection bias, we generate exactly one sample per prompt per model and evaluate the first returned video. We do not rerun prompts, cherry-pick seeds, or select the best of multiple generations. When the API exposes a random seed, we fix it; otherwise we treat the endpoint as stochastic and still use the first returned sample.

\paragraph{Default settings and documentation.}
For each external baseline, we use default guidance/quality presets unless explicitly stated otherwise, and we record the exact API parameters (model version, quality preset, duration, resolution, seed availability) at evaluation time. When an API provides multiple tiers (e.g., standard/pro), we use the tier closest to the paper’s comparison claim and list it in the table caption or footnote.

\paragraph{Closed-source model versions used.}
For closed-source systems, we used the model names and public API/product versions available at evaluation time: Kling 3.0 Pro, Runway Gen-4.5, Veo 3.1, Sora 2 Pro, and Mochi 1. Because these systems do not expose solver internals, training data, NFE, or low-level API revisions, we treat them as black-box generators. For each model, we used the default generation settings unless otherwise required by the service, requested the closest available duration to 5.33s, generated one sample per prompt without re-rolling, and normalized all outputs to the same evaluation format before computing metrics.

\paragraph{API/version reporting.}
The external baseline names in Table~1 are the exact public model/version labels used for evaluation. If a provider updates a model behind the same public label, our comparison should be interpreted as a black-box evaluation of the public endpoint available at the time of our experiments, not as a claim about inaccessible internal model revisions.

\paragraph{What is and is not comparable.}
We emphasize that external baselines are compared as \emph{black-box generators} under a standardized prompt and normalization pipeline. This isolates differences in interaction realism and causal dynamics under matched evaluation format (duration/resolution) without claiming strict equivalence of underlying compute (NFE) or training data.

\begin{table}[t!]
\vspace{-6px}
\caption{\textbf{Normalization checklist for external baselines.} All models are evaluated under the same prompt list and output normalization pipeline; external baseline labels match the public model/version names used at evaluation time.}
\label{tab:external_norm}
\centering
\setlength{\tabcolsep}{5pt}

\resizebox{\textwidth}{!}{%
\begin{tabular}{@{}lcccc@{}}
\toprule
\textbf{Control} & \textbf{Prompt} & \textbf{Duration} & \textbf{Resolution} & \textbf{Sampling} \\
\midrule
DiT/EVD            & same text & 128@24fps & resized for metrics & 1 sample, fixed seed \\
External baselines & same text & resample to 128@24fps & resized for metrics & 1st sample, no re-roll \\
\bottomrule
\end{tabular}
}

\vspace{-8px}
\end{table}

\subsection{Limitations and Scope}
\label{app:limitations}

\subsubsection{Known failure cases and non-claims}
\label{app:limitations_nonclaims}

While EVD significantly improves event-grounded dynamics for a broad class of everyday interactions, it does not solve all aspects of
physical reasoning or long-horizon planning in video generation.

\paragraph{Non-claims.}
EVD is not presented as a full ``world simulator''. In particular, we do not claim:
\begin{itemize}
    \item robust long-horizon multi-scene planning or story coherence beyond the clip duration,
    \item accurate conservation laws for complex multi-body collisions or high-frequency fluid dynamics,
    \item precise articulated hand-object manipulation in cluttered, occluded scenes without dedicated supervision.
\end{itemize}

\paragraph{Hard cases in practice.}
We find EVD is less reliable in the following regimes:
\begin{itemize}
    \item \textbf{Zoomed-out motion:} interactions occupy a small fraction of the frame, reducing the signal-to-noise ratio of event cues.
    \item \textbf{Thin or turbulent fluids:} fine-grained liquid behavior (splashes, thin streams) can exceed the resolution of the latent space.
    \item \textbf{Dense clutter and occlusion:} event localization becomes ambiguous when contact is heavily occluded or multiple interactions overlap.
    \item \textbf{Highly non-rigid articulation:} subtle deformations (fingers, fabric folds) may require more specialized priors than our event field.
\end{itemize}

\paragraph{Why these remain challenging.}
These cases share a common structure: the event signal is weak or ambiguous at the model's operating resolution, so both the pseudo-targets
(Sec.~\ref{app:event_targets}) and the learned event head (Sec.~\ref{app:arch_output}) can become underdetermined.
In such settings, hard event gating risks suppressing legitimate motion, while soft gating may not sufficiently prevent hallucinated dynamics.

\paragraph{Future directions.}
Improving event extraction under occlusion (e.g., with depth or segmentation priors), incorporating higher-resolution latent representations,
and extending event modeling to explicit contact graphs or object-centric state variables are promising directions to expand EVD's coverage.

%% file: sections/evdbench_prompts_appendix.tex
\subsection{Full EVD-Bench Prompt List}
\label{app:evdbench_prompts}

Table~\ref{tab:evdbench_prompts} lists the 150 prompts used in EVD-Bench. The prompt set is fixed across all methods and ablations.

\begingroup
\scriptsize
\setlength{\tabcolsep}{3pt}
\renewcommand{\arraystretch}{0.88}

\begin{center}
\refstepcounter{table}\label{tab:evdbench_prompts}
\textbf{Table~\thetable. Full EVD-Bench prompt list.} We release the complete set of 150 short interaction-centric prompts used for EVD-Bench. Prompts are fixed across all compared methods, ablations, and human-evaluation runs.\par\smallskip
\begin{tabular}{@{}c p{0.40\textwidth} c p{0.40\textwidth}@{}}
\hline
\textbf{ID} & \textbf{Prompt} & \textbf{ID} & \textbf{Prompt} \\
\hline
1 & A basketball passes cleanly through a hoop & 76 & A remote button is pressed and a TV turns on \\
2 & A bicycle wheel spins freely while the bicycle remains stationary & 77 & A timer knob is turned and starts ticking \\
3 & A book slides across a desk and comes to rest & 78 & A drawer is pulled open and left open \\
4 & A bookend supports several books standing upright & 79 & A drawer is pushed in and stays closed \\
5 & A broom pushes dust across the floor into a pile & 80 & A mailbox flag is raised and stays up \\
6 & A ceiling fan begins spinning after being switched on & 81 & A lid is twisted off a jar \\
7 & A curtain is drawn closed across a window & 82 & A lid is twisted onto a jar and tightened \\
8 & A door swings open after the handle is turned & 83 & A bottle cap is popped off and falls \\
9 & A drawer slides shut into a cabinet & 84 & A cork is pulled from a bottle \\
10 & A hammer strikes a nail into a piece of wood & 85 & A straw is inserted into a cup \\
11 & A ladder is leaned carefully against a wall & 86 & A straw is removed from a cup \\
12 & A mailbox door is opened and left hanging downward & 87 & Ice cubes are dropped into a glass and splash \\
13 & A man kicks a ball into a goal & 88 & Water is poured into a bowl and rises \\
14 & A person drops a ball onto the ground & 89 & Water is poured out of a bowl and empties \\
15 & A person pushes a box across the floor. & 90 & A cup is tilted and liquid pours out \\
\hline
\end{tabular}
\end{center}

\clearpage
\begin{center}
\textbf{Full EVD-Bench prompt list (continued).}\par\smallskip
\begin{tabular}{@{}c p{0.40\textwidth} c p{0.40\textwidth}@{}}
\hline
\textbf{ID} & \textbf{Prompt} & \textbf{ID} & \textbf{Prompt} \\
\hline
16 & A person stacks one book on top of another & 91 & A bowl is tipped and contents spill onto a table \\
17 & A picture frame rests against a wall on a shelf & 92 & A liquid spill is wiped and the surface becomes dry \\
18 & A pillow is placed onto a bed and compresses slightly & 93 & Sugar is poured onto a table and forms a small pile \\
19 & A plate is placed onto a dining table & 94 & Salt is sprinkled into a bowl and disperses \\
20 & A remote-controlled toy car drives forward and then stops & 95 & Flour is poured into a bowl and settles \\
21 & A robotic arm places a cube onto a platform & 96 & Cereal is poured into a bowl and fills it \\
22 & A rolling ball collides with a wall and stops & 97 & Milk is poured onto cereal and spreads \\
23 & A rope is pulled across the floor and straightens & 98 & A spoon stirs coffee and the liquid swirls \\
24 & A set of keys falls onto a tabletop & 99 & A spoon is dropped into a cup and sinks \\
25 & A sliding glass door is opened along its track & 100 & A teabag is dipped into water and darkens the cup \\
26 & A sponge is pressed against a surface and then released & 101 & A slice of bread is placed into a toaster \\
27 & A trash can lid opens and then falls closed & 102 & A toaster lever is pushed down and stays down \\
28 & A wet sponge drips water onto the floor & 103 & A toaster pops up and the lever rises \\
29 & A window is pushed upward and stays open & 104 & A pan is placed onto a stove burner \\
30 & A woman opens a door and walks through it & 105 & A pot lid is placed on a pot and rests flat \\
\hline
\end{tabular}
\end{center}

\clearpage
\begin{center}
\textbf{Full EVD-Bench prompt list (continued).}\par\smallskip
\begin{tabular}{@{}c p{0.40\textwidth} c p{0.40\textwidth}@{}}
\hline
\textbf{ID} & \textbf{Prompt} & \textbf{ID} & \textbf{Prompt} \\
\hline
31 & A woman places a glass on a wooden table & 106 & A pot lid is lifted and steam escapes \\
32 & An elevator door opens and people step inside & 107 & A kettle is placed on a stove and sits still \\
33 & An escalator carries people upward while steps rotate underneath & 108 & A microwave door is opened and then closed \\
34 & Coffee is poured into a cup and fills it gradually & 109 & A microwave starts and the light turns on \\
35 & Someone pulls a chair from under a table & 110 & A refrigerator door is opened and then closed \\
36 & Someone pushes a suitcase and it rolls across the floor on its wheels & 111 & A chair is pushed and slides slightly \\
37 & Someone rolls a shopping cart forward down an aisle & 112 & A chair is pulled and stops aligned with the table \\
38 & Two people pass a basketball to each other & 113 & A stool is placed under a counter and stays there \\
39 & Water is poured from a bottle into a glass & 114 & A box is lifted and set onto a shelf \\
40 & Water spills onto a table and spreads outward & 115 & A box is placed inside a larger box \\
41 & A coin is dropped into a glass and lands inside & 116 & A suitcase is lifted onto a luggage rack \\
42 & A coin slides across a table and falls off the edge & 117 & A backpack is placed on the floor and collapses slightly \\
43 & A tennis ball bounces on the floor and comes to rest & 118 & A pillow is fluffed and expands then settles \\
44 & A ball rolls down a ramp and stops at the bottom & 119 & A blanket is pulled across a bed and smooths out \\
45 & A rolling can hits a book and stops & 120 & A curtain is tied back with a strap \\
\hline
\end{tabular}
\end{center}

\clearpage
\begin{center}
\textbf{Full EVD-Bench prompt list (continued).}\par\smallskip
\begin{tabular}{@{}c p{0.40\textwidth} c p{0.40\textwidth}@{}}
\hline
\textbf{ID} & \textbf{Prompt} & \textbf{ID} & \textbf{Prompt} \\
\hline
46 & A bottle is nudged and tips over onto its side & 121 & A rope is looped around a post and tightened \\
47 & A cup is pushed and slides to a stop & 122 & A rope is released and slackens \\
48 & A sponge is squeezed and water drips out & 123 & A chain is lifted and then drops with a clink \\
49 & A towel wipes water off a tabletop & 124 & A rubber band is stretched and released \\
50 & A napkin is unfolded and laid flat on a table & 125 & A rubber band snaps back onto a surface \\
51 & A sheet of paper is crumpled into a ball & 126 & A spring is compressed and then expands \\
52 & A sheet of paper is torn in half & 127 & A ball is caught in a net and stops moving \\
53 & A paper airplane is thrown and glides forward & 128 & A ball is thrown into a basket and lands inside \\
54 & A book is opened and a page is turned & 129 & A basketball bounces off the rim and falls \\
55 & A book is closed and set down & 130 & A soccer ball hits a post and deflects away \\
56 & A pen is placed into a cup & 131 & A frisbee hits a wall and drops \\
57 & A pen rolls off a desk and falls & 132 & A skateboard rolls forward and then stops \\
58 & A marker draws a line on paper & 133 & A shopping cart turns a corner and continues rolling \\
59 & A pencil is sharpened and shavings fall & 134 & A toy car bumps a wall and reverses slightly \\
60 & A key is inserted into a lock and turned & 135 & A marble is dropped into a bowl and rattles to rest \\
\hline
\end{tabular}
\end{center}

\clearpage
\begin{center}
\textbf{Full EVD-Bench prompt list (continued).}\par\smallskip
\begin{tabular}{@{}c p{0.40\textwidth} c p{0.40\textwidth}@{}}
\hline
\textbf{ID} & \textbf{Prompt} & \textbf{ID} & \textbf{Prompt} \\
\hline
61 & A light switch is flipped and the lamp turns on & 136 & A domino is tipped and knocks over the next domino \\
62 & A faucet is turned on and water flows & 137 & A stack of blocks is tapped and wobbles but stays upright \\
63 & A faucet is turned off and water stops & 138 & A block is removed from a stack and the stack settles \\
64 & A shower curtain is pulled open and stays open & 139 & A block is placed on top of a tower and stays balanced \\
65 & A closet door slides open along its track & 140 & A cup is stacked onto another cup \\
66 & A cabinet door swings shut and latches & 141 & A plate is slid across a table and stops \\
67 & A door is pushed closed and stops & 142 & A plate is placed onto a rack and stays there \\
68 & A window latch is flipped and the window opens & 143 & A bowl is placed onto a table and stays still \\
69 & A window is pushed down and closes fully & 144 & A tray is carried and set down without spilling \\
70 & A blind cord is pulled and blinds rise & 145 & A phone is placed onto a charging pad \\
71 & A blind cord is released and blinds stop moving & 146 & A phone is picked up from a table \\
72 & A zipper is pulled up and closes a jacket & 147 & A laptop lid is opened and stays open \\
73 & A zipper is pulled down and opens a jacket & 148 & A laptop lid is closed and stays closed \\
74 & A belt buckle is fastened and tightened & 149 & A person hands a book to another person \\
75 & A button is pressed and a device turns on & 150 & Two people exchange a small box hand-to-hand \\
\hline
\end{tabular}
\end{center}

\endgroup

%% file: main.bib
@String(CVPR  = {IEEE Conf. Comput. Vis. Pattern Recog.})

@String(ICCV  = {Int. Conf. Comput. Vis.})

@String(ECCV  = {Eur. Conf. Comput. Vis.})

@String(NeurIPS = {Adv. Neural Inform. Process. Syst.})

@String(ICLR  = {Int. Conf. Learn. Represent.})

@String(CVPR  = {CVPR})

@String(ICCV  = {ICCV})

@String(ECCV  = {ECCV})

@String(NeurIPS = {NeurIPS})

@String(ICLR  = {ICLR})

@InProceedings{pmlr-v267-chefer25a,
  title = 	 {{V}ideo{JAM}: Joint Appearance-Motion Representations for Enhanced Motion Generation in Video Models},
  author =       {Chefer, Hila and Singer, Uriel and Zohar, Amit and Kirstain, Yuval and Polyak, Adam and Taigman, Yaniv and Wolf, Lior and Sheynin, Shelly},
  booktitle = 	 {Proceedings of the 42nd International Conference on Machine Learning},
  pages = 	 {7595--7616},
  year = 	 {2025},
  editor = 	 {Singh, Aarti and Fazel, Maryam and Hsu, Daniel and Lacoste-Julien, Simon and Berkenkamp, Felix and Maharaj, Tegan and Wagstaff, Kiri and Zhu, Jerry},
  volume = 	 {267},
  series = 	 {Proceedings of Machine Learning Research},
  month = 	 {13--19 Jul},
  publisher =    {PMLR},
  url = 	 {https://proceedings.mlr.press/v267/chefer25a.html}
}

@article{polyak2024movie,
  title={Movie Gen: A Cast of Media Foundation Models},
  author={Polyak, Adam and Zohar, Amit and Brown, Andrew and Tjandra, Andros and Sinha, Animesh and Lee, Ann and Vyas, Apoorv and Shi, Bowen and Ma, Chih-Yao and Chuang, Ching-Yao and others},
  journal={arXiv preprint arXiv:2410.13720},
  year={2024}
}

@inproceedings{
ho2021classifierfree,
title={Classifier-Free Diffusion Guidance},
author={Jonathan Ho and Tim Salimans},
booktitle={NeurIPS 2021 Workshop on Deep Generative Models and Downstream Applications},
year={2021},
url={https://openreview.net/forum?id=qw8AKxfYbI}
}

@InProceedings{10.1007/978-3-031-19790-1_26,
author="Liu, Nan
and Li, Shuang
and Du, Yilun
and Torralba, Antonio
and Tenenbaum, Joshua B.",
editor="Avidan, Shai
and Brostow, Gabriel
and Ciss{\'e}, Moustapha
and Farinella, Giovanni Maria
and Hassner, Tal",
title="Compositional Visual Generation with Composable Diffusion Models",
booktitle="Computer Vision -- ECCV 2022",
year="2022",
publisher="Springer Nature Switzerland",
address="Cham",
pages="423--439",
isbn="978-3-031-19790-1"
}

@INPROCEEDINGS {10204579,
author = { Brooks, Tim and Holynski, Aleksander and Efros, Alexei A. },
booktitle = { 2023 IEEE/CVF Conference on Computer Vision and Pattern Recognition (CVPR) },
title = {{ InstructPix2Pix: Learning to Follow Image Editing Instructions }},
year = {2023},
volume = {},
ISSN = {},
pages = {18392-18402},
doi = {10.1109/CVPR52729.2023.01764},
url = {https://doi.ieeecomputersociety.org/10.1109/CVPR52729.2023.01764},
publisher = {IEEE Computer Society},
address = {Los Alamitos, CA, USA},
month =Jun}

@InProceedings{Huang_2024_CVPR,
    author    = {Huang, Ziqi and He, Yinan and Yu, Jiashuo and Zhang, Fan and Si, Chenyang and Jiang, Yuming and Zhang, Yuanhan and Wu, Tianxing and Jin, Qingyang and Chanpaisit, Nattapol and Wang, Yaohui and Chen, Xinyuan and Wang, Limin and Lin, Dahua and Qiao, Yu and Liu, Ziwei},
    title     = {VBench: Comprehensive Benchmark Suite for Video Generative Models},
    booktitle = {Proceedings of the IEEE/CVF Conference on Computer Vision and Pattern Recognition (CVPR)},
    month     = {June},
    year      = {2024},
    pages     = {21807-21818}
}

@INPROCEEDINGS {9878449,
author = { Rombach, Robin and Blattmann, Andreas and Lorenz, Dominik and Esser, Patrick and Ommer, Bjorn },
booktitle = { 2022 IEEE/CVF Conference on Computer Vision and Pattern Recognition (CVPR) },
title = {{ High-Resolution Image Synthesis with Latent Diffusion Models }},
year = {2022},
volume = {},
ISSN = {},
pages = {10674-10685},
doi = {10.1109/CVPR52688.2022.01042},
url = {https://doi.ieeecomputersociety.org/10.1109/CVPR52688.2022.01042},
publisher = {IEEE Computer Society},
address = {Los Alamitos, CA, USA},
month =Jun}

@article{Blattmann2023StableVD,
  title={Stable Video Diffusion: Scaling Latent Video Diffusion Models to Large Datasets},
  author={A. Blattmann and Tim Dockhorn and Sumith Kulal and Daniel Mendelevitch and Maciej Kilian and Dominik Lorenz},
  journal={ArXiv},
  year={2023},
  volume={abs/2311.15127},
  url={https://api.semanticscholar.org/CorpusID:265312551}
}

@INPROCEEDINGS {10377858,
author = { Peebles, William and Xie, Saining },
booktitle = { 2023 IEEE/CVF International Conference on Computer Vision (ICCV) },
title = {{ Scalable Diffusion Models with Transformers }},
year = {2023},
volume = {},
ISSN = {},
pages = {4172-4182},
doi = {10.1109/ICCV51070.2023.00387},
url = {https://doi.ieeecomputersociety.org/10.1109/ICCV51070.2023.00387},
publisher = {IEEE Computer Society},
address = {Los Alamitos, CA, USA},
month =Oct}

@INPROCEEDINGS{10377881,
  author={Zhang, Lvmin and Rao, Anyi and Agrawala, Maneesh},
  booktitle={2023 IEEE/CVF International Conference on Computer Vision (ICCV)}, 
  title={Adding Conditional Control to Text-to-Image Diffusion Models}, 
  year={2023},
  volume={},
  number={},
  pages={3813-3824},
  doi={10.1109/ICCV51070.2023.00355}}

@inproceedings{
hu2022lora,
title={Lo{RA}: Low-Rank Adaptation of Large Language Models},
author={Edward J Hu and yelong shen and Phillip Wallis and Zeyuan Allen-Zhu and Yuanzhi Li and Shean Wang and Lu Wang and Weizhu Chen},
booktitle={International Conference on Learning Representations},
year={2022},
url={https://openreview.net/forum?id=nZeVKeeFYf9}
}

@InProceedings{Zhang_2023_ICCV,
    author    = {Zhang, Lvmin and Rao, Anyi and Agrawala, Maneesh},
    title     = {Adding Conditional Control to Text-to-Image Diffusion Models},
    booktitle = {Proceedings of the IEEE/CVF International Conference on Computer Vision (ICCV)},
    month     = {October},
    year      = {2023},
    pages     = {3836-3847}
}

@InProceedings{pmlr-v97-houlsby19a,
  title = 	 {Parameter-Efficient Transfer Learning for {NLP}},
  author =       {Houlsby, Neil and Giurgiu, Andrei and Jastrzebski, Stanislaw and Morrone, Bruna and De Laroussilhe, Quentin and Gesmundo, Andrea and Attariyan, Mona and Gelly, Sylvain},
  booktitle = 	 {Proceedings of the 36th International Conference on Machine Learning},
  pages = 	 {2790--2799},
  year = 	 {2019},
  editor = 	 {Chaudhuri, Kamalika and Salakhutdinov, Ruslan},
  volume = 	 {97},
  series = 	 {Proceedings of Machine Learning Research},
  month = 	 {09--15 Jun},
  publisher =    {PMLR},
  url = 	 {https://proceedings.mlr.press/v97/houlsby19a.html}
}

@misc{ma2025stepvideot2vtechnicalreportpractice,
      title={Step-Video-T2V Technical Report: The Practice, Challenges, and Future of Video Foundation Model}, 
      author={Guoqing Ma and Haoyang Huang and Kun Yan and Liangyu Chen and Nan Duan and Shengming Yin and Changyi Wan and Ranchen Ming and Xiaoniu Song and Xing Chen and Yu Zhou and Deshan Sun and Deyu Zhou and Jian Zhou and Kaijun Tan and Kang An and Mei Chen and Wei Ji and Qiling Wu and Wen Sun and Xin Han and Yanan Wei and Zheng Ge and Aojie Li and Bin Wang and Bizhu Huang and Bo Wang and Brian Li and Changxing Miao and Chen Xu and Chenfei Wu and Chenguang Yu and Dapeng Shi and Dingyuan Hu and Enle Liu and Gang Yu and Ge Yang and Guanzhe Huang and Gulin Yan and Haiyang Feng and Hao Nie and Haonan Jia and Hanpeng Hu and Hanqi Chen and Haolong Yan and Heng Wang and Hongcheng Guo and Huilin Xiong and Huixin Xiong and Jiahao Gong and Jianchang Wu and Jiaoren Wu and Jie Wu and Jie Yang and Jiashuai Liu and Jiashuo Li and Jingyang Zhang and Junjing Guo and Junzhe Lin and Kaixiang Li and Lei Liu and Lei Xia and Liang Zhao and Liguo Tan and Liwen Huang and Liying Shi and Ming Li and Mingliang Li and Muhua Cheng and Na Wang and Qiaohui Chen and Qinglin He and Qiuyan Liang and Quan Sun and Ran Sun and Rui Wang and Shaoliang Pang and Shiliang Yang and Sitong Liu and Siqi Liu and Shuli Gao and Tiancheng Cao and Tianyu Wang and Weipeng Ming and Wenqing He and Xu Zhao and Xuelin Zhang and Xianfang Zeng and Xiaojia Liu and Xuan Yang and Yaqi Dai and Yanbo Yu and Yang Li and Yineng Deng and Yingming Wang and Yilei Wang and Yuanwei Lu and Yu Chen and Yu Luo and Yuchu Luo and Yuhe Yin and Yuheng Feng and Yuxiang Yang and Zecheng Tang and Zekai Zhang and Zidong Yang and Binxing Jiao and Jiansheng Chen and Jing Li and Shuchang Zhou and Xiangyu Zhang and Xinhao Zhang and Yibo Zhu and Heung-Yeung Shum and Daxin Jiang},
      year={2025},
      eprint={2502.10248},
      archivePrefix={arXiv},
      primaryClass={cs.CV},
      url={https://arxiv.org/abs/2502.10248}, 
}

@inproceedings{
zheng2023dpmsolverv,
title={{DPM}-Solver-v3: Improved Diffusion {ODE} Solver with Empirical Model Statistics},
author={Kaiwen Zheng and Cheng Lu and Jianfei Chen and Jun Zhu},
booktitle={Thirty-seventh Conference on Neural Information Processing Systems},
year={2023},
url={https://openreview.net/forum?id=9fWKExmKa0}
}

@inproceedings{10.1145/3680528.3687614,
author = {Bar-Tal, Omer and Chefer, Hila and Tov, Omer and Herrmann, Charles and Paiss, Roni and Zada, Shiran and Ephrat, Ariel and Hur, Junhwa and Liu, Guanghui and Raj, Amit and Li, Yuanzhen and Rubinstein, Michael and Michaeli, Tomer and Wang, Oliver and Sun, Deqing and Dekel, Tali and Mosseri, Inbar},
title = {Lumiere: A Space-Time Diffusion Model for Video Generation},
year = {2024},
isbn = {9798400711312},
publisher = {Association for Computing Machinery},
address = {New York, NY, USA},
url = {https://doi.org/10.1145/3680528.3687614},
doi = {10.1145/3680528.3687614},
booktitle = {SIGGRAPH Asia 2024 Conference Papers},
articleno = {94},
numpages = {11},
location = {Tokyo, Japan},
series = {SA '24}
}

@misc{kong2025hunyuanvideosystematicframeworklarge,
      title={HunyuanVideo: A Systematic Framework For Large Video Generative Models}, 
      author={Weijie Kong and Qi Tian and Zijian Zhang and Rox Min and Zuozhuo Dai and Jin Zhou and Jiangfeng Xiong and Xin Li and Bo Wu and Jianwei Zhang and Kathrina Wu and Qin Lin and Junkun Yuan and Yanxin Long and Aladdin Wang and Andong Wang and Changlin Li and Duojun Huang and Fang Yang and Hao Tan and Hongmei Wang and Jacob Song and Jiawang Bai and Jianbing Wu and Jinbao Xue and Joey Wang and Kai Wang and Mengyang Liu and Pengyu Li and Shuai Li and Weiyan Wang and Wenqing Yu and Xinchi Deng and Yang Li and Yi Chen and Yutao Cui and Yuanbo Peng and Zhentao Yu and Zhiyu He and Zhiyong Xu and Zixiang Zhou and Zunnan Xu and Yangyu Tao and Qinglin Lu and Songtao Liu and Dax Zhou and Hongfa Wang and Yong Yang and Di Wang and Yuhong Liu and Jie Jiang and Caesar Zhong},
      year={2025},
      eprint={2412.03603},
      archivePrefix={arXiv},
      primaryClass={cs.CV},
      url={https://arxiv.org/abs/2412.03603}, 
}

@misc{zheng2024opensorademocratizingefficientvideo,
      title={Open-Sora: Democratizing Efficient Video Production for All}, 
      author={Zangwei Zheng and Xiangyu Peng and Tianji Yang and Chenhui Shen and Shenggui Li and Hongxin Liu and Yukun Zhou and Tianyi Li and Yang You},
      year={2024},
      eprint={2412.20404},
      archivePrefix={arXiv},
      primaryClass={cs.CV},
      url={https://arxiv.org/abs/2412.20404}, 
}

@misc{huang2025stepvideoti2vtechnicalreportstateoftheart,
      title={Step-Video-TI2V Technical Report: A State-of-the-Art Text-Driven Image-to-Video Generation Model}, 
      author={Haoyang Huang and Guoqing Ma and Nan Duan and Xing Chen and Changyi Wan and Ranchen Ming and Tianyu Wang and Bo Wang and Zhiying Lu and Aojie Li and Xianfang Zeng and Xinhao Zhang and Gang Yu and Yuhe Yin and Qiling Wu and Wen Sun and Kang An and Xin Han and Deshan Sun and Wei Ji and Bizhu Huang and Brian Li and Chenfei Wu and Guanzhe Huang and Huixin Xiong and Jiaxin He and Jianchang Wu and Jianlong Yuan and Jie Wu and Jiashuai Liu and Junjing Guo and Kaijun Tan and Liangyu Chen and Qiaohui Chen and Ran Sun and Shanshan Yuan and Shengming Yin and Sitong Liu and Wei Chen and Yaqi Dai and Yuchu Luo and Zheng Ge and Zhisheng Guan and Xiaoniu Song and Yu Zhou and Binxing Jiao and Jiansheng Chen and Jing Li and Shuchang Zhou and Xiangyu Zhang and Yi Xiu and Yibo Zhu and Heung-Yeung Shum and Daxin Jiang},
      year={2025},
      eprint={2503.11251},
      archivePrefix={arXiv},
      primaryClass={cs.CV},
      url={https://arxiv.org/abs/2503.11251}, 
}

@misc{wu2025hunyuanvideo15technicalreport,
      title={HunyuanVideo 1.5 Technical Report}, 
      author={Bing Wu and Chang Zou and Changlin Li and Duojun Huang and Fang Yang and Hao Tan and Jack Peng and Jianbing Wu and Jiangfeng Xiong and Jie Jiang and Linus and Patrol and Peizhen Zhang and Peng Chen and Penghao Zhao and Qi Tian and Songtao Liu and Weijie Kong and Weiyan Wang and Xiao He and Xin Li and Xinchi Deng and Xuefei Zhe and Yang Li and Yanxin Long and Yuanbo Peng and Yue Wu and Yuhong Liu and Zhenyu Wang and Zuozhuo Dai and Bo Peng and Coopers Li and Gu Gong and Guojian Xiao and Jiahe Tian and Jiaxin Lin and Jie Liu and Jihong Zhang and Jiesong Lian and Kaihang Pan and Lei Wang and Lin Niu and Mingtao Chen and Mingyang Chen and Mingzhe Zheng and Miles Yang and Qiangqiang Hu and Qi Yang and Qiuyong Xiao and Runzhou Wu and Ryan Xu and Rui Yuan and Shanshan Sang and Shisheng Huang and Siruis Gong and Shuo Huang and Weiting Guo and Xiang Yuan and Xiaojia Chen and Xiawei Hu and Wenzhi Sun and Xiele Wu and Xianshun Ren and Xiaoyan Yuan and Xiaoyue Mi and Yepeng Zhang and Yifu Sun and Yiting Lu and Yitong Li and You Huang and Yu Tang and Yixuan Li and Yuhang Deng and Yuan Zhou and Zhichao Hu and Zhiguang Liu and Zhihe Yang and Zilin Yang and Zhenzhi Lu and Zixiang Zhou and Zhao Zhong},
      year={2025},
      eprint={2511.18870},
      archivePrefix={arXiv},
      primaryClass={cs.CV},
      url={https://arxiv.org/abs/2511.18870}, 
}

@misc{wan2025wanopenadvancedlargescale,
      title={Wan: Open and Advanced Large-Scale Video Generative Models}, 
      author={Team Wan and Ang Wang and Baole Ai and Bin Wen and Chaojie Mao and Chen-Wei Xie and Di Chen and Feiwu Yu and Haiming Zhao and Jianxiao Yang and Jianyuan Zeng and Jiayu Wang and Jingfeng Zhang and Jingren Zhou and Jinkai Wang and Jixuan Chen and Kai Zhu and Kang Zhao and Keyu Yan and Lianghua Huang and Mengyang Feng and Ningyi Zhang and Pandeng Li and Pingyu Wu and Ruihang Chu and Ruili Feng and Shiwei Zhang and Siyang Sun and Tao Fang and Tianxing Wang and Tianyi Gui and Tingyu Weng and Tong Shen and Wei Lin and Wei Wang and Wei Wang and Wenmeng Zhou and Wente Wang and Wenting Shen and Wenyuan Yu and Xianzhong Shi and Xiaoming Huang and Xin Xu and Yan Kou and Yangyu Lv and Yifei Li and Yijing Liu and Yiming Wang and Yingya Zhang and Yitong Huang and Yong Li and You Wu and Yu Liu and Yulin Pan and Yun Zheng and Yuntao Hong and Yupeng Shi and Yutong Feng and Zeyinzi Jiang and Zhen Han and Zhi-Fan Wu and Ziyu Liu},
      year={2025},
      eprint={2503.20314},
      archivePrefix={arXiv},
      primaryClass={cs.CV},
      url={https://arxiv.org/abs/2503.20314}, 
}

@inproceedings{
jin2025pyramidal,
title={Pyramidal Flow Matching for Efficient Video Generative Modeling},
author={Yang Jin and Zhicheng Sun and Ningyuan Li and Kun Xu and Kun Xu and Hao Jiang and Nan Zhuang and Quzhe Huang and Yang Song and Yadong MU and Zhouchen Lin},
booktitle={The Thirteenth International Conference on Learning Representations},
year={2025},
url={https://openreview.net/forum?id=66NzcRQuOq}
}

@inproceedings{
li2024tvturbo,
title={T2V-Turbo: Breaking the Quality Bottleneck of Video Consistency Model with Mixed Reward Feedback},
author={Jiachen Li and Weixi Feng and Tsu-Jui Fu and Xinyi Wang and S Basu and Wenhu Chen and William Yang Wang},
booktitle={The Thirty-eighth Annual Conference on Neural Information Processing Systems},
year={2024},
url={https://openreview.net/forum?id=53daI9kbvf}
}

@inproceedings{
li2025tvturbov,
title={T2V-Turbo-v2: Enhancing Video Model Post-Training through Data, Reward, and Conditional Guidance Design},
author={Jiachen Li and Qian Long and Jian Zheng and Xiaofeng Gao and Robinson Piramuthu and Wenhu Chen and William Yang Wang},
booktitle={The Thirteenth International Conference on Learning Representations},
year={2025},
url={https://openreview.net/forum?id=BZwXMqu4zG}
}

@ARTICLE{11250949,
author={Huang, Ziqi and Zhang, Fan and Xu, Xiaojie and He, Yinan and Yu, Jiashuo and Dong, Ziyue and Ma, Qianli and Chanpaisit, Nattapol and Si, Chenyang and Jiang, Yuming and Wang, Yaohui and Chen, Xinyuan and Chen, Ying-Cong and Wang, Limin and Lin, Dahua and Qiao, Yu and Liu, Ziwei},
journal={ IEEE Transactions on Pattern Analysis \& Machine Intelligence },
title={{ VBench++: Comprehensive and Versatile Benchmark Suite for Video Generative Models }},
year={2026},
volume={48},
number={03},
ISSN={1939-3539},
pages={3268-3285},
doi={10.1109/TPAMI.2025.3633890},
url = {https://doi.ieeecomputersociety.org/10.1109/TPAMI.2025.3633890},
publisher={IEEE Computer Society},
address={Los Alamitos, CA, USA},
month=mar}

@misc{zheng2025vbench20advancingvideogeneration,
      title={VBench-2.0: Advancing Video Generation Benchmark Suite for Intrinsic Faithfulness}, 
      author={Dian Zheng and Ziqi Huang and Hongbo Liu and Kai Zou and Yinan He and Fan Zhang and Lulu Gu and Yuanhan Zhang and Jingwen He and Wei-Shi Zheng and Yu Qiao and Ziwei Liu},
      year={2025},
      eprint={2503.21755},
      archivePrefix={arXiv},
      primaryClass={cs.CV},
      url={https://arxiv.org/abs/2503.21755}, 
}

@misc{
qin2025worldsimbench,
title={WorldSimBench: Towards Video Generation  Models as World Simulators},
author={Yiran Qin and Zhelun Shi and Jiwen Yu and Xijun Wang and Enshen Zhou and Lijun Li and Zhenfei Yin and Xihui Liu and Lu Sheng and Jing Shao and LEI BAI and Wanli Ouyang and Ruimao Zhang},
year={2025},
url={https://openreview.net/forum?id=ejGAytoWoe}
}

@inproceedings{
yang2025cogvideox,
title={CogVideoX: Text-to-Video Diffusion Models with An Expert Transformer},
author={Zhuoyi Yang and Jiayan Teng and Wendi Zheng and Ming Ding and Shiyu Huang and Jiazheng Xu and Yuanming Yang and Wenyi Hong and Xiaohan Zhang and Guanyu Feng and Da Yin and Yuxuan.Zhang and Weihan Wang and Yean Cheng and Bin Xu and Xiaotao Gu and Yuxiao Dong and Jie Tang},
booktitle={The Thirteenth International Conference on Learning Representations},
year={2025},
url={https://openreview.net/forum?id=LQzN6TRFg9}
}

@misc{maduabuchi2026corruptionawaretraininglatentvideo,
      title={Corruption-Aware Training of Latent Video Diffusion Models for Robust Text-to-Video Generation}, 
      author={Chika Maduabuchi and Hao Chen and Yujin Han and Jindong Wang},
      year={2026},
      eprint={2505.21545},
      archivePrefix={arXiv},
      primaryClass={cs.CV},
      url={https://arxiv.org/abs/2505.21545}, 
}

@misc{maduabuchi2026temporalpairconsistencyvariancereduced,
      title={Temporal Pair Consistency for Variance-Reduced Flow Matching}, 
      author={Chika Maduabuchi and Jindong Wang},
      year={2026},
      eprint={2602.04908},
      archivePrefix={arXiv},
      primaryClass={cs.LG},
      url={https://arxiv.org/abs/2602.04908}, 
}

@misc{maduabuchi2026entropycontrolledflowmatching,
      title={Entropy-Controlled Flow Matching}, 
      author={Chika Maduabuchi},
      year={2026},
      eprint={2602.22265},
      archivePrefix={arXiv},
      primaryClass={cs.LG},
      url={https://arxiv.org/abs/2602.22265}, 
}

@inproceedings{
lipman2023flow,
title={Flow Matching for Generative Modeling},
author={Yaron Lipman and Ricky T. Q. Chen and Heli Ben-Hamu and Maximilian Nickel and Matthew Le},
booktitle={The Eleventh International Conference on Learning Representations },
year={2023},
url={https://openreview.net/forum?id=PqvMRDCJT9t}
}

@INPROCEEDINGS{masala2023gest,
  author={Masala, Mihai and Cudlenco, Nicolae and Rebedea, Traian and Leordeanu, Marius},
  booktitle={2023 IEEE/CVF International Conference on Computer Vision Workshops (ICCVW)}, 
  title={Explaining Vision and Language through Graphs of Events in Space and Time}, 
  year={2023},
  volume={},
  number={},
  pages={2818-2823},
  doi={10.1109/ICCVW60793.2023.00302}}

@inproceedings{
cudlenco2026gestengine,
title={[Tiny Paper] {GEST}-Engine: Controllable Multi-Actor Video Synthesis with Perfect Spatiotemporal Annotations},
author={Nicolae Cudlenco and Mihai Masala and Marius Leordeanu},
booktitle={ICLR 2026 the 2nd Workshop on World Models: Understanding, Modelling and Scaling},
year={2026},
url={https://openreview.net/forum?id=uUofPYVMZH}
}

@INPROCEEDINGS{sun2025t2vcompbench,
  author={Sun, Kaiyue and Huang, Kaiyi and Liu, Xian and Wu, Yue and Xu, Zihan and Li, Zhenguo and Liu, Xihui},
  booktitle={2025 IEEE/CVF Conference on Computer Vision and Pattern Recognition (CVPR)}, 
  title={T2V-CompBench: A Comprehensive Benchmark for Compositional Text-to-video Generation}, 
  year={2025},
  volume={},
  number={},
  pages={8406-8416},
  doi={10.1109/CVPR52734.2025.00787}}

@INPROCEEDINGS{sharan2025neusv,
author = { Sharan, S P and Choi, Minkyu and Shah, Sahil and Goel, Harsh and Omama, Mohammad and Chinchali, Sandeep },
booktitle = { 2025 IEEE/CVF Conference on Computer Vision and Pattern Recognition (CVPR) },
title = {{ Neuro-Symbolic Evaluation of Text-to-Video Models using Formal Verification }},
year = {2025},
volume = {},
ISSN = {},
pages = {8395-8405},
doi = {10.1109/CVPR52734.2025.00786},
url = {https://doi.ieeecomputersociety.org/10.1109/CVPR52734.2025.00786},
publisher = {IEEE Computer Society},
address = {Los Alamitos, CA, USA},
month =Jun}

@INPROCEEDINGS{kodaira2025streamdiffusion,
  author={Kodaira, Akio and Xu, Chenfeng and Hazama, Toshiki and Yoshimoto, Takanori and Ohno, Kohei and Mitsuhori, Shogo and Sugano, Soichi and Cho, Hanying and Liu, Zhijian and Tomizuka, Masayoshi and Keutzer, Kurt},
  booktitle={2025 IEEE/CVF International Conference on Computer Vision (ICCV)}, 
  title={StreamDiffusion: A Pipeline-Level Solution for Real-Time Interactive Generation}, 
  year={2025},
  volume={},
  number={},
  pages={12371-12380},
  doi={10.1109/ICCV51701.2025.01150}}

@inproceedings{liang2025streamv2v,
 author = {Liang, Feng and Kodaira, Akio and Xu, Chenfeng and Tomizuka, Masayoshi and Keutzer, Kurt and Marculescu, Diana},
 booktitle = {International Conference on Learning Representations},
 editor = {Y. Yue and A. Garg and N. Peng and F. Sha and R. Yu},
 pages = {46425--46445},
 title = {Looking Backward: Streaming Video-to-Video Translation with Feature Banks},
 url = {https://proceedings.iclr.cc/paper_files/paper/2025/file/7280f65ed571b7b28321f2c7cf4c60c8-Paper-Conference.pdf},
 volume = {2025},
 year = {2025}
}

@InProceedings{munir2025objectalign,
    author    = {Munir, Mustafa and Goel, Harsh and Wei, Xiwen and Choi, Minkyu and Shah, Sahil and Bhardwaj, Kartikeya and Whatmough, Paul and Chinchali, Sandeep and Marculescu, Radu},
    title     = {ObjectAlign: Neuro-Symbolic Object Consistency Verification and Correction},
    booktitle = {Proceedings of the IEEE/CVF Conference on Computer Vision and Pattern Recognition (CVPR) Workshops},
    month     = {June},
    year      = {2026},
    pages     = {3458-3468}
}
